\RequirePackage{fix-cm}
\documentclass[smallcondensed]{svjour3}     
\smartqed  
\usepackage{graphicx}
\usepackage[table]{xcolor}
\usepackage{bm}
\usepackage{amsmath, amssymb}
\usepackage{mathrsfs}  
\usepackage{array}
\usepackage{epsfig}
\usepackage{enumerate}
\usepackage{braket}
\usepackage{multirow}
\usepackage{bbm}
\usepackage{color}
\usepackage{natbib}
 \usepackage{relsize}
\usepackage{graphicx,subcaption}
\usepackage{verbatim}
\newcommand{\bl[1]}{\boldsymbol{#1}}
\newcommand{\bz}{\bl[z]}
\newcommand{\bxi}{\bl[\xi]}
\newcommand{\bgamma}{\bl[\gamma]}
\newcommand{\bx}{\bl[x]}
\newcommand{\Dmat}{W}
\newcommand{\Domega}{D_\Omega}
\begin{document}

\title{Skew Gaussian Processes for Classification
}


\author{Alessio Benavoli  \and         Dario Azzimonti \and Dario Piga}


\institute{{Alessio Benavoli \at
Department of Computer Science and Information Systems, University of Limerick, Ireland \\
              Tel.: +353 61 213128\\
              \email{alessio.benavoli@ul.ie}           
           \and
           Dario Azzimonti and Dario Piga \at
           Dalle Molle Institute for Artificial Intelligence Research (IDSIA) - USI/SUPSI, Manno, Switzerland. 
}}

\date{Received: date / Accepted: date}

\maketitle

\begin{abstract}
Gaussian processes (GPs) are  distributions over
functions, which provide a Bayesian nonparametric approach to regression and classification. 
In spite of their success, GPs have limited use in some applications, for example, in some cases a symmetric distribution with respect to its mean is an unreasonable model.  This implies, for instance, that  the mean and the median coincide, while the mean and median in an asymmetric (skewed)  distribution can be different numbers.
In this paper, we  propose Skew-Gaussian processes (SkewGPs)    as a non-parametric  prior  over  functions.
A SkewGP extends the multivariate \textit{Unified Skew-Normal} distribution over finite dimensional vectors to a stochastic processes.
 The SkewGP class of distributions includes GPs
 and, therefore, SkewGPs inherit all good properties of GPs and increase their flexibility 
by allowing asymmetry in the probabilistic model.
By exploiting the fact that SkewGP and probit likelihood are conjugate model, we derive  closed form expressions for the marginal likelihood  and  predictive  distribution  of  
this new nonparametric classifier.
 We verify empirically that the  proposed SkewGP classifier 
 provides a better performance than a GP classifier based 
 on either Laplace's method  or Expectation Propagation.
\end{abstract}

\section{Introduction}
Gaussian processes  (GPs)  extend multivariate Gaussian distributions over finite dimensional vectors to infinite dimensionality. Specifically, a GP  defines a distribution over functions, that is each draw from a Gaussian process is a function. Therefore, GPs provide  a principled, practical, and probabilistic  approach to nonparametric  regression and classification and they have successfully been applied to different domains \citep{rasmussen2006gaussian}. 

GPs  have several desirable mathematical properties.
The most appealing of them is that, for regression with Gaussian noise, the prior distribution is \textit{conjugate} for the likelihood function. Therefore the Bayesian update step is analytic, as is computing the predictive distribution for the function behavior at unknown locations.
In spite of their success, GPs have several known shortcomings. 

First, the Gaussian distribution is not a ``heavy-tailed'' distribution, and so it is not robust to extreme outliers. 
{ Alternative to GPs have been proposed of which the most notable example is represented by the class of elliptical processes
\citep{fang2018symmetric}, such as  Student-t processes
\citep{o1991bayes,zhang2007semiparametric},  where  any  collection  of  function values has a desired elliptical distribution, with a covariance matrix built using a kernel. 
}

Second, the Gaussian distribution is symmetric with respect to its mean.  This implies, for instance, that  the mean and the median coincide, while the mean and median in an asymmetric (skewed)  distribution can be different numbers.
{
This constraint limits GPs' flexibility and affects
the coverage of their credible intervals (regions) --
%
especially when considering that symmetry must hold for all components of the (latent) function and that, as for instance
discussed by \cite{kuss2005assessing,nickisch2008approximations}, the exact posterior of a GP classifier is skewed.}

To overcome this second limitation, in this paper, we propose \textit{Skew-Gaussian processes} (SkewGPs)    as a non-parametric  prior  over  functions.
A SkewGP extends the multivariate \textit{Unified Skew-Normal} distribution defined over finite dimensional vectors to a stochastic process, i.e. a distribution over infinite dimensional objects. 
A SkewGP is completely defined by a location function, a scale function and three additional parameters that depend on a latent dimension: a skewness function, a truncation vector and a covariance matrix.
It is worth noting that a SkewGP reduces to a GP
 when the latent variables have dimension zero. 
 Therefore, SkewGPs inherit all good properties of GPs and increase their flexibility 
by allowing asymmetry in the probabilistic model.

{We focus on applying this new nonparametric model
to a \textit{classification} problem. In the case of parametric models, 
\cite{durante2018conjugate} shows  that the posterior distribution of a probit likelihood and Gaussian prior has a unified skew-normal distribution. Such a novel result allowed the author to efficiently compute full posterior inferences for Bayesian logistic regression (for small datasets $n\approx 100$). Moreover the author showed  that the unified skew-normal distribution is a \textit{conjugate prior} for the probit likelihood (without using this prior model for data analysis).} 

Here we extend this result to the nonparametric case, we  derive a semi-analytical expression for the posterior distribution of the latent function and predictive probabilities for SkewGPs.
The term \textit{semi-analytical}  is adopted to indicate
that posterior inferences require  the computation of the cumulative distribution function of a multivariate Gaussian distribution (i.e., the computation of Gaussian orthant probabilities).
By using a new formulation \citep{gessner2019integrals} of elliptical slice sampling \citep{pmlrv9murray10a}, \textit{lin-ess}, which permits efficient sampling
from a linearly constrained (e.g., orthant)  Gaussian domain,
we show that we can compute efficiently posterior inferences for SkewGP binary classifiers.
\text{Lin-ess} is a special case of elliptical slice sampling that leverages the analytic tractability of intersections of ellipses and hyperplanes to speed up the elliptical slice algorithm. In particular, this  guarantees  rejection-free sampling and it is therefore also highly parallelizable. 

{
The main contributions of this paper are
\begin{enumerate}
	\item we propose a new class of stochastic processes called Skew-Gaussian processes (SkewGP) which generalize GP models;
	\item we show that a SkewGP prior process is conjugate for the probit likelihood thus deriving for the first time the posterior distribution of a GP classifier in an analytic form;
	\item we derive an efficient way to learn the hyperparameters of SkewGP and compute Monte Carlo predictions using \textit{lin-ess}, showing that our model has similar bottleneck computational complexity of GPs;
	\item we evaluate the proposed SkewGP classifier against state-of-the-art implementations of the GP classifier which approximate the posterior with the Laplace method or with Expectation propagation;
	\item we show on a small image classification dataset that a SkewGP prior can lead to better uncertainty quantification than a GP prior. 
\end{enumerate}
}

\section{Background}
The skew normal distribution is a continuous probability distribution that generalises the normal distribution to allow for non-zero skewness. 
The probability density function (PDF) of the univariate skew-normal distribution with location $\xi \in \mathbb{R}$, scale $\sigma>0$ and skew parameter $\alpha \in \mathbb{R}$ is given by \citep{o1976bayes}:
$$
p(z)={\frac {2}{\sigma }}\phi \left({\frac {z-\xi }{\sigma }}\right)\Phi \left(\alpha \left({\frac {z-\xi }{\sigma }}\right)\right), \qquad z \in \mathbb{R}
$$
where $\phi$ and $\Phi$ are the PDF and, respectively, Cumulative Distribution Function (CDF) of the standard univariate Normal distribution.
This distribution has been generalised in several ways, see \citep{azzalini2013skew} for a review.
In particular, \cite{arellano2006unification} provided a
unification of the above generalizations within a
single and tractable multivariate \textit{Unified  Skew-Normal} distribution that satisfies closure properties for marginals and conditionals
and allows more flexibility due the introduction of additional
parameters.

\subsection{Unified Skew-Normal distribution}

\begin{figure}
	\centering
	\begin{tabular}{c @{\quad} c }
		\includegraphics[width=.48\linewidth]{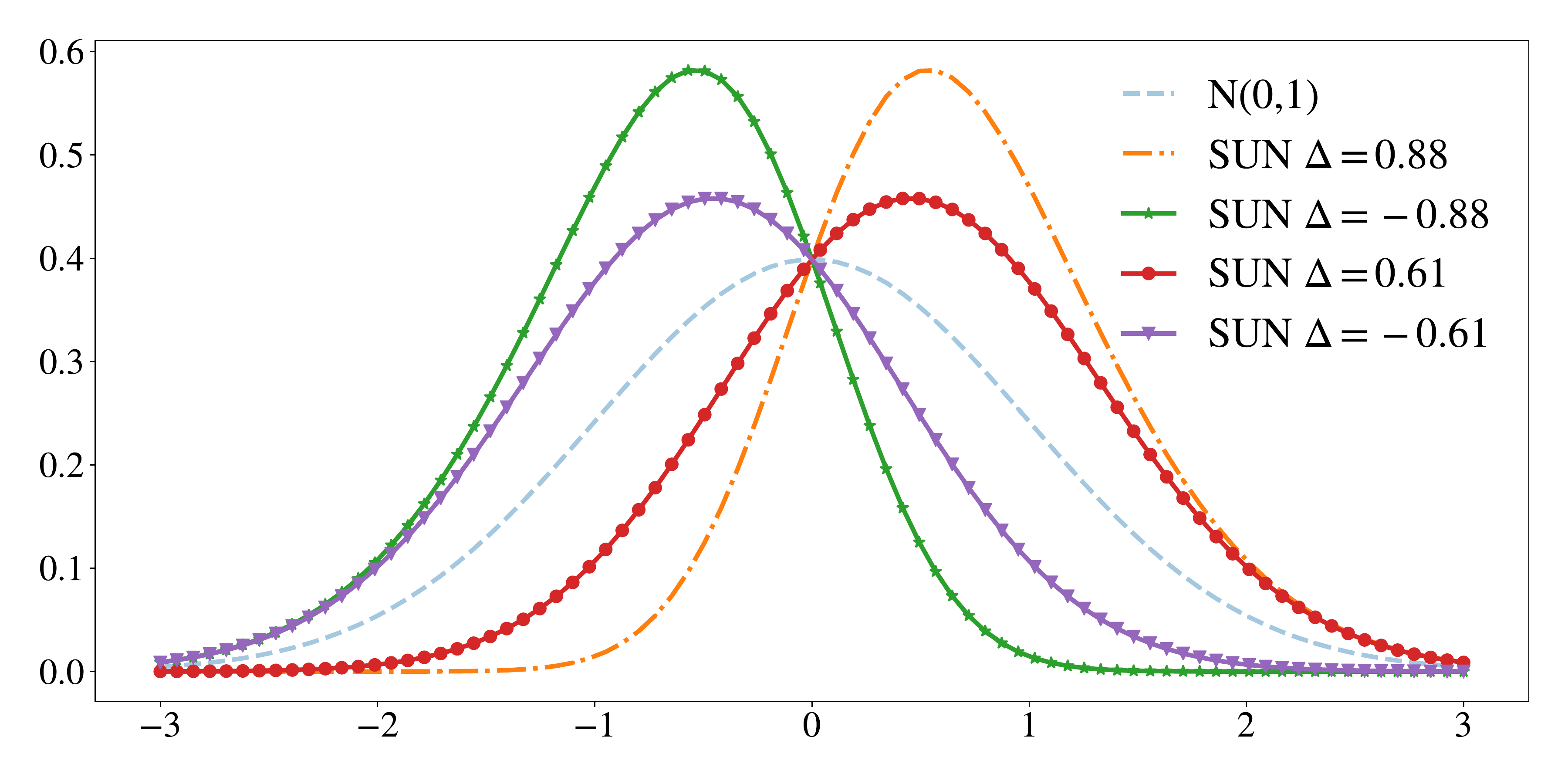} &
		\includegraphics[width=.48\linewidth]{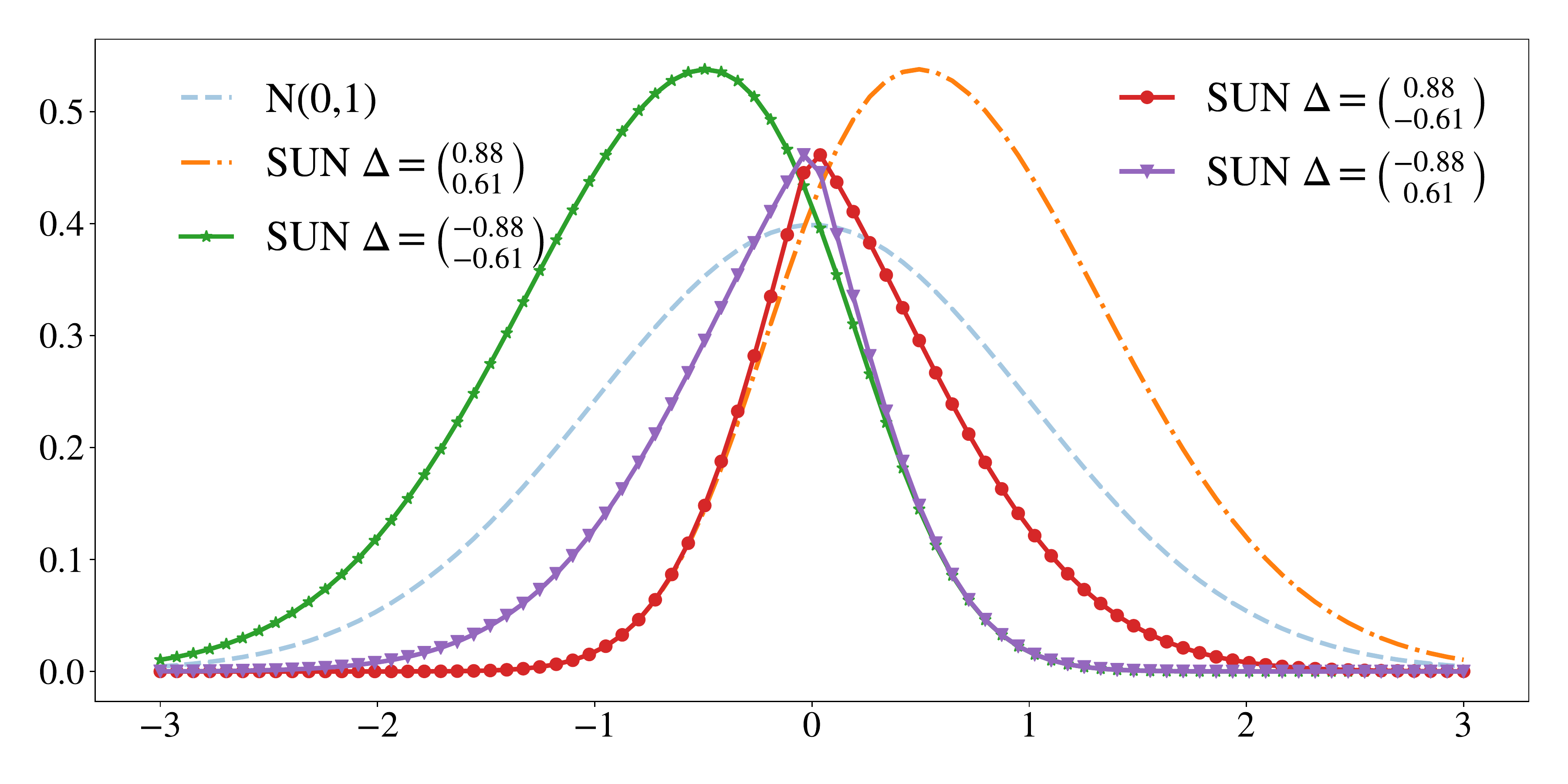} \\
		\small (a1) $s=1$, $\Gamma=1$  & \small (a2) $s=2$, $\Gamma_{1,2}=0.88$
	\end{tabular}
	\caption{Density plots for $\text{SUN}_{1,s}(0,1,\Delta,\gamma,\Gamma)$. For all plots $\Gamma$ is a correlation matrix, $\gamma = 0$, dashed lines are the contour plots of $y \sim N_1(0,1)$.}
	\label{fig:SUN1d}
\end{figure}

\begin{figure}
	\centering
	\begin{tabular}{c @{\quad} c @{\quad} c @{\quad} c}
		\includegraphics[width=.24\linewidth]{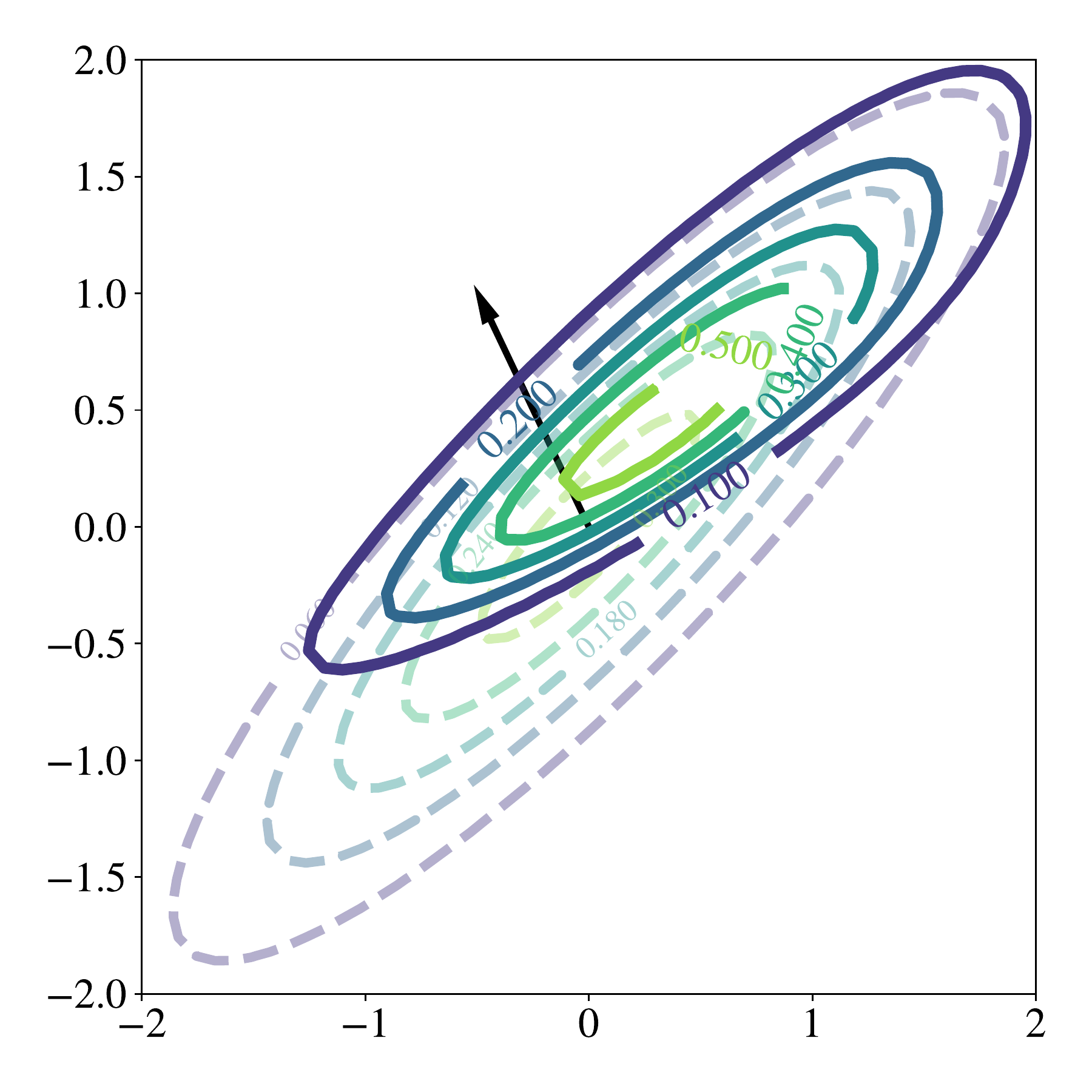} &
		\includegraphics[width=.24\linewidth]{{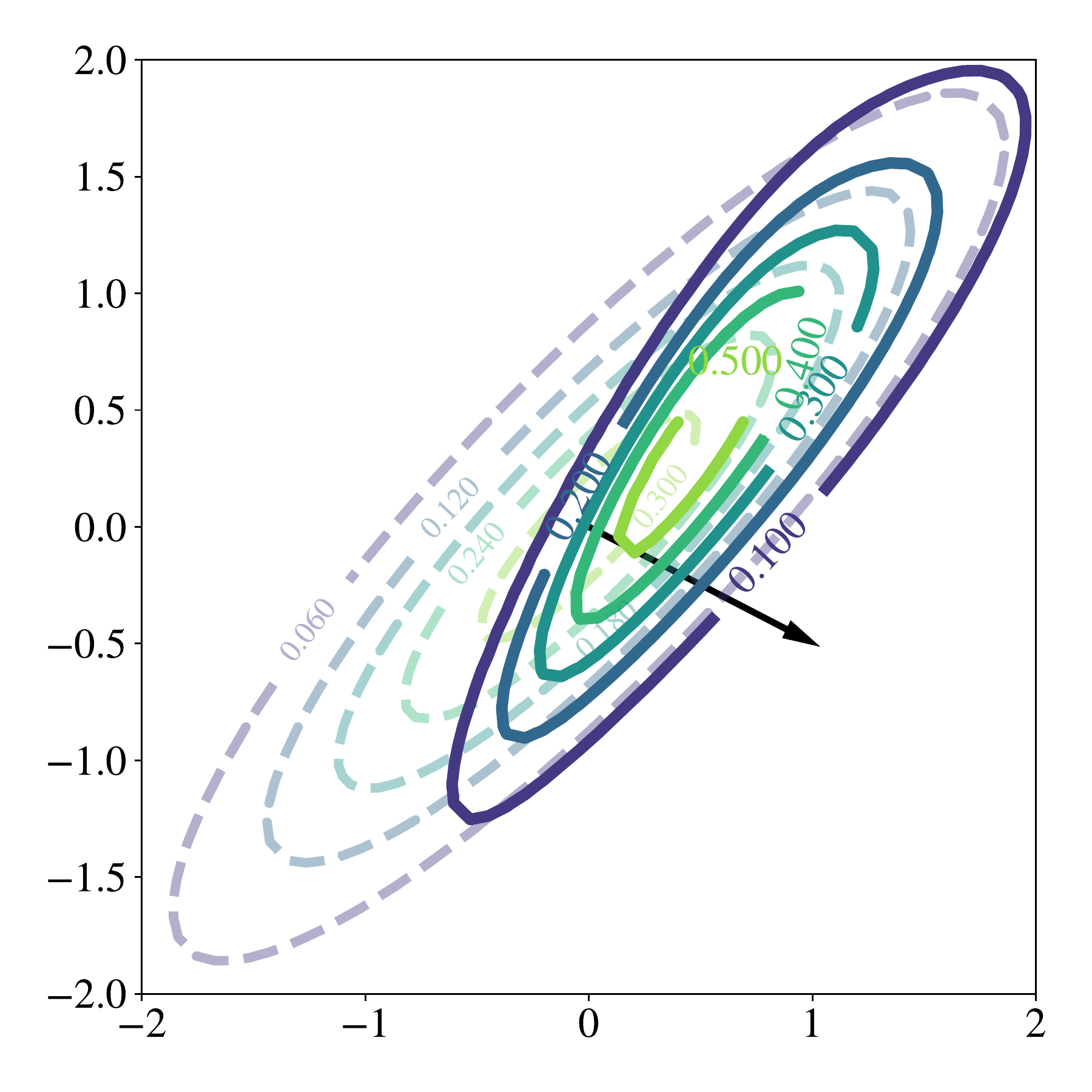}} &
		\includegraphics[width=.24\linewidth]{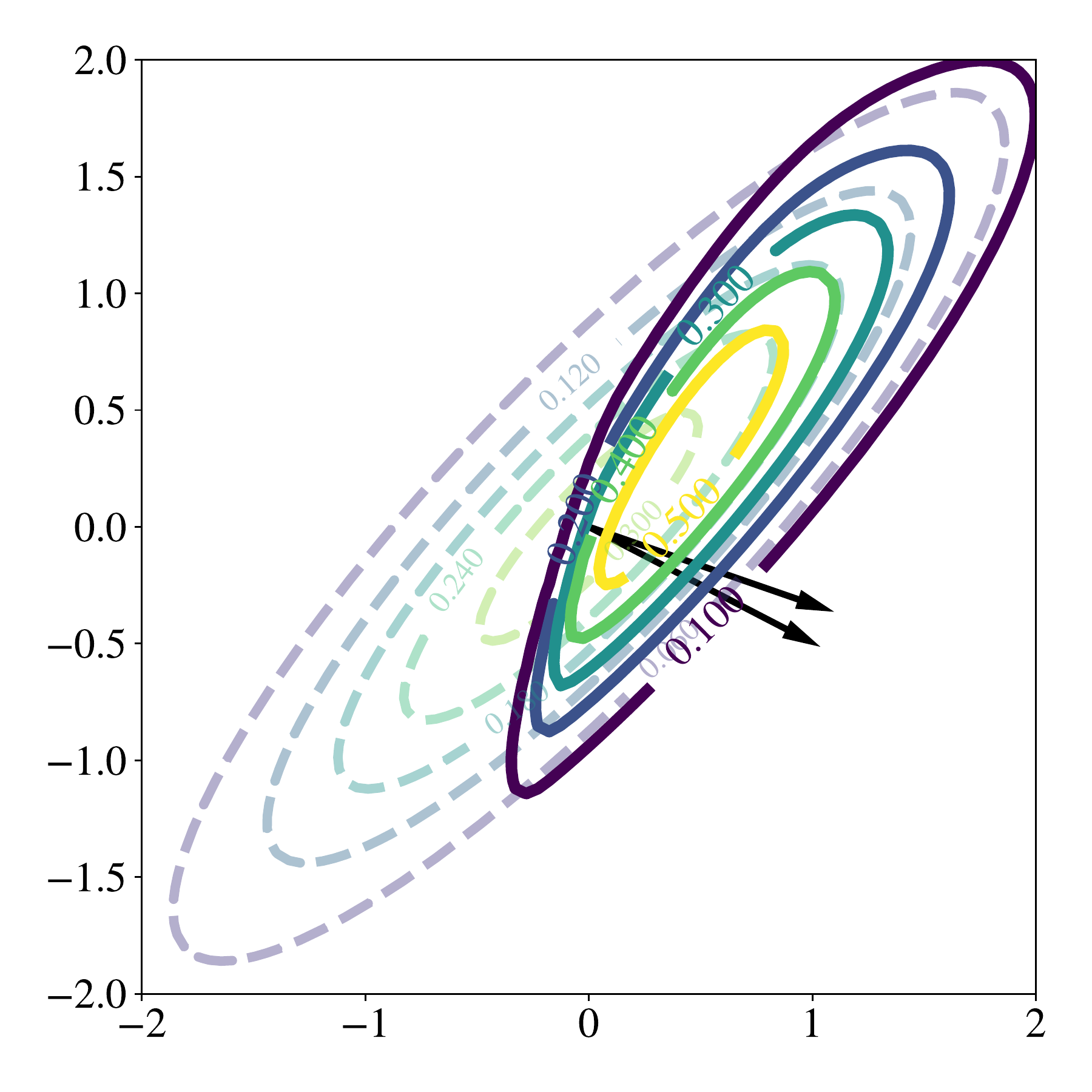} &
		\includegraphics[width=.24\linewidth]{{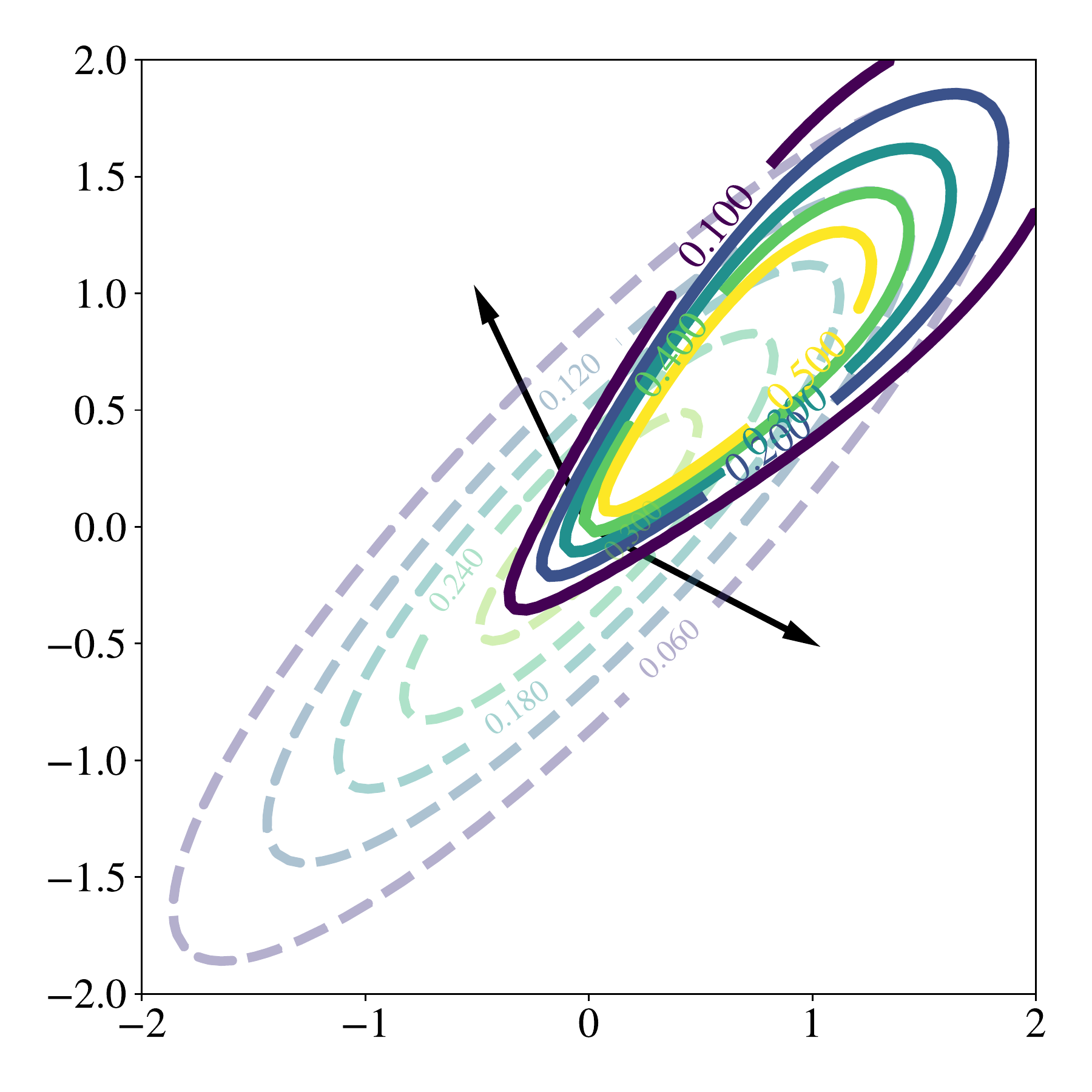}} \\
		\small (a1) $s=1$, $\Gamma=1$  & \small (a2) $s=1$, $\Gamma=1$ & \small (a3) $s=2$, $\Gamma_{1,2}=0.96$ & \small (a4) $s=2$, $\Gamma_{1,2}=0.33$ \\
		\small $\Delta = [0.88,0.61]^T$,  & \small  $\Delta = [0.61,0.88]^T$ & \small $\Delta = \begin{bmatrix}
		0.61 & 0.76 \\
		0.88 &, 0.97
		\end{bmatrix}$ & $\Delta = \begin{bmatrix}
		0.61 & 0.88 \\
		0.88 & 0.61
		\end{bmatrix}$ 
	\end{tabular}
	\caption{Contour density plots for four unified skew-normal. For all plots $p=2$, $\bxi=[0,0]^T$, $\Omega$ and $\Gamma$ are correlation matrices with $\Omega_{1,2}= 0.88$, $\gamma = 0$, dashed lines are the contour plots of $y \sim N_2(\bxi,\Omega)$.}
	\label{fig:SUN}
\end{figure}

A vector $\bz \in \mathbb{R}^p$ is said to have a multivariate
Unified Skew-Normal distribution with latent skewness dimension $s$, $ \bz \sim \text{SUN}_{p,s}(\bxi,\Omega,\Delta,\bgamma,\Gamma)$, if its probability density function \citep[Ch.7]{azzalini2013skew} is:
\begin{equation}
 \label{eq:sun}
 p(\bz) = \phi_p(\bz-\bxi;\Omega)\frac{\Phi_s\left(\bgamma+\Delta^T\bar{\Omega}^{-1}\Domega^{-1}(\bz-\bxi);\Gamma-\Delta^T\bar{\Omega}^{-1}\Delta\right)}{\Phi_s\left(\bgamma;\Gamma\right)} 
\end{equation}
where $\phi_p(\bz-\bxi;\Omega)$ represents the PDF of a multivariate Normal distribution with mean $\bxi \in \mathbb{R}^p$ and covariance $\Omega=\Domega\bar{\Omega} \Domega\in \mathbb{R}^{p\times p}$,
with $\bar{\Omega}$ being a correlation matrix and $\Domega$ a diagonal matrix  containing the square root of the diagonal elements in $\Omega$. The notation $\Phi_s(\bl[a];M)$ denotes the CDF of $N_s(0,M)$ evaluated at $\bl[a]\in \mathbb{R}^s$. 
 The parameters $\bgamma \in \mathbb{R}^s, \Gamma \in \mathbb{R}^{s\times s},\Delta^{p \times s}$ of the SUN distribution are related to a latent variable that controls the skewness, in particular $\Delta$ is called Skewness matrix.
The PDF  \eqref{eq:sun} is well-defined provided that the matrix
 \begin{equation}
  \label{eq:positivity}
   M:=\begin{bmatrix}
  \Gamma & \Delta^T\\
  \Delta & \bar{\Omega}
 \end{bmatrix} \in \mathbb{R}^{(s+p)\times(s+p)}>0,
 \end{equation}
 i.e., $M$ is positive definite. 
 { 
 Note that when $\Delta=0$, \eqref{eq:sun}  reduces  to  $\phi_p(\bz-\bxi;\Omega)$. Moreover we assume that $\Phi_0(\cdot)=1$, so that, for $s=0$, \eqref{eq:sun} becomes a multivariate Normal distribution. 
 
 The role of the latent dimension $s$ can be briefly explained as follows. 
 Consider now a random vector $\begin{bmatrix}
 \bx_0 \\
 \bx_1 
 \end{bmatrix} \sim N_{s+p}(0,M)$ with $M$ as in \eqref{eq:positivity} and define $\mathbf{y}$ as the vector with distribution $(\bx_1 \mid \bx_0+\bgamma>0)$, then it can be shown \citep[Ch. 7]{azzalini2013skew} that $\bz = \bxi + \Domega\mathbf{y}\sim\text{SUN}_{p,s}(\bxi,\Omega,\Delta,\bgamma,\Gamma)$. This representation will be used in Section \ref{sec:compComplex} to draw samples from the distribution. Figure~\ref{fig:SUN1d} shows the density of a univariate SUN distribution with latent dimensions $s=1$ (a1) and $s=2$ (a2). The effect of a higher latent dimension can be better observed in bivariate SUN densities as shown in  Figure~\ref{fig:SUN}. 
 The contours of the corresponding bivariate normal are dashed. We also plot the skewness directions given by $\bar{\Omega}^{-1}\Delta$.
}
 
 The Skew-Normal family has several interesting properties, see \citet[Ch.7]{azzalini2013skew} for details. Most notably, it is close under marginalization and affine transformations. 
Specifically, if we partition $z = [z_1 , z_2]^T$,
where $z_1 \in \mathbb{R}^{p_1}$ and $z_2 \in \mathbb{R}^{p_2}$
with $p_1+p_2=p$, then
\begin{equation}
\label{eq:marginalFinDim}
\begin{array}{c}
z_1  \sim SUN_{p_1,s}(\bxi_1,\Omega_{11},\Delta_1,\bgamma,\Gamma), \vspace{0.2cm}\\
\text{with }~~
\bxi=\begin{bmatrix}
 \bxi_1\\\bxi_2
\end{bmatrix},~~~
\Delta=\begin{bmatrix}
 \Delta_1\\\Delta_2
\end{bmatrix},~~~
\Omega=\begin{bmatrix}
 \Omega_{11} & \Omega_{12}\\
 \Omega_{21} & \Omega_{22}
\end{bmatrix}.
\end{array}
\end{equation}
Moreover, \citep[Ch.7]{azzalini2013skew} the conditional distribution is a unified skew-Normal, i.e.,  
%
$(Z_2|Z_1=z_1) \sim SUN_{p_2,s}(\bxi_{2|1},\Omega_{2|1},\Delta_{2|1},\bgamma_{2|1},\Gamma_{2|1})$, where 
\begin{align*}
\bxi_{2|1} & :=\bxi_{2}+\Omega_{21}\Omega_{11}^{-1}(z_1-\bxi_1), \quad
\Omega_{2|1} := \Omega_{22}-\Omega_{21}\Omega_{11}^{-1}\Omega_{12},\\
\Delta_{2|1} &:=\Delta_2 -\bar{\Omega}_{21}\bar{\Omega}_{11}^{-1}\Delta_1,\\
\bgamma_{2|1}& :=\bgamma+\Delta_1^T \Omega_{11}^{-1}(z_1-\bxi_1), \quad
\Gamma_{2|1}:=\Gamma-\Delta_1^T\bar{\Omega}_{11}^{-1}\Delta_1,
\end{align*}
and $\bar{\Omega}_{11}^{-1}:=(\bar{\Omega}_{11})^{-1}$.

\section{Skew-Gaussian process}
In this section, we define a Skew-Gaussian  Process (SkewGP). Consider the functions $\xi:\mathbb{R}^p \rightarrow \mathbb{R}$, a location function, $\Omega: \mathbb{R}^p \times \mathbb{R}^p \rightarrow \mathbb{R}$,
a scale function, $\Delta:\mathbb{R}^p \rightarrow \mathbb{R}^s$, the 
Skewness vector function, and $\bgamma \in \mathbb{R}^s,\Gamma \in \mathbb{R}^{s \times s}$.

We say that a real function $f: \mathbb{R}^p \rightarrow \mathbb{R}$ is distributed as a skew-Gaussian process with latent dimension $s$, if, for any sequence of $n$ points $\bx_1,\dots,\bx_n \in \mathbb{R}^p$, the vector
$f(X)=[f(\bx_1),\dots,f(\bx_n)]^T$ is Skew-Gaussian  distributed with parameters $\bgamma,\Gamma$, location, scale and skewness matrices,
respectively, given by
\begin{equation}
\begin{array}{rl}
\xi(X):=\begin{bmatrix}
\xi(\bx_1)\\
\xi(\bx_2)\\
\vdots\\
\xi(\bx_n)\\
\end{bmatrix},~~
\Omega(X,X)&:=
\begin{bmatrix}
\Omega(\bx_1,\bx_1) & \Omega(\bx_1,\bx_2) &\dots & \Omega(\bx_1,\bx_n)\\
\Omega(\bx_2,\bx_1) & \Omega(\bx_2,\bx_2) &\dots & \Omega(\bx_2,\bx_n)\\
\vdots & \vdots &\dots & \vdots\\
\Omega(\bx_n,\bx_1) & \Omega(\bx_n,\bx_2) &\dots & \Omega(\bx_n,\bx_n)\\
\end{bmatrix},\vspace{0.2cm}\\
\Delta(X)&:=\begin{bmatrix}
~~\Delta(\bx_1) & \quad\quad\Delta(\bx_2) &~~\dots & ~\quad\Delta(\bx_n)\\
\end{bmatrix}.
\end{array}
\end{equation}

The Skew-Gaussian distribution above is well defined
provided that the matrix 
\begin{equation*}
\begin{bmatrix}
\Gamma & \Delta(X) \\
\Delta(X)^T & \Omega(X,X)
\end{bmatrix}
\end{equation*}
is positive definite 
for all $X=\{\bx_1,\dots,\bx_n\} \subset \mathbb{R}^p$ and for all $n$. In this case we can write 
 \begin{equation}
 \label{eq:SGprocess}
 f(\bx) \sim  \text{SkewGP}_{s}(\xi(\bx),\Omega(\bx,\bx'),\Delta(\bx,\bx'),\bgamma,\Gamma).
 \end{equation}
 {
 We detail how to select the parameters in Section~\ref{sec:hyperparams}, the proposition below shows that SkewGP is a proper stochastic process.

\begin{proposition}
The construction of a Skew-Gaussian process from a finite-dimensional distribution is well-posed.
\label{prop:1}
\end{proposition}
All the proofs are in appendix.
}

\subsection{Binary classification}
Consider the training data  $\mathcal{D}=\{(\bx_i,y_i)\}_{i=1}^n$,
where $\bx_i \in \mathbb{R}^p$ and $ y_i \in \{0,1\}$.
We aim to build a nonparametric binary  classifier. We first define a probabilistic model $\mathcal{M}$ by assuming that
$f \sim \text{SkewGP}(\xi, \Omega, \Delta, \bgamma, \Gamma)$ and considering
a \textit{probit} model for the likelihood:
\begin{equation}
 \label{eq:probit}
 \begin{aligned}
  p(\mathcal{D}|f)&=\prod\limits_{i=1}^n \Phi(f(\bx_i);1)^{y_i}(1-\Phi(f(\bx_i);1))^{1-y_i}=\prod\limits_{i=1}^n \Phi((2y_i-1)f(\bx_i);1)\\
 &=\Phi_n(\Dmat f(X);I_n),
 \end{aligned} 
\end{equation}
where $\Dmat=\text{diag}(2y_1-1,\dots,2y_n-1)$.
A SkewGP prior combined with a
probit likelihood gives rise to a posterior SkewGP over functions,  this because Skew-Gaussian distributions
are conjugate priors for probit models. {In the finite dimensional parametric case, this property was shown by \cite{durante2018conjugate}, hereafter we extend it to the nonparametric one.}
\begin{theorem}
\label{th:1}
The posterior of $f(X)$ is a Skew-Gaussian distribution:
\begin{align}
\label{eq:posteriorclass}
p(f(X)|\mathcal{D})&= \text{SUN}_{n,s+n}(\tilde{\xi},\tilde{\Omega},\tilde{\Delta},\tilde{\bgamma},\tilde{\Gamma})\\
\tilde{\xi} & =\xi,\qquad
\tilde{\Omega} = \Omega, \\
\tilde{\Delta} &=[\Delta,~~\bar{\Omega}\Domega \Dmat^T],\\
\tilde{\bgamma}& =[\bgamma,~~\Dmat\xi]^T, \\
\tilde{\Gamma}&=\begin{bmatrix}
         \Gamma & ~~\Delta^T \Domega \Dmat^T \\
        \Dmat \Domega \Delta & ~~(\Dmat \Omega \Dmat^T + I_n) \end{bmatrix},
\end{align}
where, for simplicity of notation, we have denoted $\xi(X),\Omega(X,X),\Delta(X)$ as $\xi,\Omega,\Delta $ and
$\Omega = \Domega \bar{\Omega} \Domega$.
\end{theorem}
From  Theorem
 \ref{th:1} we can immediately derive the following result.
\begin{corollary}
\label{co:ml}
The marginal likelihood of the observations $\mathcal{D}=\{(\bx_i,y_i)\}_{i=1}^n$ given the probabilistic model $\mathcal{M}$, that is the prior \eqref{eq:SGprocess} and 
 likelihood \eqref{eq:probit}, is
 \begin{equation}
  p(\mathcal{D}|\mathcal{M}) = \frac{ \Phi_{s+n}(\tilde{\bgamma};~\tilde{\Gamma})}{\Phi_{s}(\bgamma;~\Gamma)},
  \label{eq:marginalLikelihood}
 \end{equation}
 with $\tilde{\bgamma},\tilde{\Gamma}$ defined in Theorem
 \ref{th:1}.
\end{corollary}
In classification, based on the training data  $\mathcal{D}=\{(\bx_i,y_i)\}_{i=1}^n$, and given 
 test inputs $\bx^* $, we aim to predict the probability that $y^*=1$.
 \begin{corollary}
 \label{co:predictive}
 The posterior predictive probability of $y^*=1$ given the test input $\bx^* \in \mathbb{R}^p$ and the training data $\mathcal{D}=\{(\bx_i,y_i)\}_{i=1}^n$ is
 \begin{equation}
  p(y^*=1|\bx^*,X,y) = \frac{\Phi_{s+n+1}(\tilde{\bgamma}^*;~\tilde{\Gamma}^*)}{\Phi_{s+n}(\tilde{\bgamma};~\tilde{\Gamma})},
  \label{eq:predPost}
 \end{equation}
 where $\tilde{\bgamma}^*,\tilde{\Gamma}^*$ are the corresponding matrices  of the posterior computed for the augmented dataset
 $\hat{X}=[X^T,{\bx^*}^T]^T$, $\hat{y}=[y^T,1/2]^T$.
\end{corollary}

{
Note that the dummy value $\frac{1}{2}$ in $\hat{y}$ does not influence the value of $p(y^*=1 | \bx^*, X,y)$ and it was chosen only for mathematical convenience, as it allows for marginalization over $f(\bx^*)$ and to derive the expression of $\tilde{\bgamma}^*,\tilde{\Gamma}^*$ similarly to the ones in Theorem \ref{th:1}.\footnote{Note in fact that, for $y=1/2$, the likelihood  $\Phi((2y-1)f(\bx^*),1)=0.5$ and  it does not depend on $f(\bx^*)$, and so it is marginalised out.}
}

\section{Prior functions, parameters and hyperparameters}
\label{sec:hyperparams}
A SkewGP prior is completely defined by
the location function $\xi(\bx)$, the scale function $\Omega(\bx,\bx')$,
the latent dimension $s\in \mathbb{N}$, the skewness vector function $\Delta(\bx,\bx') \in \mathbb{R}^s$  and $\bgamma \in \mathbb{R}^s,\Gamma \in \mathbb{R}^{s \times s}$.
As it is common for GPs,  we will take the location function $\xi(\bx)$ to be zero, although this need not be done.
Let $K(\bx,\bx')$ be a  positive definite covariance function (kernel) and let  $\Omega = K(X,X)$ be the covariance matrix obtained
 by applying $K(\bx,\bx')$  elementwise to the training data $X$.
In this paper, we propose the following way to
define the location, scale and skewness
functions of a SkewGP:
 \begin{equation}
  \label{eq:positivityKernel}
  M= \begin{bmatrix}
  \Gamma & \Delta(X,R)^T\\
  \Delta(X,R) & \bar{\Omega}(X,X)
 \end{bmatrix}
 :=   \begin{bmatrix}
  L\bar{K}(R,R)L & ~~L\bar{K}(R,X)\\
  \bar{K}(X,R)L & ~~\bar{K}(X,X)
 \end{bmatrix},
 \end{equation}
 with $L \in \mathbb{R}^{s \times s}$ is a diagonal matrix
 whose elements  $L_{ii} \in \{-1,1\}$ (it is a phase),
 $R=[\bl[r]_1,\dots,\bl[r]_s]^T \in \mathbb{R}^{s \times p}$ is
 a vector of $s$ pseudo-points  and $\bar{K}(\bx,\bx')=\frac{1}{\sigma^2}K(\bx,\bx')$ (for stationary kernels) with $K(\bx,\bx')$ being the kernel function
 and $\sigma^2$ the variance parameter of the kernel, e.g.,
 for the RBF kernel
 $$
 K(\bx,\bx') := \sigma^2 \exp \left(-\frac {\|{\bx} -{\bx'} \|^{2}}{2\ell^{2}}\right).
 $$
 It can easily be proven that $M>0$ and, therefore,
 \eqref{eq:positivity} holds.
 We select the parameters of the kernel $\sigma,\ell$, the locations $\bl[r]_i$ of the pseudo-points and the phase diagonal matrix $L$ by maximizing the marginal likelihood. In particular we exploit the lower bound
 \eqref{eq:lowerbound} explained in Section~\ref{sec:compComplex}.

\begin{figure}
\centering
  \begin{tabular}{c @{\qquad} c }
    \includegraphics[width=.48\linewidth]{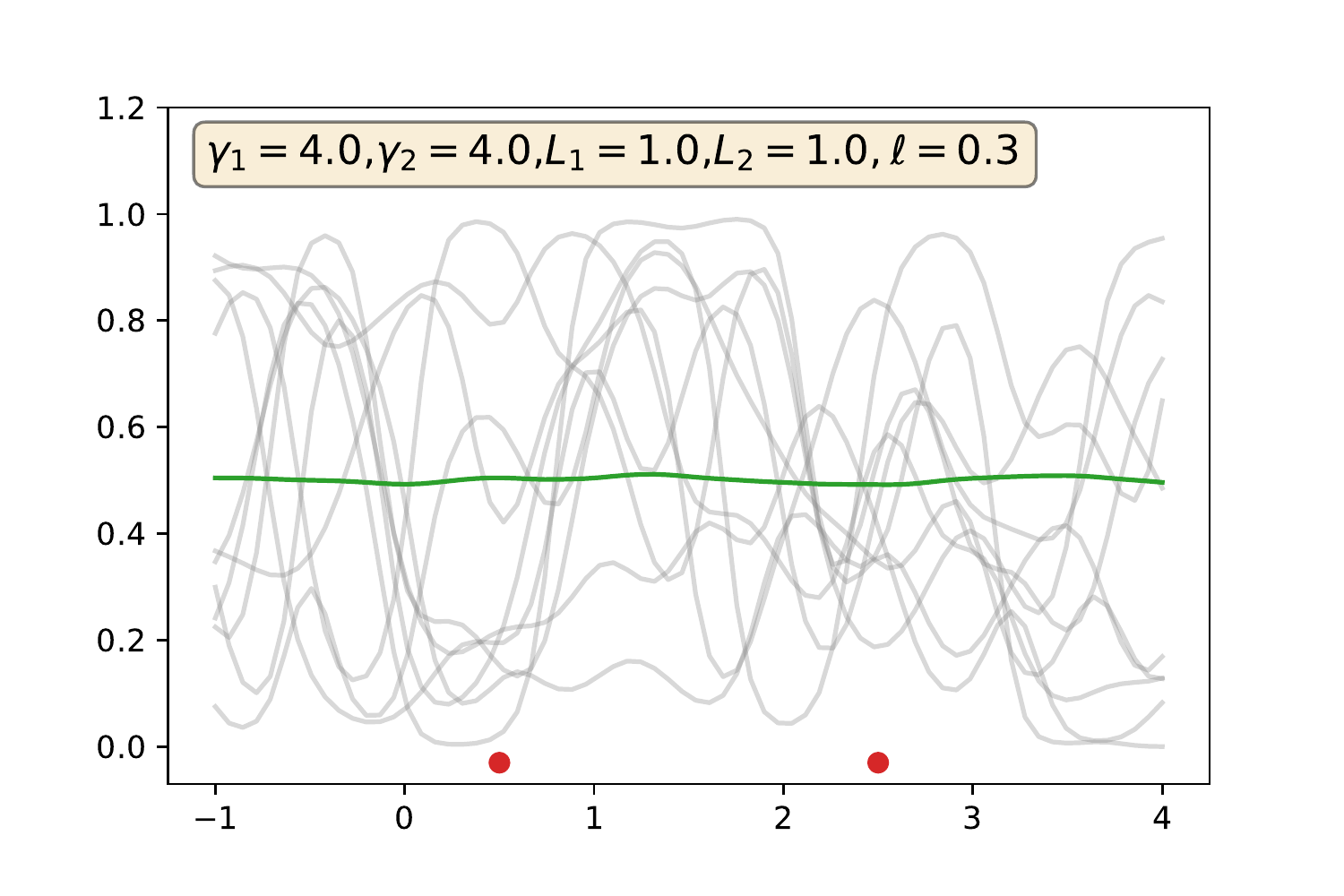} &
    \includegraphics[width=.48\linewidth]{{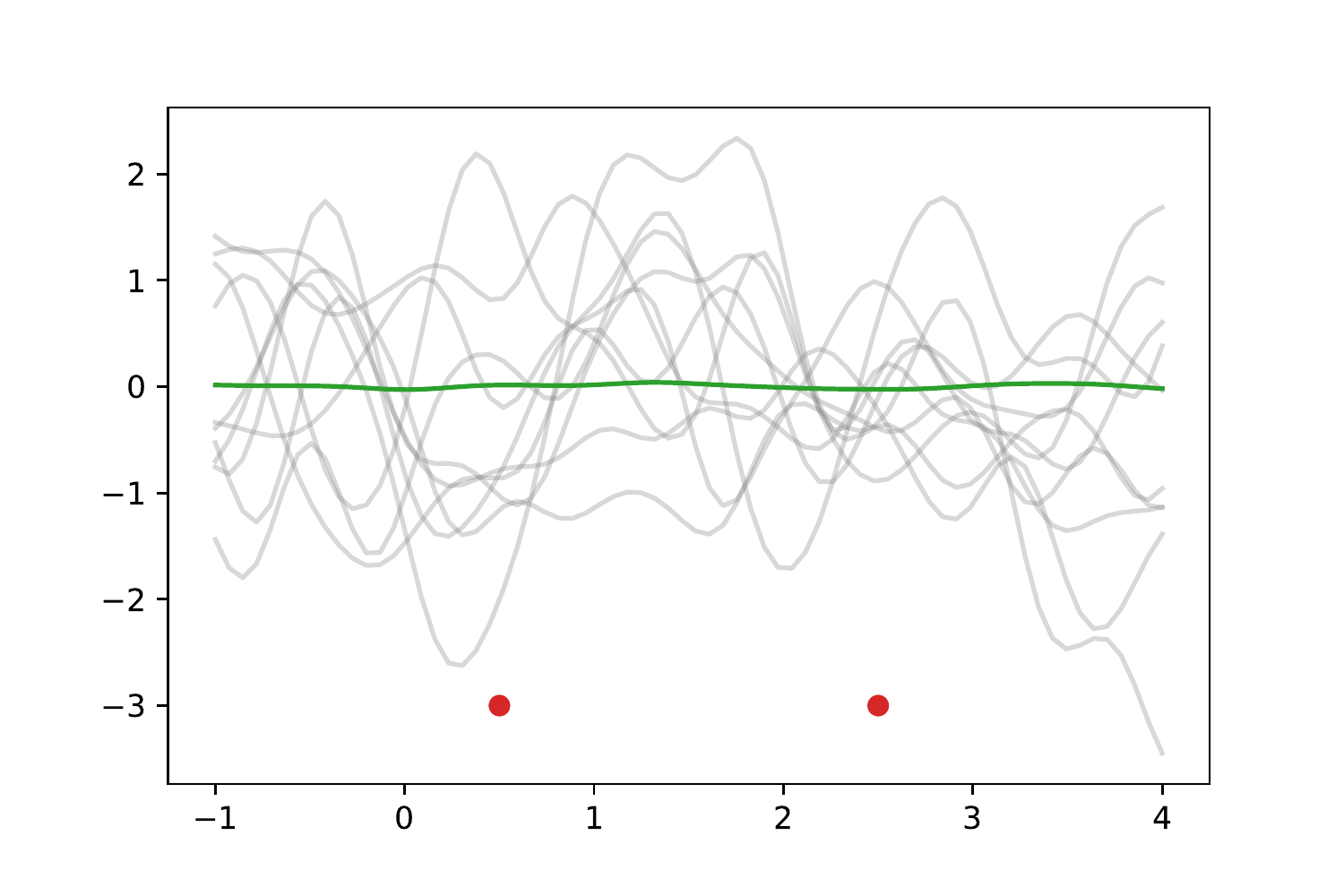}} \\
    \small (a1) & \small (a2)\\
        \includegraphics[width=.48\linewidth]{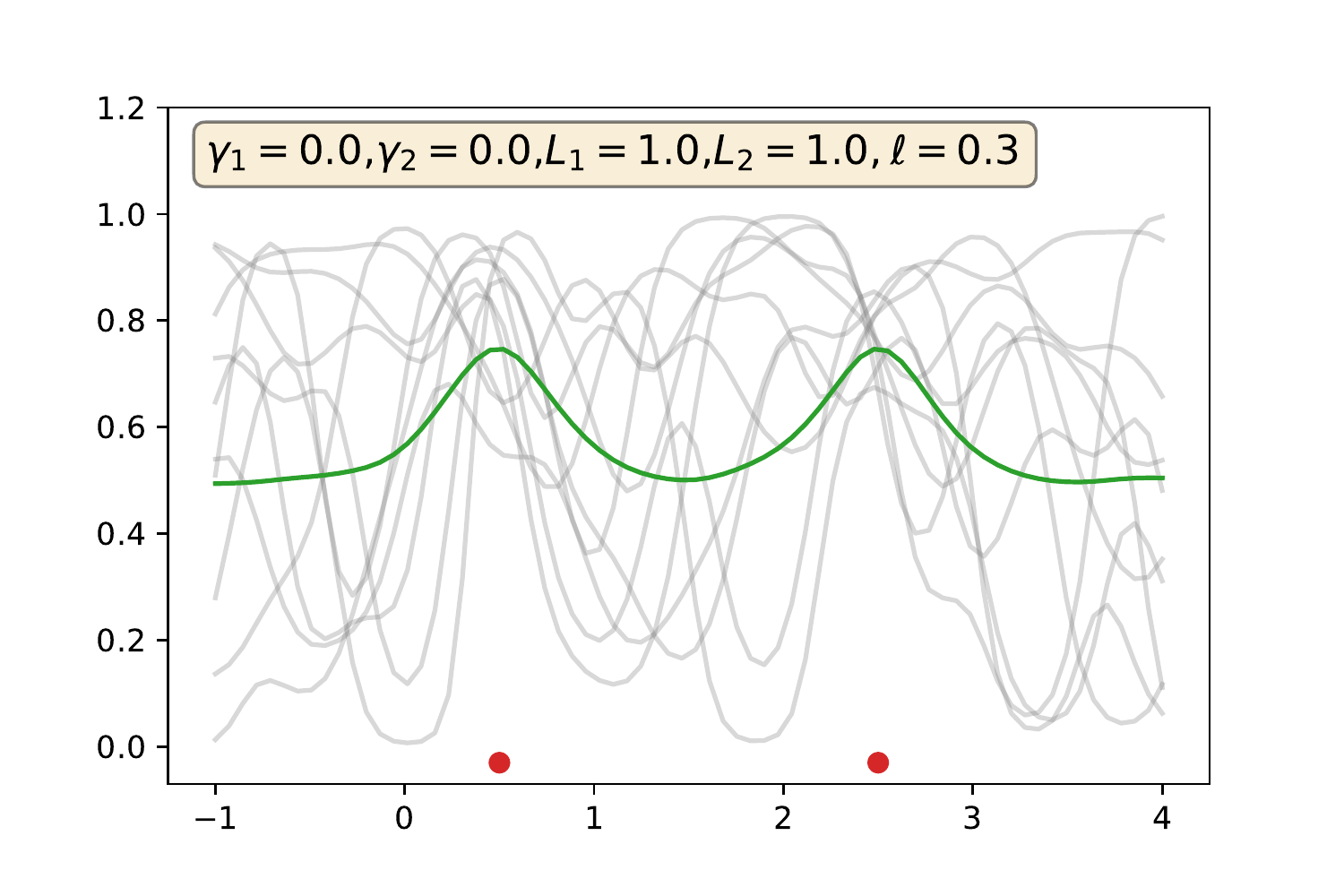} &
    \includegraphics[width=.48\linewidth]{{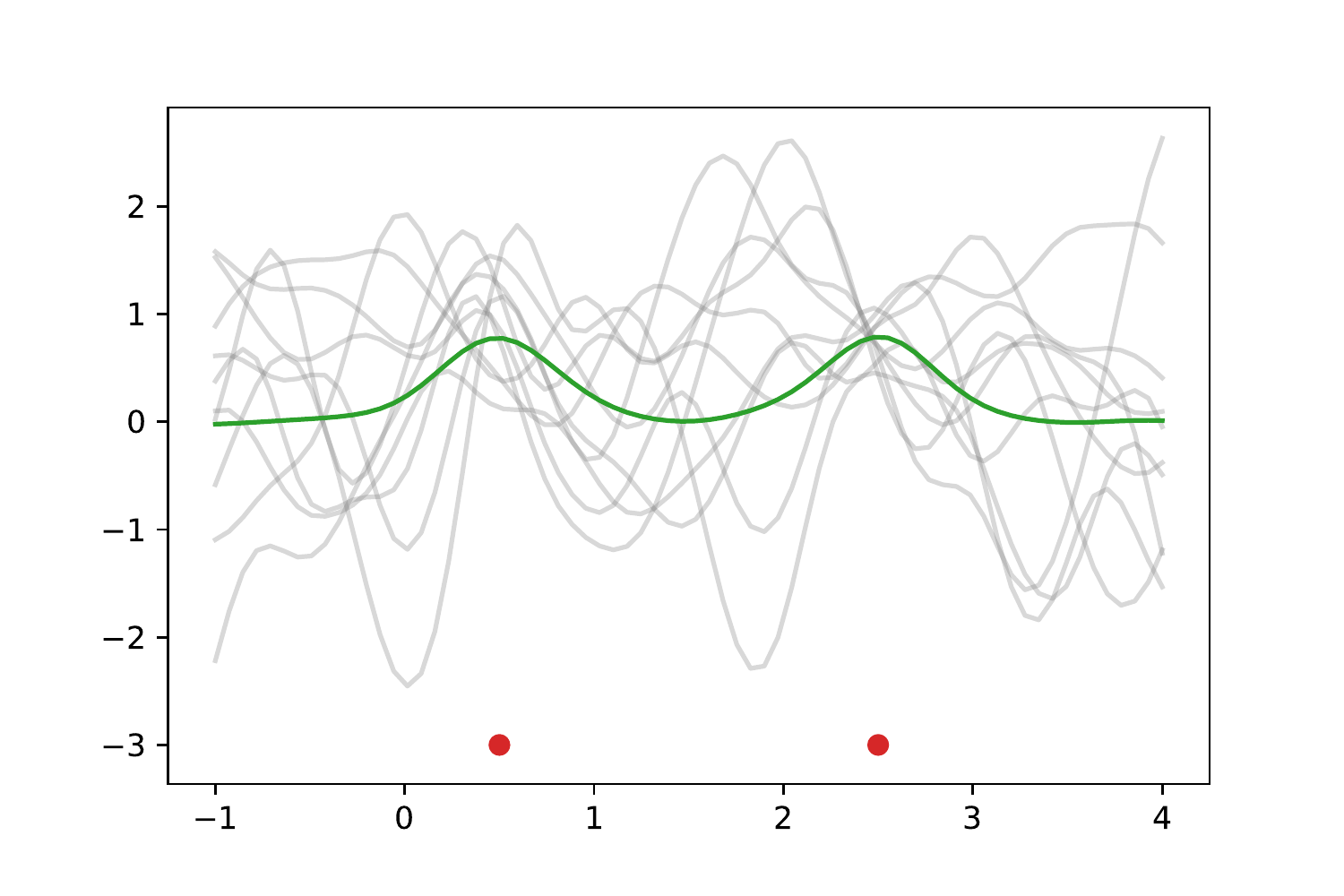}} \\
    \small (b1) & \small (b2)\\
        \includegraphics[width=.48\linewidth]{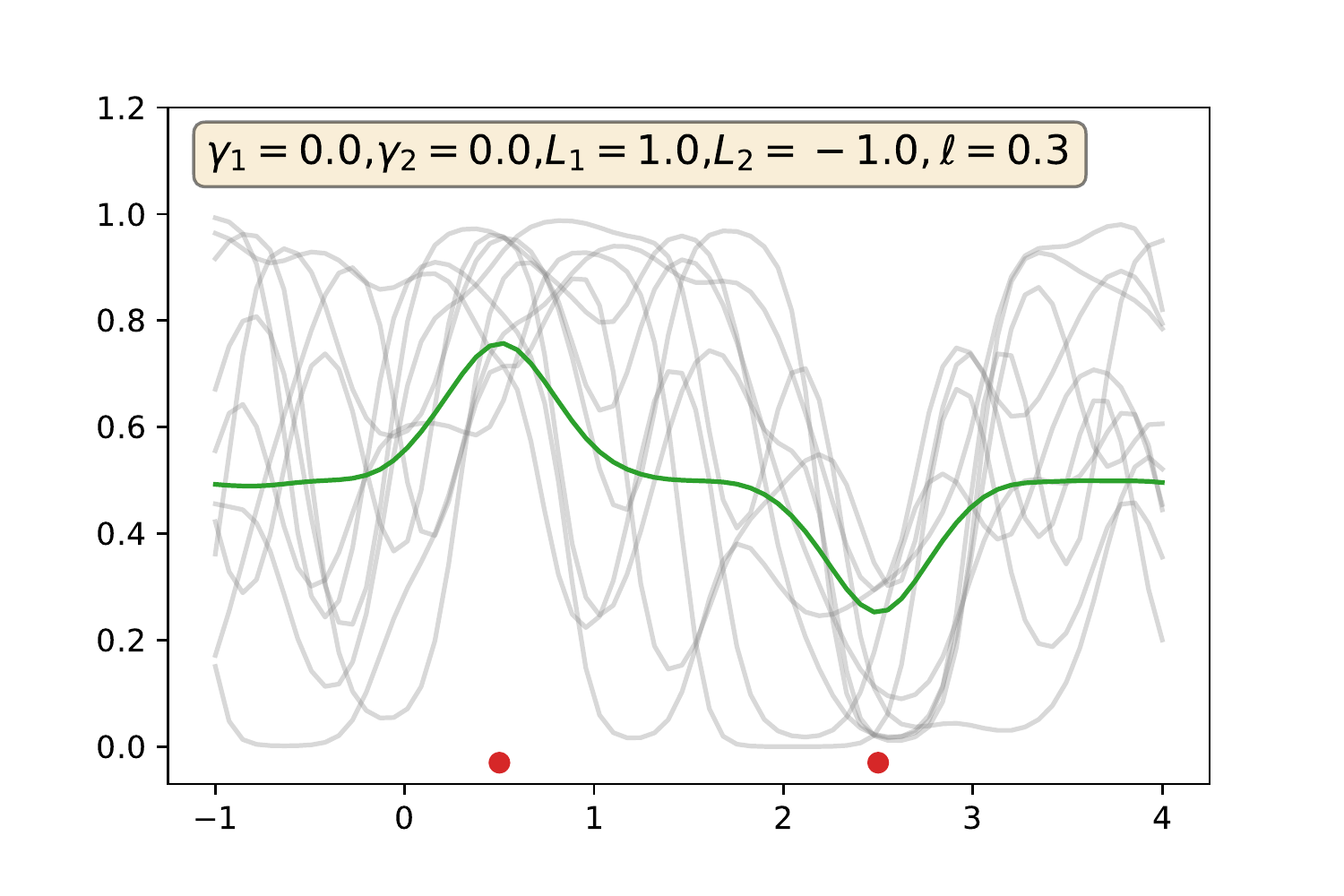} &
    \includegraphics[width=.48\linewidth]{{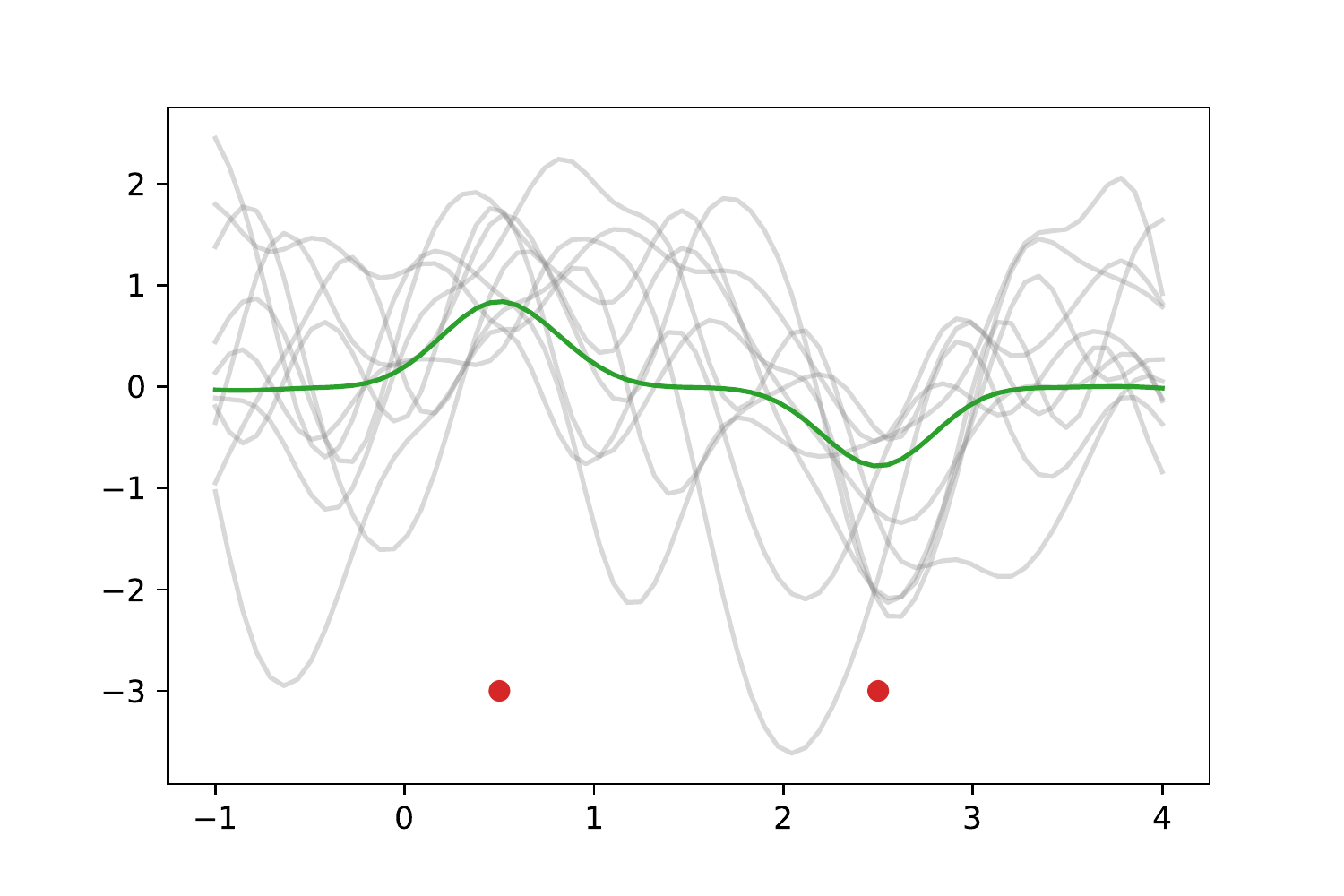}} \\
    \small (c1) & \small (c2)\\
        \includegraphics[width=.48\linewidth]{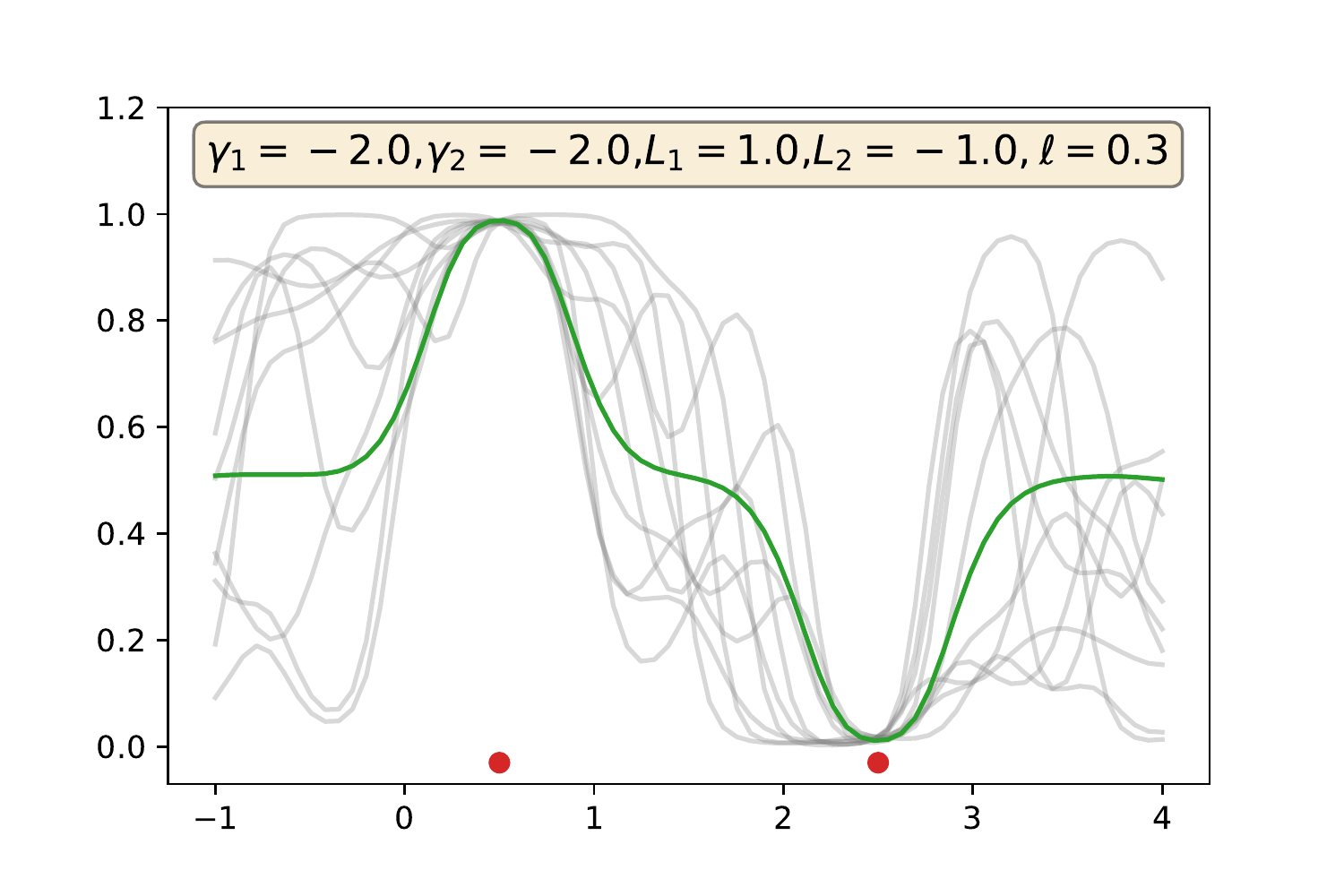} &
    \includegraphics[width=.48\linewidth]{{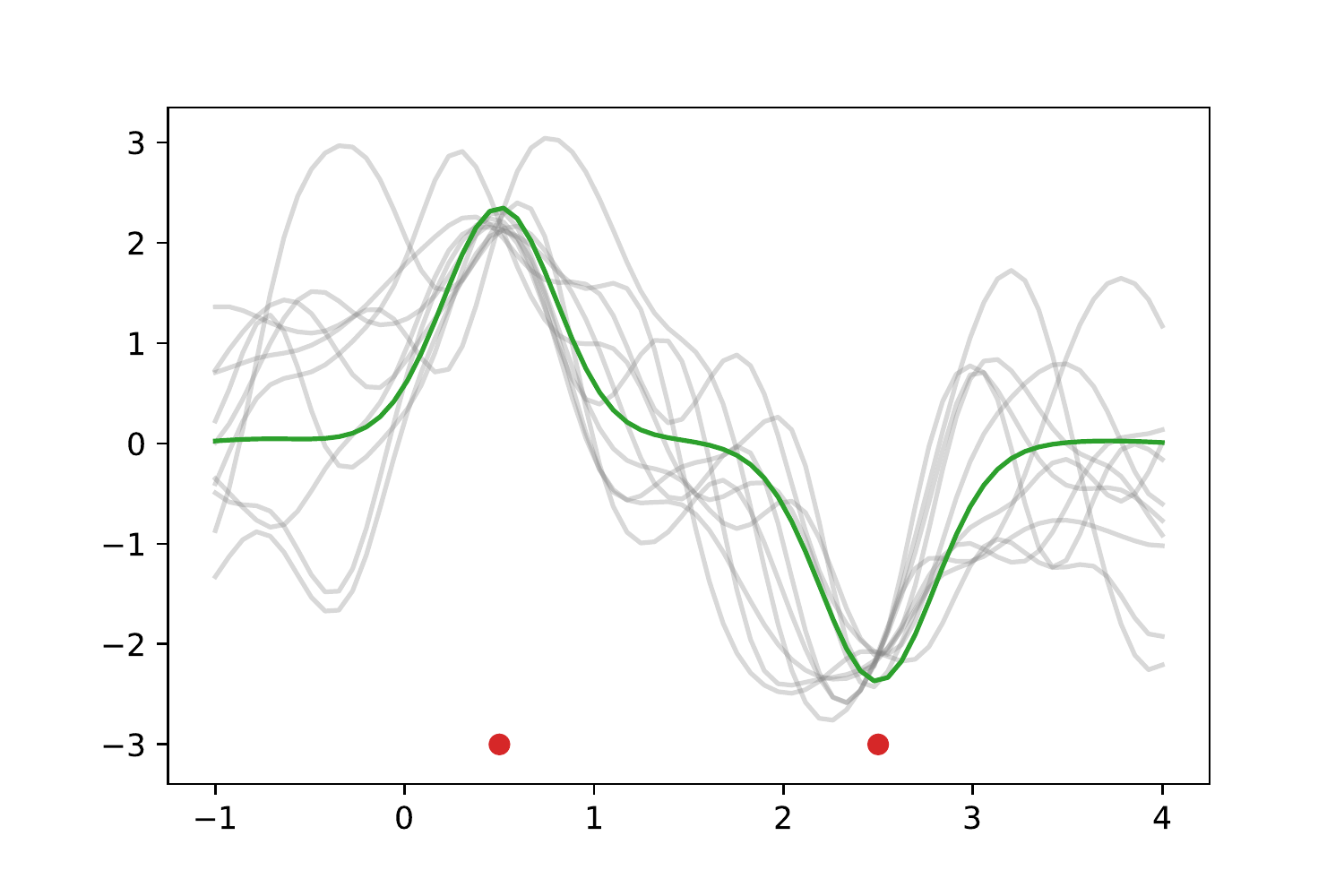}} \\
    \small (d1) & \small (d2)\\
        \includegraphics[width=.48\linewidth]{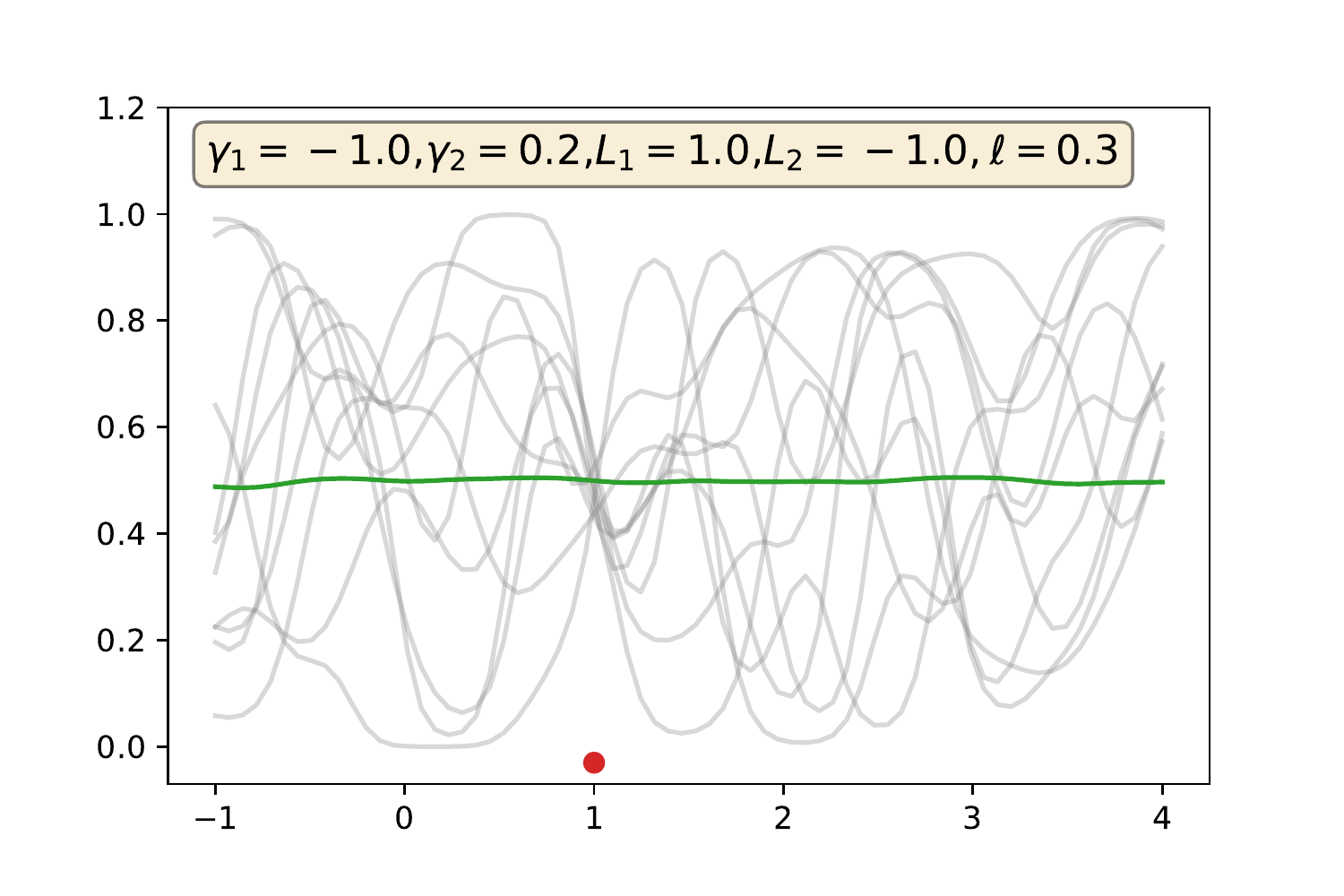} &
    \includegraphics[width=.48\linewidth]{{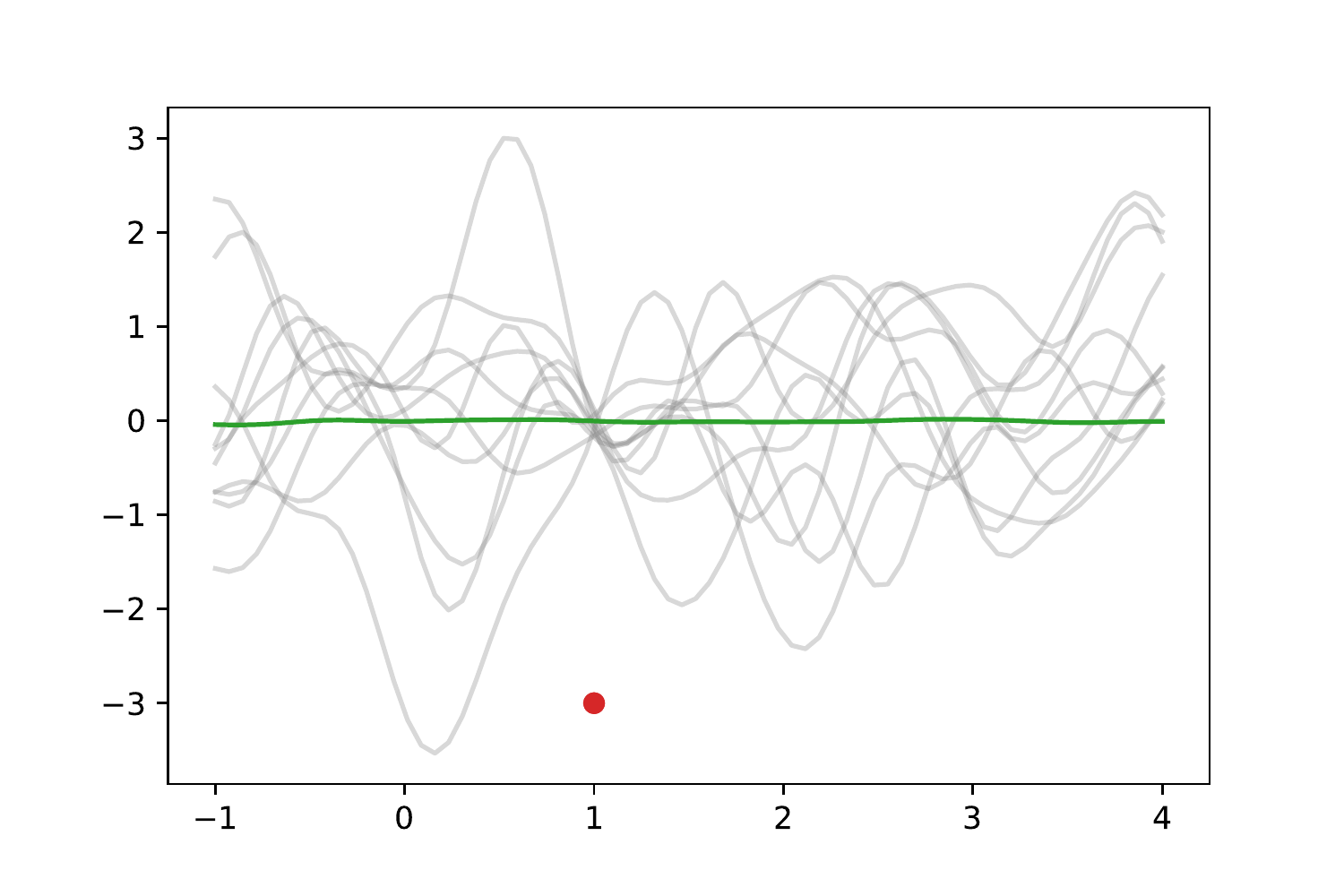}} \\
    \small (e1) & \small (e2)\\
  \end{tabular}
\caption{The second column shows random latent functions (in gray) drawn from a SkewGP with latent dimension $2$ and the first column shows the result of squashing these sample functions through the probit logistic function. The values
of the parameters of the SkewGP are reported at the top
of the plots in the first column. The green line is the  mean function and the red dots represent the location of the $s=2$ 
latent  pseudo-points.}
\label{fig:0}
\end{figure}

 Similarly to the inducing points in the sparse approximation of GPs \citep{quinonero2005unifying,snelson2006sparse,pmlrv5titsias09a,bauer2016understanding}, the  points $\bl[r]_i$  can be viewed as a set of $s$ latent variables. However, their role is totally different 
 from that of the inducing points, they allow us to locally modulate the skewness of SkewGP.
 Figure \ref{fig:0} shows  latent functions (in gray, second column) drawn from a SkewGP with latent dimension $2$ and  the result of squashing these sample functions through the probit logistic function (first column).  In all cases, we have considered  $\xi(\bx)=0$  and a  RBF kernel with $\ell=0.3$ and $\sigma^2=1$.  The values of the other parameters of the SkewGP are reported at the top
of the plots in the first column. The green line is the  mean function and the red dots represent the location of the $s=2$ 
latent  pseudo-points.
For large positive values of $\gamma_1, \gamma_2$, SkewGP is equivalent to a GP (plots (a1)-(a2)). At the decreasing  of $\gamma_i$, $i=1,2$ (plots (b1)-(b2)), the mean shifts up and the mass of the distribution is concentrated on the top of the figure.
By changing the phase (the sign of) $L_{22}$ (plots (c1)-(c2)),
the mean  and the mass of the distribution shift down 
at the location of the second pseudo-observation $r_2$.
We can magnifying this effect by decreasing both $\gamma_i$
(plots (d1)-(d2)). It is also possible to introduce skewness
without changing the mean (plots (e1)-(e2)). In this latter case, $r_1=r_2$ and the mass of the distribution is shifted up.

\section{Computational complexity}
\label{sec:compComplex}

Corollaries~\ref{co:ml} and \ref{co:predictive} provide two straightforward ways to compute the marginal likelihood and the predictive 
posterior probability however equations \eqref{eq:marginalLikelihood} and \eqref{eq:predPost} both require the accurate computation of $\Phi_{s+n}$. 
Quasi-randomized Monte Carlo methods \citep{genz1992numerical,genz2009computation,botev2017normal}  allows calculation of $\Phi_{s+n}$ for small $n$ (few hundreds observations).
Therefore, these procedures are not in general suitable 
for medium and large $n$ (apart from special cases
\citet{phinikettos2011fast,genton2018hierarchical,azzimonti2018estimating}). We overcome this issue with an effective use of sampling for the predictive posterior and a mini-batch approach to marginal likelihood. 

\paragraph{Posterior predictive distribution.} In order to compute the predictive distribution we generate samples from the posterior distribution at training points and then exploit the closure properties of the SUN distribution to obtain samples at test points. The following result from \citet{azzalini2013skew} allows us to draw independent samples from the posterior in \eqref{eq:posteriorclass}:
\begin{align}
\label{eq:sampling}
f(X) &\sim \tilde{\xi}+ \Domega\left(U_0+\tilde{\Delta}\tilde{\Gamma}^{-1}U_1\right),\\
\nonumber
U_0 &\sim \mathcal{N}(0;\bar{\Omega}-\tilde{\Delta}\tilde{\Gamma}^{-1}\tilde{\Delta}^T),\qquad 
U_1 \sim \mathcal{T}_{\tilde{\bgamma}}(0;\tilde{\Gamma}),
\end{align}
where $\mathcal{T}_{\tilde{\bgamma}}(0;\tilde{\Gamma})$ is the pdf of a multivariate Gaussian distribution truncated component-wise below $-\tilde{\bgamma}$. { Equation~\eqref{eq:sampling} is a consequence of the additive representation of skew-normal distributions, see \citep[Ch. 7.1.2 and 5.1.3]{azzalini2013skew} for more details. Note that sampling $U_0$ can be achieved with standard methods, however using standard rejection sampling for the variable $U_1$ would incur in exponentially growing sampling time as the dimension increases.
A commonly used sampling technique for this type of problems is Elliptical Slice Sampling (\emph{ess}) \citep{pmlrv9murray10a} which is a Markov Chain Monte Carlo algorithm that performs inference in models with Gaussian priors. This method looks for acceptable samples along elliptical slices and by doing so drastically reduces the number of rejected samples. Recently, \citet{gessner2019integrals} proposed an extension of \emph{ess}, called linear elliptical slice sampling (\textit{lin-ess}), for multivariate Gaussian distribution truncated on a region defined by linear constraints. In particular, this approach analytically derives the acceptable regions on the elliptical slices used in \emph{ess} and thus guarantees rejection-free sampling. This leads to a large speed up over \emph{ess}, especially in high dimensions. }


Given posterior samples at the training points it is possible to compute the predictive posterior at a new test points $\bx^*$ thanks to the following result. 

\begin{theorem}
	\label{th:predictive}
	The posterior samples of the latent function computed at the test point $\bx^*$ can be obtained by
	sampling $ f(\bx^*)$ from:
	\begin{align}
	\nonumber
	f(\bx^*) &\sim SUN_{1,s}({\xi}_{*},{\Omega}_{*},{\Delta}_{*},{\bgamma}_{*},{\Gamma}_{*}),\\
	\nonumber
	{\xi}_{*}  &=\xi(\bx^*)+K(\bx^*,X)K(X,X)^{-1}(f(X)-\xi(X)),\\
	\nonumber
	{\Omega}_{*} &= K(\bx^*,\bx^*)-K(\bx^*,X)K(X,X)^{-1}K(X,\bx^*),\\
	\nonumber
	{\Delta_{*}} &=\Delta(\bx^*) -\bar{K}(\bx^*,X)\bar{K}(X,X)^{-1}\Delta(X),\\
	\nonumber
	{\bgamma_{*}}& =\bgamma+\Delta(X)^T \bar{K}(X,X)^{-1}\Domega(X,X)^{-1}(f(X)-\xi(X)), \\
	\nonumber
	{\Gamma_{*}}&=\Gamma-\Delta(X)^T\bar{K}^{-1}\Delta(X),
	\end{align} 
	with $f(X)$ sampled from the posterior $\text{SUN}_{n,s+n}(\tilde{\xi},\tilde{\Omega},\tilde{\Delta},\tilde{\bgamma},\tilde{\Gamma})$ in Theorem \ref{th:1}, $K$ is a kernel that defines the matrices $\Gamma,\Delta, \Omega$ as in eq.~\eqref{eq:positivityKernel} and where 
	$$
	\begin{bmatrix}
	\bar{K}(X,X) & \bar{K}(X,\bx^*)\\
	\bar{K}(\bx^*,X) & \bar{K}(\bx^*,\bx^*)
	\end{bmatrix} :=\Domega^{-1}\begin{bmatrix}
	K(X,X) & K(X,\bx^*)\\
	K(\bx^*,X) & K(\bx^*,\bx^*)
	\end{bmatrix}\Domega^{-1}
	$$
	and $\Domega=\text{diag}[\Domega(X),\Domega(\bx^*)])$ is a diagonal matrix  containing the square root of the diagonal elements of the inner matrix in the r.h.s..
\end{theorem}

Observe that the computation of the predictive posterior requires the inversion of a $n \times n$ matrix ($\bar{\Omega}_{11}$),
which has complexity $O(n^3)$ with storage demands of $O(n^2)$.
This means we have similar bottleneck computational complexity
of GPs. Moreover, note that sampling from $SUN_{1,s}$
is extremely efficient when the latent dimension $s$ is small
(in the experiments we use $s=2$).

\paragraph{Marginal likelihood.} As discussed in the previous section, in  practical application of SkewGP, the (hyper-)parameters of the scale function $\Omega(\bx,\bx')$, of the skewness vector function $\Delta(\bx,\bx') \in \mathbb{R}^s$  and the parameters $\bgamma \in \mathbb{R}^s,\Gamma \in \mathbb{R}^{s \times s}$  have to be selected. As for GPs, we use Bayesian model selection to consistently set such parameters. This requires the maximization of the marginal likelihood with respect to these parameters and, therefore, it is essential to provide a fast and accurate way to evaluate the marginal likelihood. 
%
In this paper, we propose a simple approximation  of the marginal likelihood that allows us to efficiently compute a lower bound.
\begin{proposition}
	\label{prop:frechet}
	Consider the marginal likelihood $p(\mathcal{D}|\mathcal{M})$ in Corollary \ref{co:ml}, then it holds
	\begin{equation}
	\label{eq:lowerbound}
	p(\mathcal{D}|\mathcal{M}) = \frac{\Phi_{s+n}(\tilde{\bgamma};~\tilde{\Gamma})}{\Phi_{s}(\bgamma;~\Gamma)}\geq\frac{  \sum_{i=1}^{b} \Phi_{s+|B_i|}(\tilde{\bgamma}_{B_i};~\tilde{\Gamma}_{B_i})-(b-1)}{\Phi_{s}(\bgamma;~\Gamma)},
	\end{equation}
	where $B_1,\dots,B_b$ denotes a partition of the training dataset  into $b$ random disjoint subsets, $|B_i|$ denotes the number of observations  in the i-th element of the partition, 
	$\tilde{\bgamma}_{B_i},~\tilde{\Gamma}_{B_i}$ are the parameters
	of the posterior computed using only the subset $B_i$ of the data.
\end{proposition}
If the batch-size is low (in the experiments we have used $|B_i|=30$), then we can efficiently compute each term $\Phi_{s+|B_i|}(\tilde{\bgamma}_{B_i};~\tilde{\Gamma}_{B_i})$ by using a quasi-randomised Monte Carlo method. We can then optimize the hyper-parameter of SkewGP by maximizing the lower bound in \eqref{eq:lowerbound}.

\begin{figure}[h]
	\centering
	\includegraphics[width=6cm]{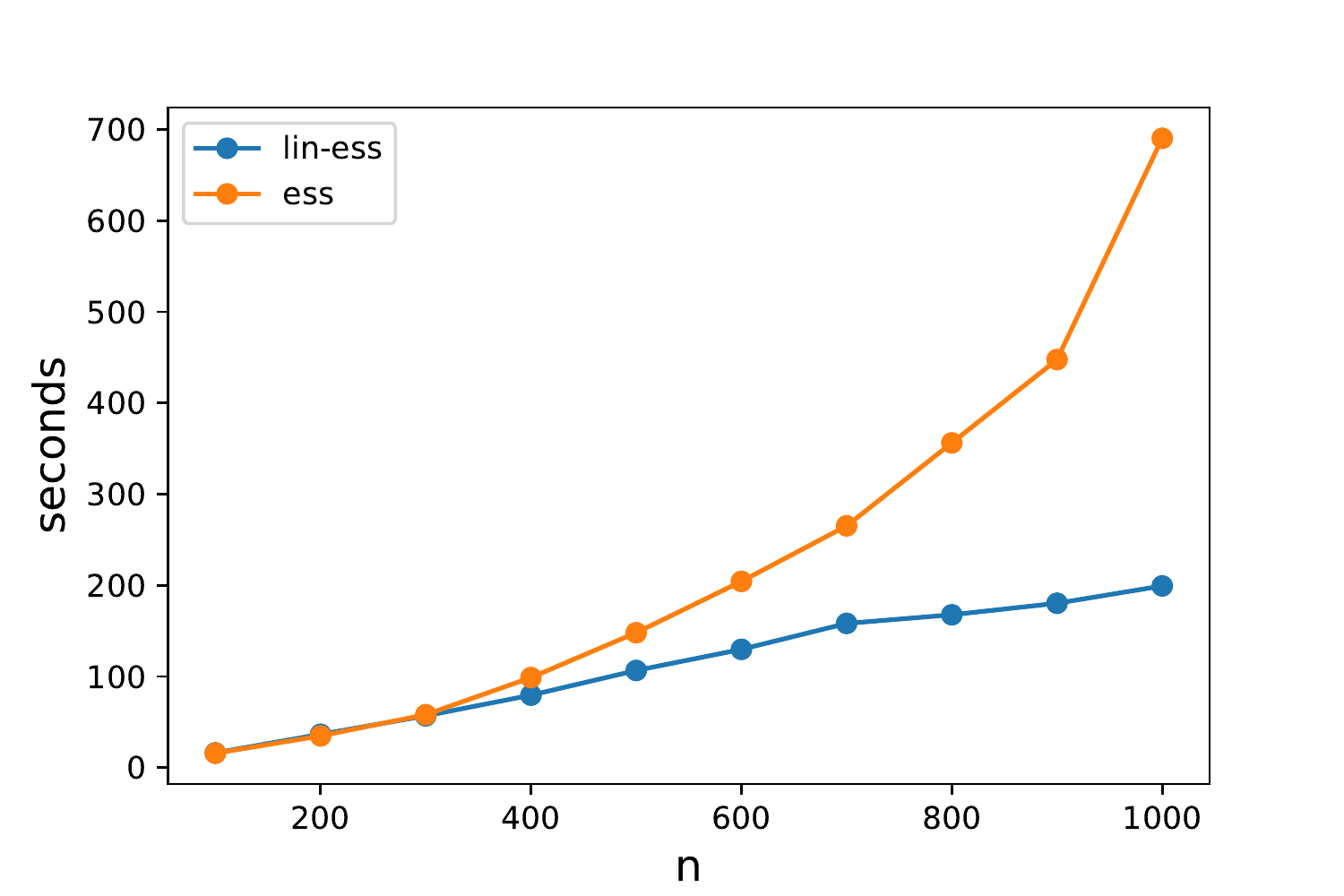}
	\caption{Computational time for sampling from the posterior
		of SkewGP0 using ess vs.\ lin-ess on a standard laptop.}
	\label{fig:computime}
\end{figure}

\paragraph{Computational load and parallelization:}
To evaluate the computational load, we have generated artificial classification data using a probit likelihood model and drawing 
the latent function $f(X)=[f(\bx_1),\dots,f(\bx_n)]$, with $\bx_i \sim N(0,1)$, from a GP with zero mean  and radial basis function kernel (lengthscale $0.5$
and variance $2$). We have then computed the full posterior
latent function from Theorem \ref{th:1}, that is a  SkewGP posterior. Figure \ref{fig:computime} compares the CPU time for sampling $1000$ instances of $f(X)$ from the SkewGP posterior  as a function of $n$ for lin-ess  vs.\ a standard Elliptical Slice Sampler (ess) ($5000$ burn in).\footnote{
	Sampling is performed according to \eqref{eq:sampling}
	and, therefore, lin-ess and ess are  applied to sample $U_1$ from $\mathcal{T}_{\tilde{\bgamma}}(0;\tilde{\Gamma})$. To increase the probability of acceptance for ess, we have replaced the indicator function that defines the truncation $I_{u_1>\tilde{\bgamma}}$ with a logistic function 
	$\text{sigmoid}(80(u_1-\tilde{\bgamma}))$. We have  verified that, using $5000$ samples for burn in, the posterior first moment of lin-ess and ess are close for all considered values of $n$.} It can be observed the computational advantage of lin-ess with respect to ess.

%

The average CPU time required to compute $\Phi_{s+|B_i|}(\tilde{\bgamma}_{B_i};~\tilde{\Gamma}_{B_i})$ with $|B_i|=30$,
using the randomized lattice routine with $5000$ points
\citep{genz1992numerical},  is $0.5$ seconds on a standard laptop.
Since the above method is randomized, we use the \textit{simultaneous perturbation stochastic approximation}
algorithm \citep{spall1998implementation} for optimizing the maximum lower bound \eqref{eq:lowerbound}.\footnote{The SkewGP classifier is implemented in Python, but we  call  Matlab to use existing implementations of both the simultaneous perturbation stochastic approximation algorithms and the randomized lattice routine. We plan to re-implement everything in Python and then release the source code for SkewGP.}

Finally notice that, both the computation of the lower bound of the marginal likelihood and the sampling from the posterior via lin-ess are highly parallelizable. In fact, each
term $\Phi_{s+|B_i|}(\tilde{\bgamma}_{B_i};~\tilde{\Gamma}_{B_i})$ can be computed independently as well as
each sample in lin-ess can be sample independently
(because lin-ess is rejection-free and, therefore,
no  ``burn in'' is necessary).

 
 \section{Properties of the Posterior}
 In the above sections, we have shown how to compute 
the posterior distribution of a SkewGP process when the likelihood
is a probit model. The full conjugacy of the model 
allows us to prove that the posterior is again a SkewGP process.
This section provides more details on the
properties of the posterior and compares it with two  approximations.
For GP classification, there are two main alternative approximation schemes for finding a Gaussian approximation to the posterior: the Laplace's method and the Expectation-Propagation (EP) method, see, e.g. \citet{rasmussen2006gaussian} chapter 3.

\begin{figure}[h!]
\centering
  \begin{tabular}{c @{\qquad} c }
    \includegraphics[width=.48\linewidth]{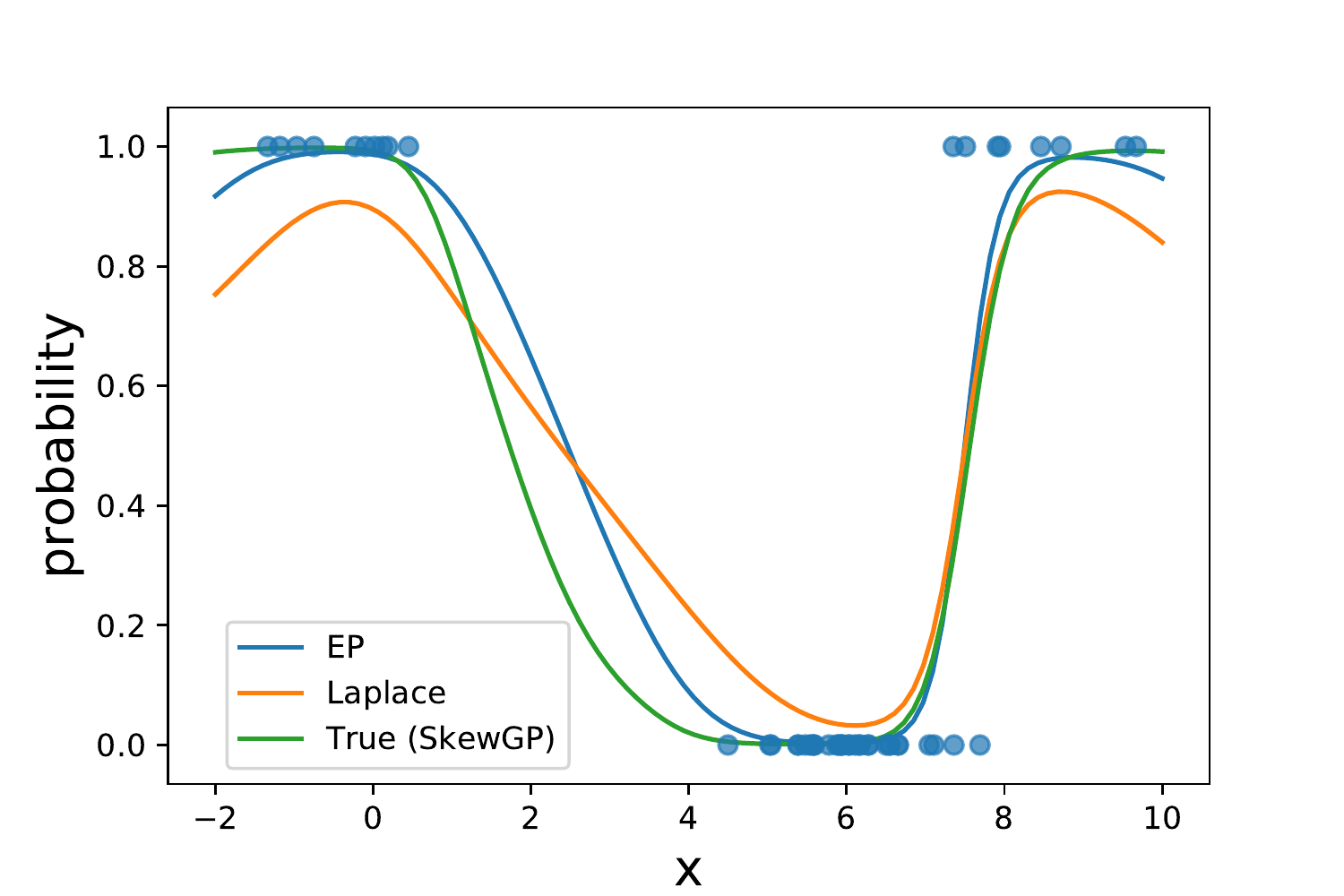} &
    \includegraphics[width=.48\linewidth]{{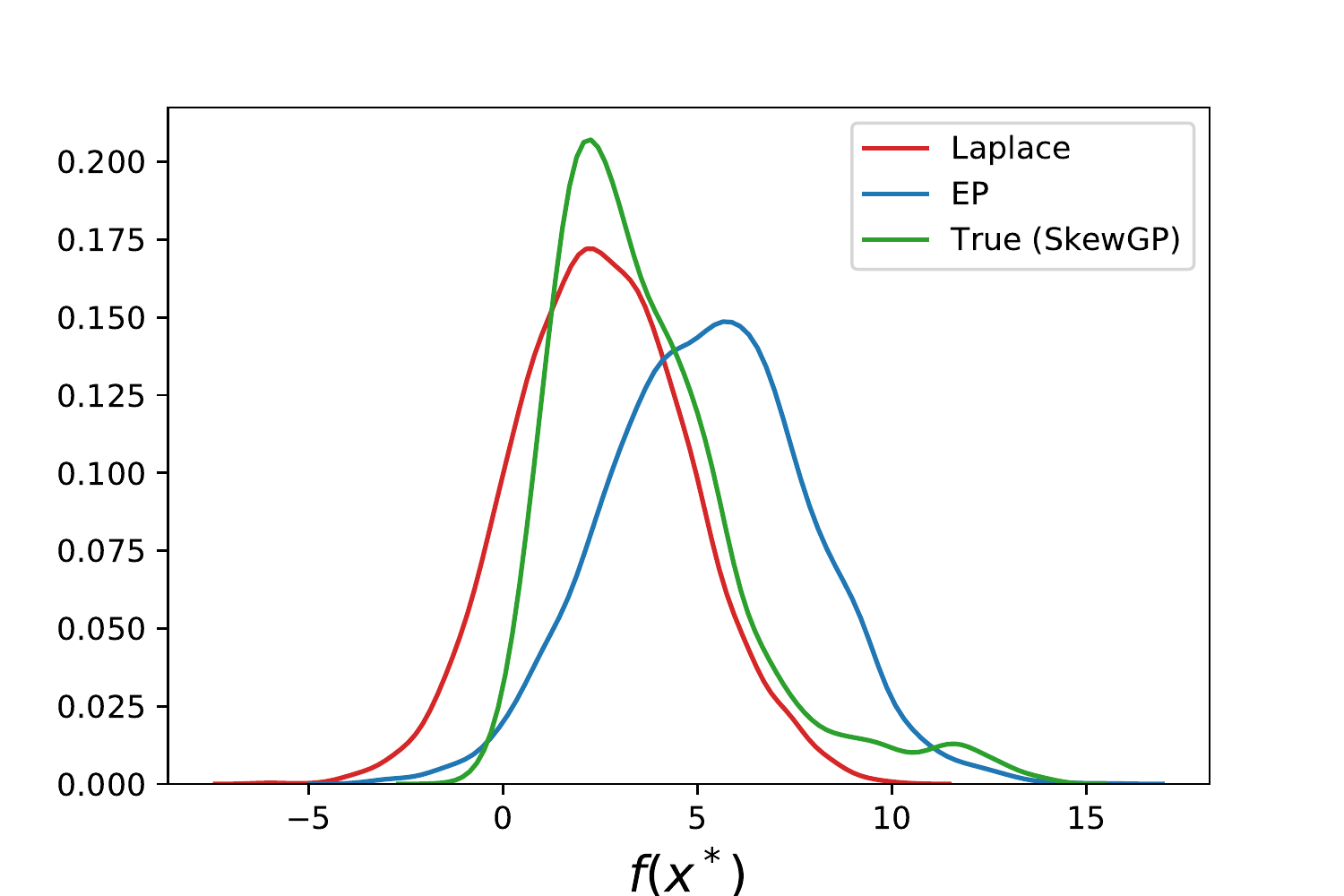}} \\
    \small (a) & \small (b)\\
   \end{tabular} 
  \includegraphics[width=\linewidth]{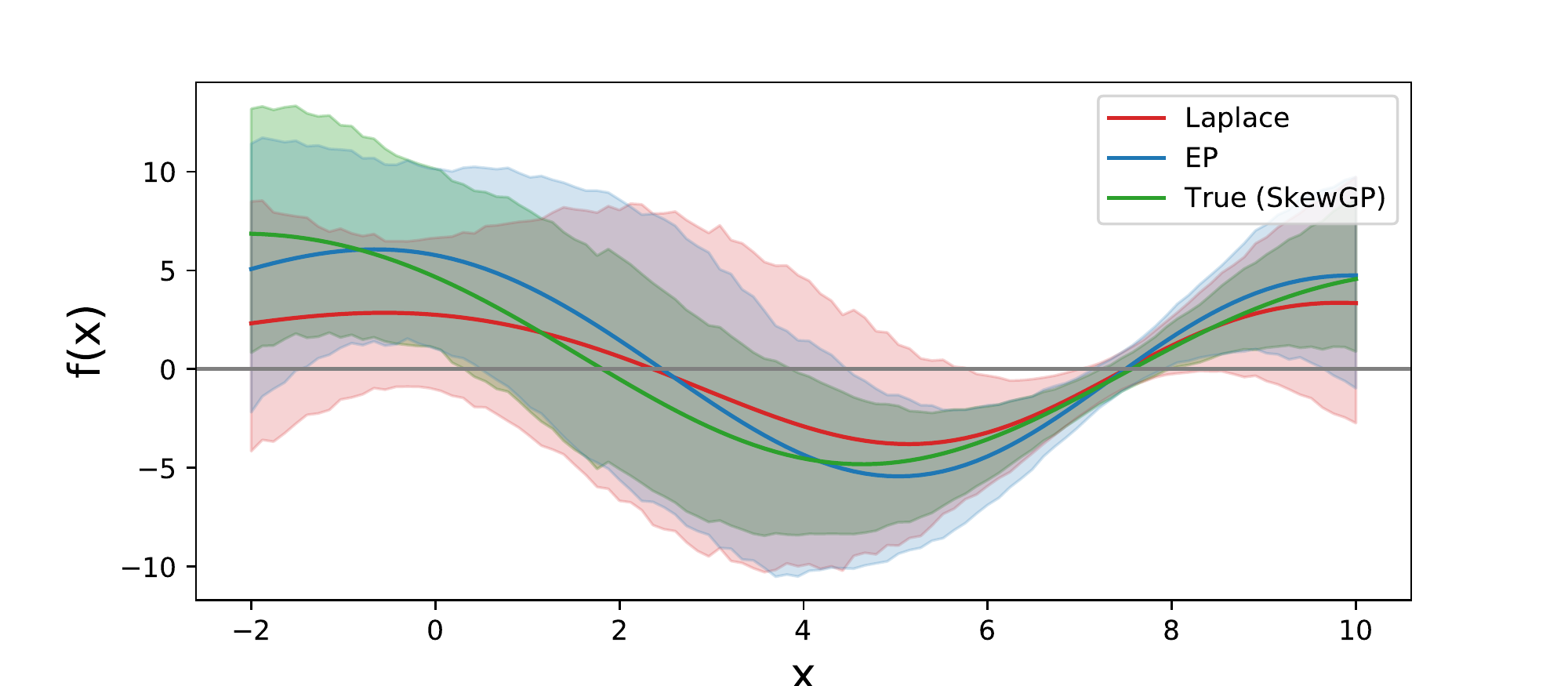}\\
  \small (c)\\
\caption{Plot (a) shows the training dataset and compares the true  posterior mean probability  of class 1, that is a SkewGP, versus
the approximations computed via Laplace's method and EP.
Plot (c) shows the posterior mean latent function and corresponding 95\% credible intervals for the three methods and Plot (b) reports
the density-plot of the latent function posterior prediction at $x^*=0.42$.}
\label{fig:1}
\end{figure}

Figure \ref{fig:1} provides a one-dimensional illustration using
a  synthetic classification problem with 50 observations and scalar inputs taken from
\citep{kuss2005assessing}.
Figure  \ref{fig:1}(a) shows the dataset and the predictive posterior probability for the Laplace and EP approximations.
Moreover, by using a SkewGP prior with latent dimension $s=0$
(that coincides with a GP prior), we have computed the exact SkewGP predictive posterior probability.
Therefore, all  three methods  have the same prior: a GP with zero mean and RBF covariance function (the lengthscale and variance of the kernel are the same for all the three methods and have been set equal to the values that maximise the Laplace's approximation to the marginal likelihood).
Figure \ref{fig:1}(c) shows the posterior mean latent function and corresponding 95\% credible intervals.
It is evident that the true posterior (SkewGP) of the latent function is skew (see for instance for $x\in [0,2]$
and the slice plot in Figure \ref{fig:1}(b)).
Laplace's approximation peaks at the posterior mode, but places  too much mass over positive values of
$f$. The EP approximation aims to match the first two moments
of the posterior and, therefore, usually obtains a better coverage
of the posterior mass. That is why EP is usually the method of choice for approximate inference in GP classification.

\begin{figure}
\centering
  \begin{tabular}{c @{\qquad} c }
    \includegraphics[width=.48\linewidth]{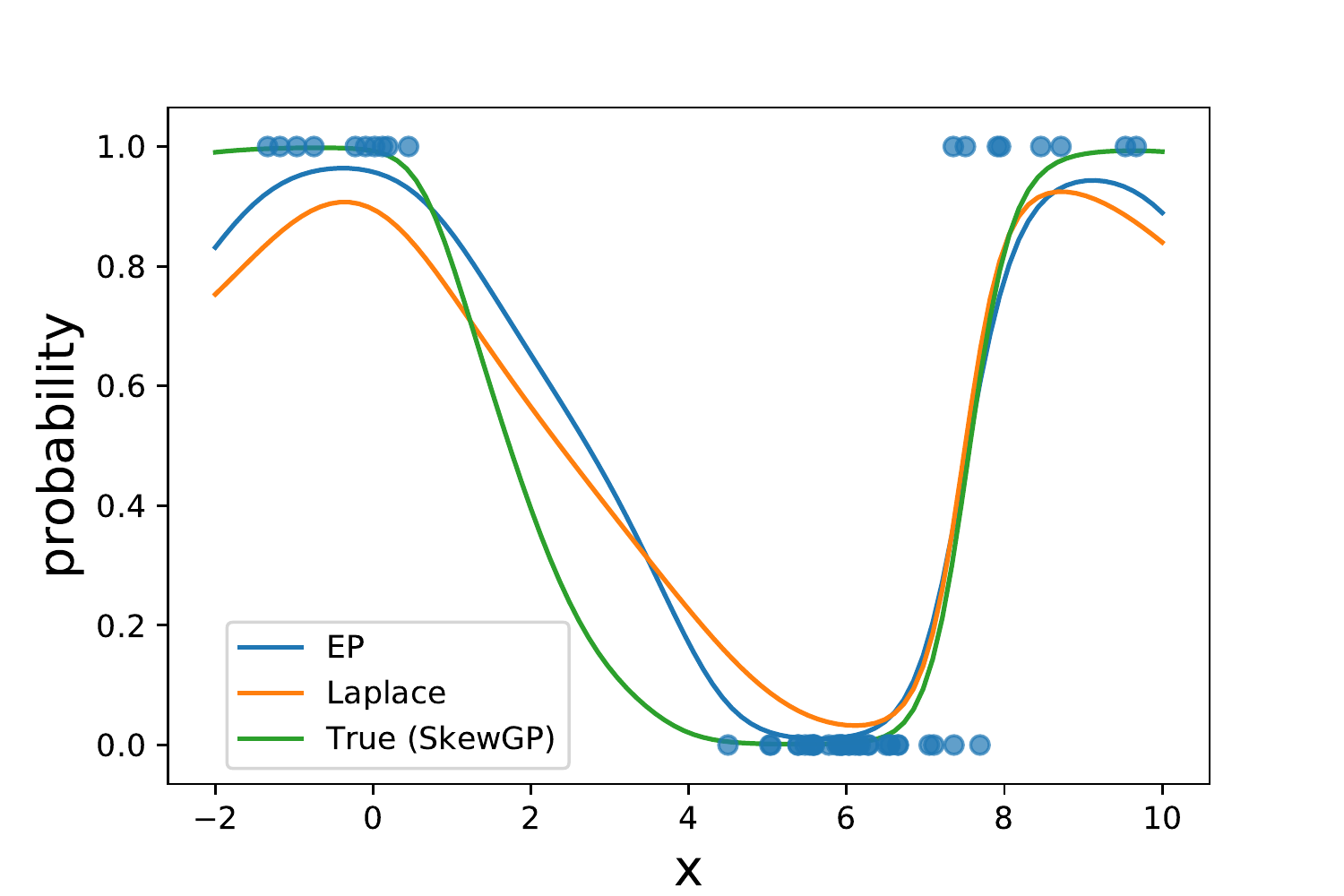} &
    \includegraphics[width=.48\linewidth]{{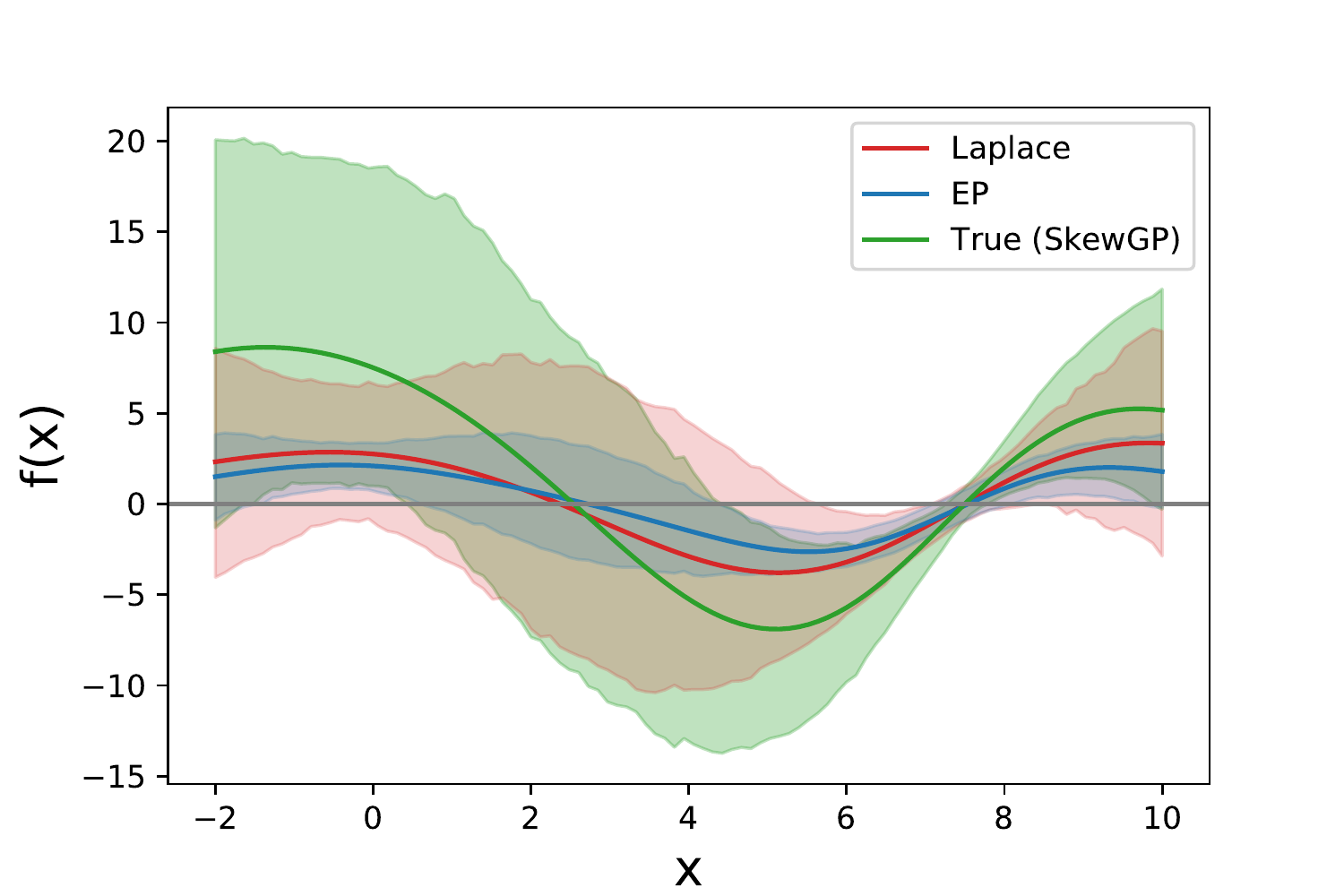}} \\
    \small (a) & \small (b)\\
   \end{tabular}
\caption{Plot (a) shows the training dataset and compares the true  posterior mean probability  of class 1, that is a SkewGP, versus
the approximations computed via Laplace's method and EP.
Plot (b) shows the posterior mean latent function and corresponding 95\% credible intervals for the three methods.}
\label{fig:2}
\end{figure}

Figure \ref{fig:2} shows the true posterior and the two approximations for the same dataset, but now the lengthscale and variance of the kernel are the optimal values  for  the three methods. It is evident that the skewness of the posterior
provides a better model fit to the data.

\begin{figure}
\centering
  \begin{tabular}{c @{\qquad} c }
    \includegraphics[width=.48\linewidth]{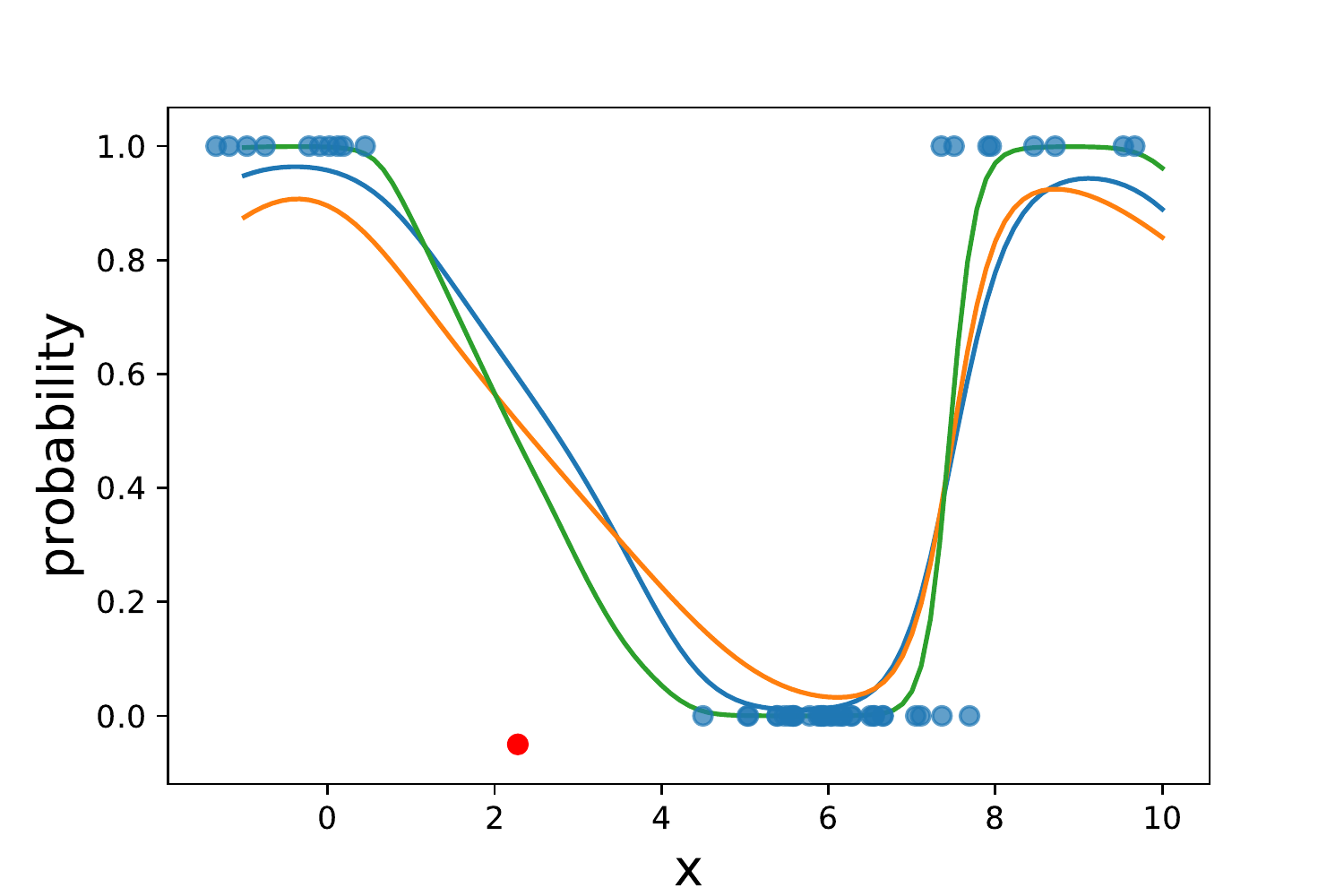} &
    \includegraphics[width=.48\linewidth]{{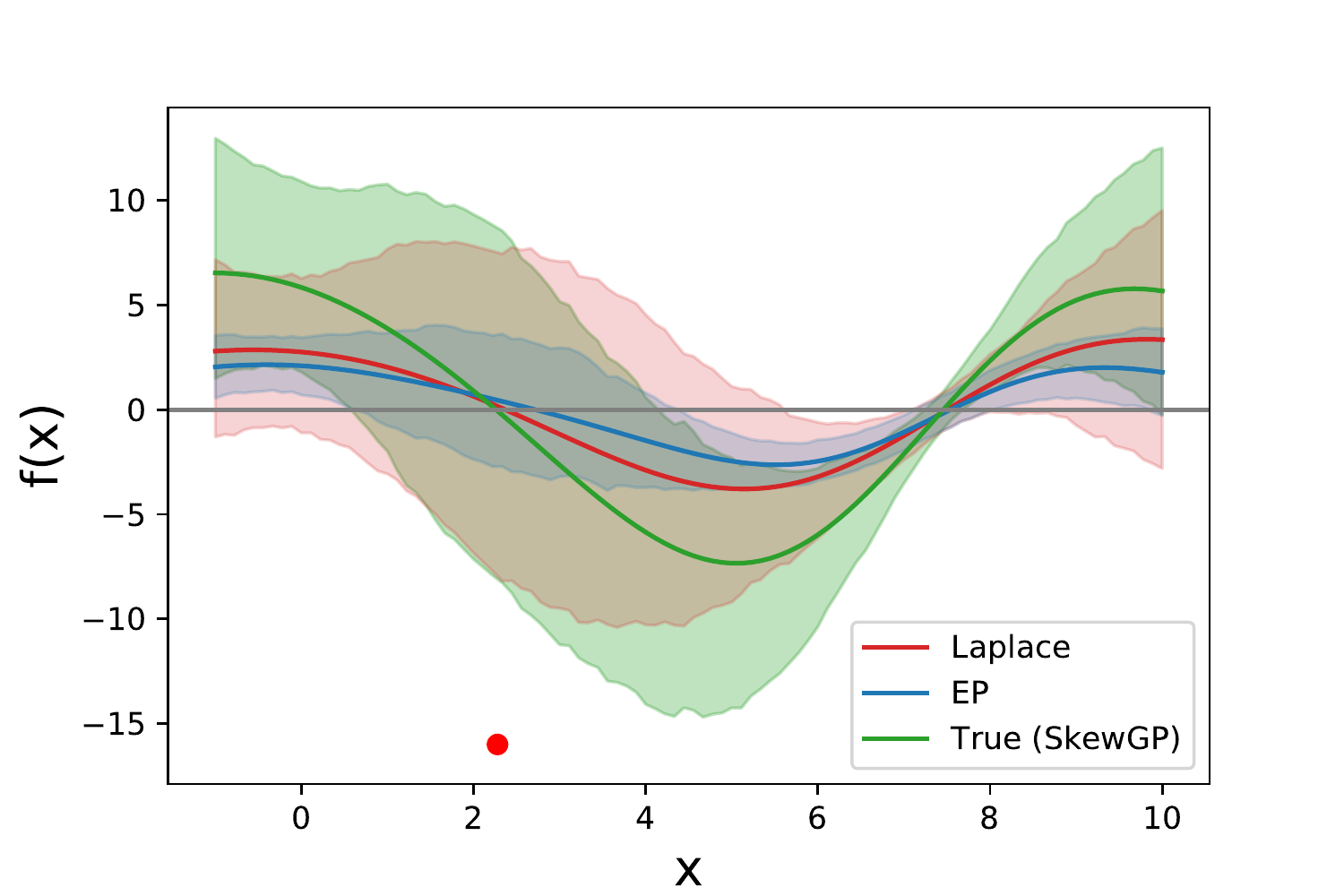}} \\
    \small (a) & \small (b)\\
   \end{tabular}
\caption{Plot (a) shows the training dataset and compares the true  posterior mean probability  of class 1, that is a SkewGP2, versus
the approximations computed via Laplace's method and EP.
Plot (b) shows the posterior mean latent function and corresponding 95\% credible intervals for the three methods.
The red dot denotes  the optimal location of the pseudo-point
$r_1$.}
\label{fig:3}
\end{figure}

Figure \ref{fig:3} shows the posteriors  corresponding to a prior SkewGP process with latent-dimension $s=2$. The red dot denotes  the optimal location of the pseudo-point
$r_1$, while $r_2=13.5$ (their initial location were $5.8$ and, respectively, $6$).
The additional degrees of freedom of the SkewGP prior process  gave a much more satisfactory answer than that obtained from a GP prior model. By comparing Figures \ref{fig:2} and \ref{fig:3}, it can be noticed that the skew-point allows us to locally modulate the skewness.
Moreover, the additional degrees of freedom do not lead to  overfitting, even with small data, as highlighted
by the  optimized location of $r_2$ (far away) that has not effect on the skewness of the posterior SkewGP. 
%

\section{Results}
We have evaluated the  proposed SkewGP classifier  on a number of benchmark  classification datasets and compared its
classification accuracy  with the
accuracy of a Gaussian process classifier that uses either
Laplace's method (GP-L) or Expectation Propagation (GP-EP) for approximate Bayesian inference.
For GP-L and GP-EP, we have used the
implementation available in GPy \citep{gpy2014}.

\subsection{Penn Machine Learning Benchmarks}
\begin{figure}
\centering
  \includegraphics[width=12cm,trim={0 0.8cm 0 0},clip]{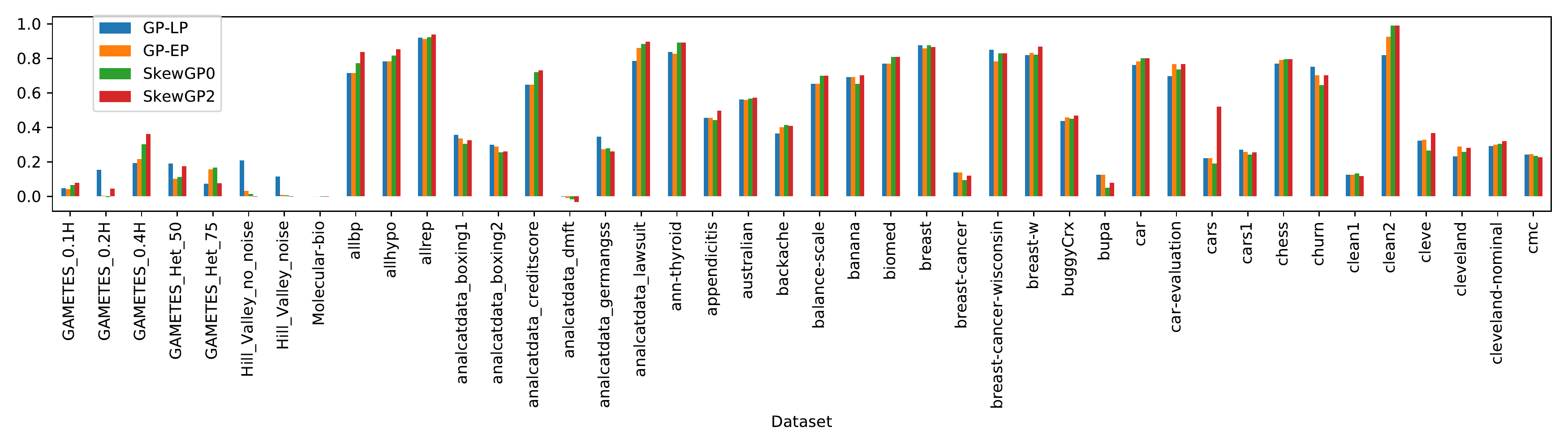}
  \includegraphics[width=12cm,trim={0 0.8cm 0 0},clip]{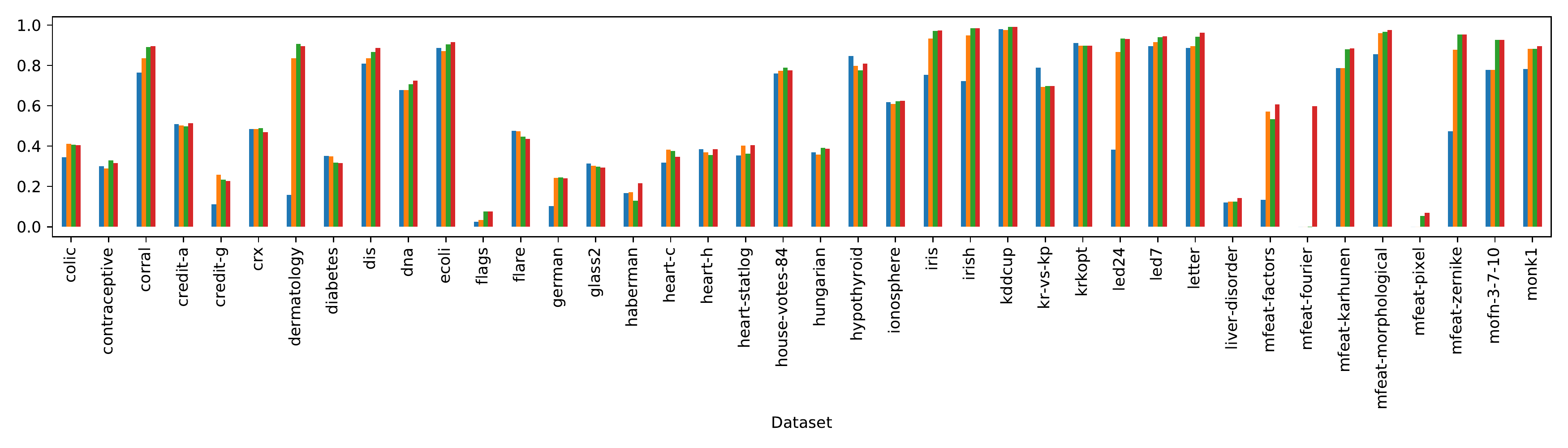}
    \includegraphics[width=12cm, height=2.5cm, trim={0 0.8cm 0 0},clip]{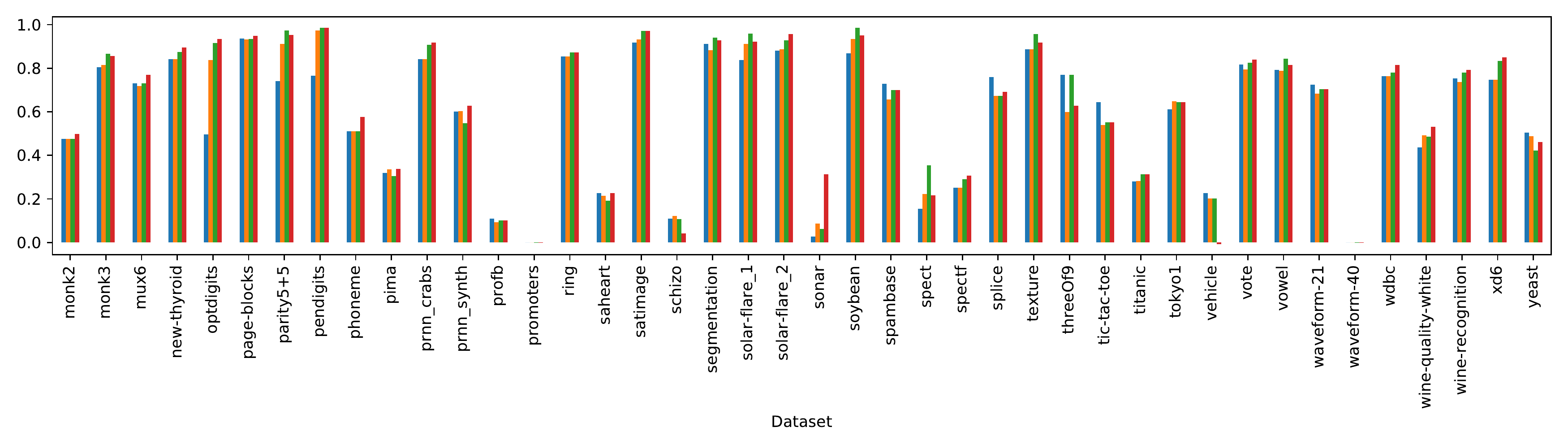}
\caption{Average information score on 124 datasets of the Penn Machnine Learning Benchmark dataset.}
\label{fig:penn}
\end{figure}

\begin{figure}
\centering
  \includegraphics[width=5.5cm]{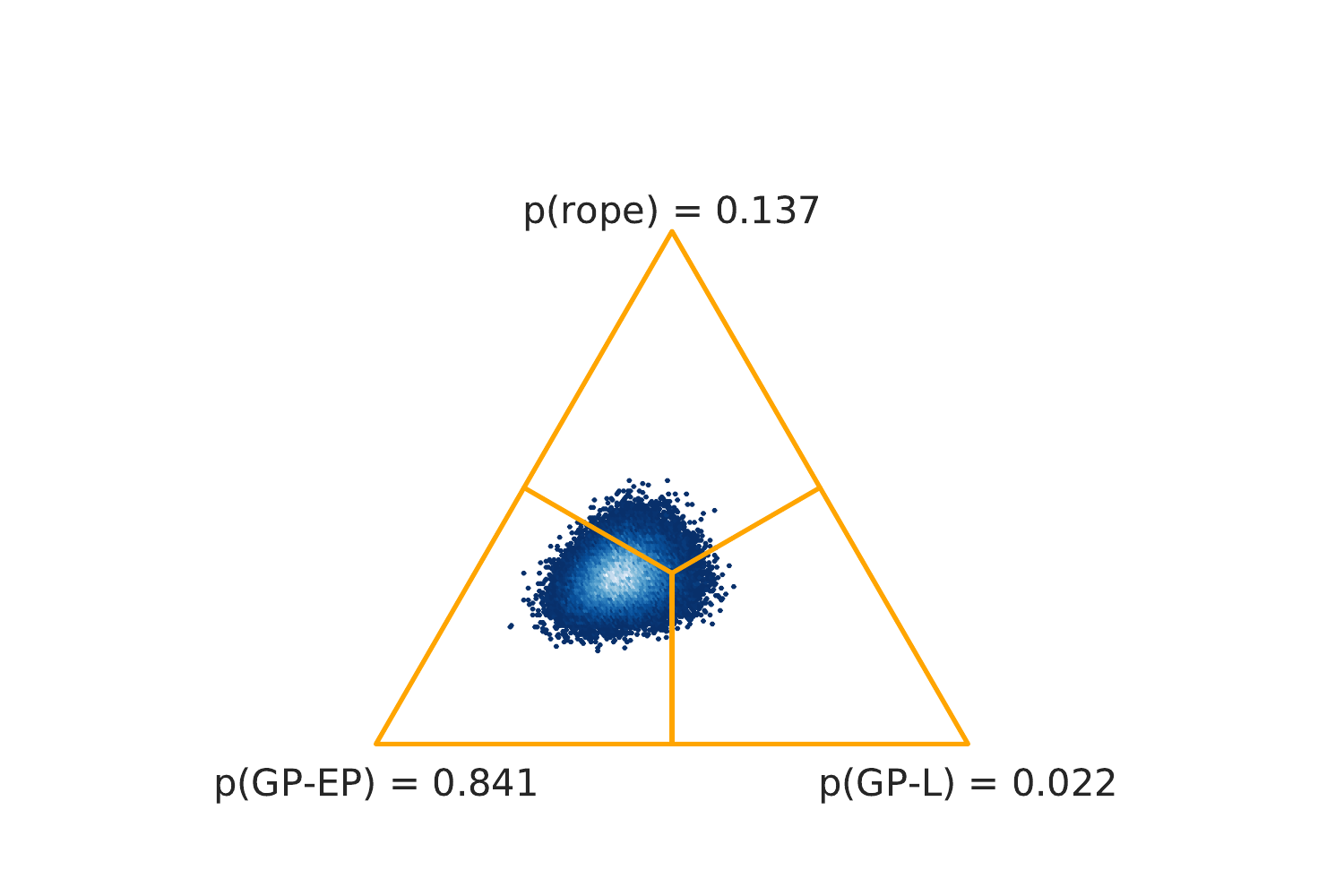}
  \includegraphics[width=5.5cm]{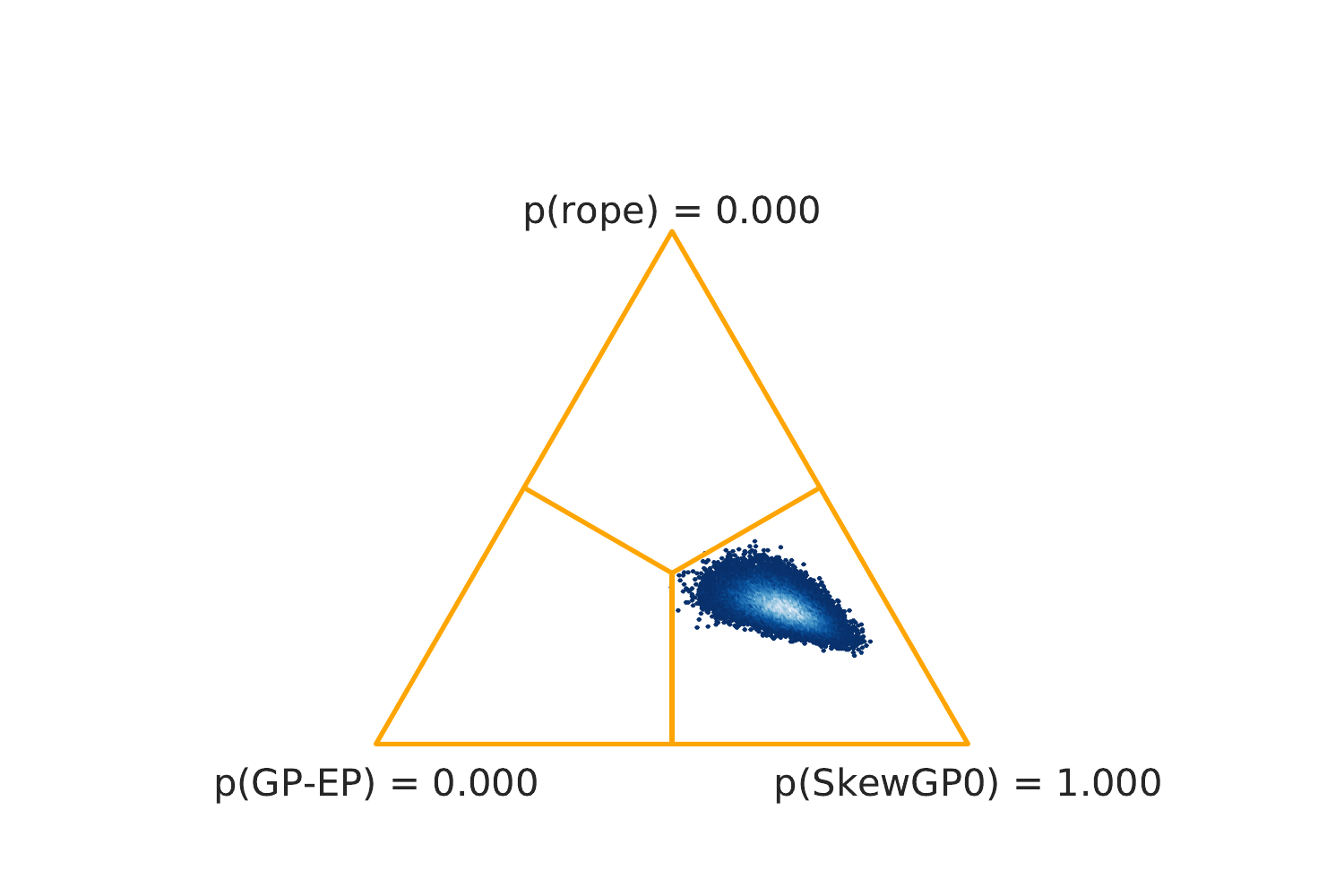}
\includegraphics[width=5.5cm]{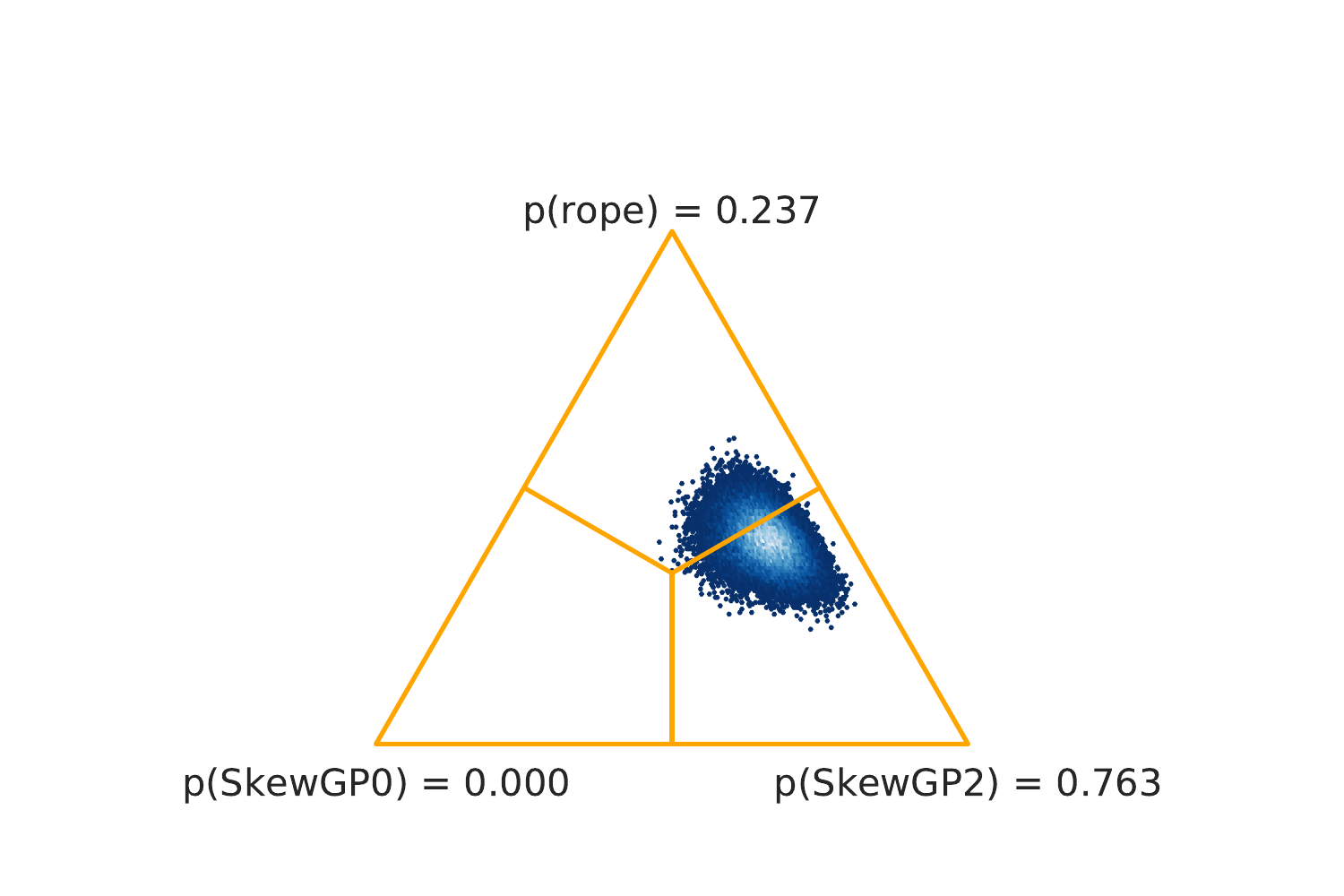}
\caption{Bayesian Wilcoxon signed-rank test}
\label{fig:triangle}
\end{figure}

\begin{figure}
\centering
  \includegraphics[width=5.5cm]{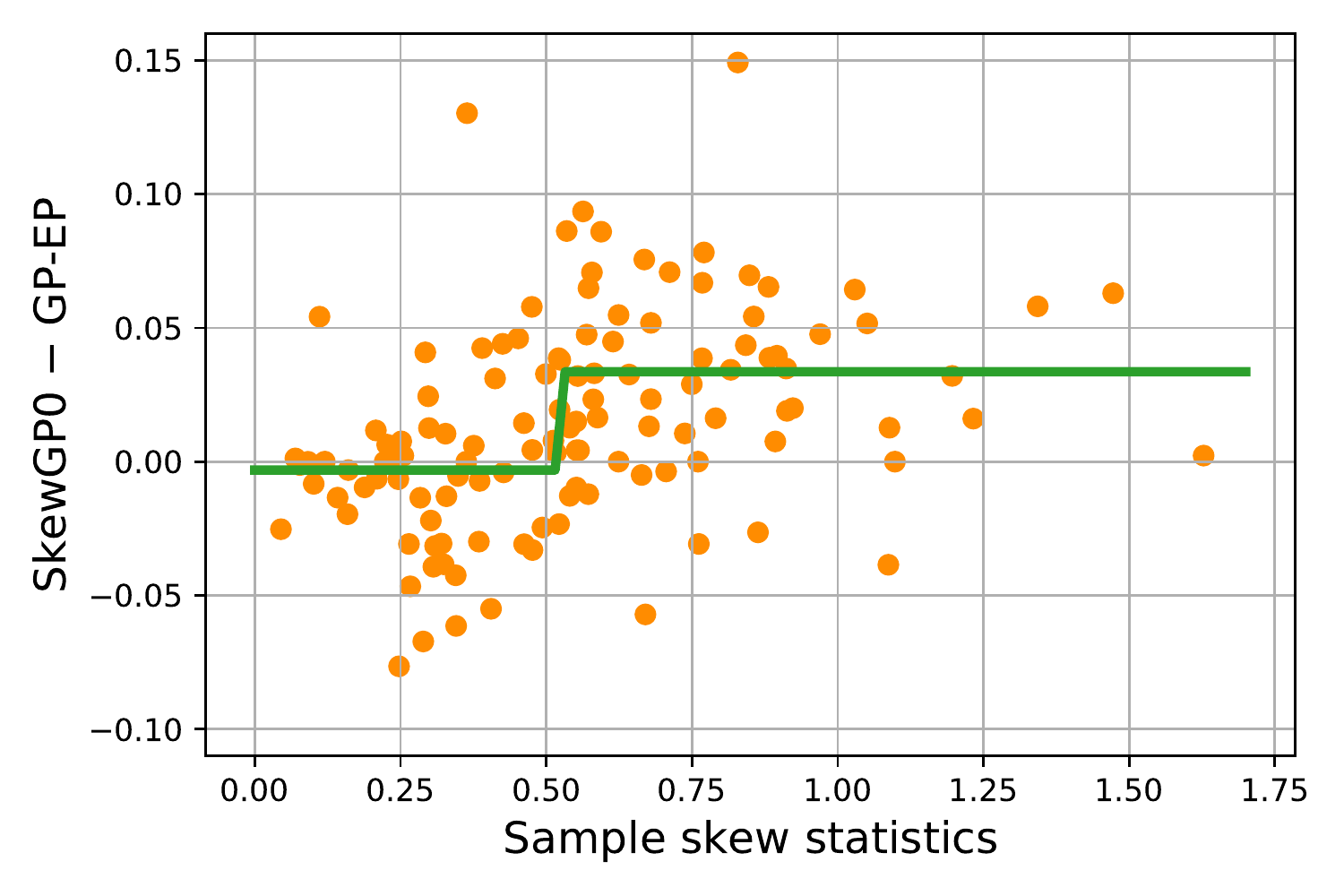}
  \includegraphics[width=5.5cm]{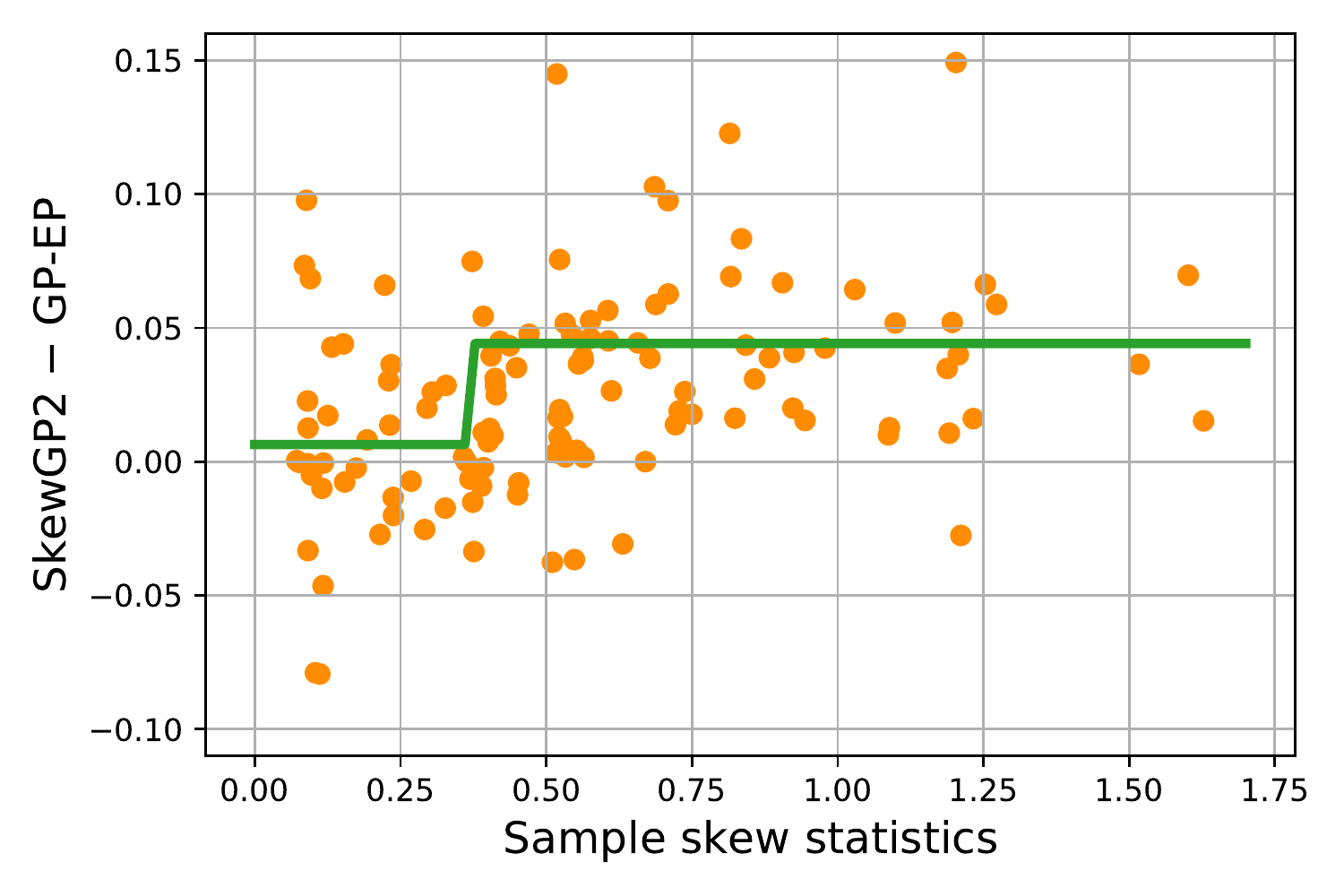}
\caption{Maximum skewness statistics versus difference 
in average information score. For visualisation purpose only, we bounded the y-axis to $[-0.11,0.16]$.}
\label{fig:tree}
\end{figure}

 From the Penn Machine Learning Benchmarks \citep{Olson2017PMLB},
 we have selected 124 datasets (number of features up to $500$). Since this pool includes non-binary class datasets, we have defined a binary sub-problem  by considering the first (class $0$) and second (class $1$) class.
 The resulting binarised subset of datasets includes datasets with  number of instances between  $100$ and $7400$. { We have  scaled the inputs to zero mean and unit standard deviation and used,  as performance measure,  the average information  in bits of the predictions about the test targets \citep{kuss2005assessing}:
 $$
 I(p_i^*,y_i)=\frac{y_i+1}{2}\log_2(p_i^*)+\frac{1-y_i}{2}\log_2(1-p_i^*)+1,
 $$
 where $p_i^*$ is the predicted probability for  class
 $1$. This score equals $1$ bit if the true label is predicted with absolute certainty, $0$ bits for random guessing and takes negative values if the prediction gives higher probability to the wrong class.  We have assessed the above performance measure for the three classifiers by using 5-fold cross-validation.
 
While we could use any kernel for GP-L, GP-EP and SkewGP, in this experiment we have chosen the RBF kernel with a lengthscale for each dimension.
Figure \ref{fig:penn} contrasts GP-L and GP-EP with
SkewGP0 (SkewGP with $s=0$) and SkewGP2 (SkewGP with $s=2$). 
We selected $s=2$ because we decided to use the same dimension for all datasets and, since there are several datasets where the ratio between the number of features and the number of instances is  high, a latent dimension $s>2$ leads to a number of  parameters that exceeds the number of instances affecting the convergence of the maximization of the marginal likelihood.  
The proposed SkewGP2 and SkewGP0  outperform the other two models for most data sets. 
The average information score  of SkewGP2 is $0.573$ (average accuracy $0.904$),
SkewGP0 is $0.557$ (acc. $0.882$), GP-EP is $0.542$ (acc. $0.859$) and GP-L  is $0.512$ (acc. $0.863$).

This claim is supported by a statistical analysis.
We have compared the three classifiers using the (nonparametric) Bayesian signed-rank test \citep{benavoli2014a,benavoli2016e}. This test declares two classifiers
practically equivalent when the difference of average information is less than 0.01 (1\%).  The interval $[-0.01,0.01]$ thus defines a region of practical equivalence (rope) for classifiers.
The test returns the posterior probability of  the vector
$[p(Cl_1 > Cl_2), p(Cl_1 \approx Cl_2), p(Cl_1 < Cl_2)]$ and, therefore, this posterior can be visualised in the probability simplex (Figure \ref{fig:triangle}). For the comparison GP-L vs.\ GP-EP, as expected it can seen  that GP-EP is better than  GP-L.\footnote{
It is well known that GP-EP usually achieves a more calibrated estimate of the class probability. Laplace's method gives over-conservative predictive probabilities \citep{kuss2005assessing}.}
Conversely, for GP-EP versus SkewGP0, 100\%  of the posterior mass is in the region in favor of SkewGP0, which is the region at the right bottom of the triangle. 
This confirms that SkewGP0 is  practically significantly better than GP-L and GP-EP.
The comparison SkewGP2 versus SkewGP0 shows that SkewGP2
has surely an average information score that is not worse than that of SkewGP0 and better with probability about $0.76$.

The  difference between GP-L and GP-EP, and SkewGP is that the  posterior of SkewGP can be skewed. Therefore, we expect  
SkewGP to outperform GP-L and GP-EP  on the datasets 
for which the  posterior is far from Normal (e.g., highly skewed). 
To verify that we have computed  the sample skewness  statistics (SS) for each test input $\mathbf{x}_i^*$:
$$
SS(\mathbf{x}_i^*)={\frac {\operatorname {E} \left[(f(\mathbf{x}_i^*)-\mu )^{3}\right]}{(\operatorname {E} \left[(f(\mathbf{x}_i^*)-\mu )^{2}\right])^{3/2}}}
$$
with $\mu=\operatorname {E} \left[(f(\mathbf{x}_i^*)\right]$
and the expectation $\operatorname {E}[\cdot]$ can be approximated using the posterior samples drawn as in  Theorem \ref{th:predictive}. Note that $SS(\mathbf{x}_i^*)=0$ for symmetric distributions.
Figure \ref{fig:tree}(left) shows, for each of the 124 datasets,
the difference between
the average information score of SkewGP0 and GP-EP in the y-axis, and $\max_{\mathbf{x}_i^*} SS(\mathbf{x}_i^*)$ for SkewGP0 in the x-axis. We used a regression tree (green line) to  
detect structural changes in the mean of these data points. It is evident that, for large values of the maximum skewness statistics,
SkewGP0 outperforms GP-EP (the average difference is positive).
Figure \ref{fig:tree}(right) reports a similar plot for 
 SkewGP2 and the difference is even more evident.
 This confirms that SkewGP on average outperforms GP-EP 
 in those datasets where the posterior is skewed and has a similar performance otherwise.
}

\subsection{Image classification}
{ We have also considered an image classification task:
\textit{Fashion MNIST}  dataset (each image is $28\times28 =784$ pixels and there are 10 classes: 0 T-shirt/top, 1 Trouser, 2 Pullover, 3 Dress, 4 Coat, 5 Sandal, 6 Shirt, 7 Sneaker, 8 Bag,
9 Ankle boot).
We randomly pooled 10000 images from the dataset and  divided them into two sets, with 5000 cases for training and 5000 for testing.
For each one of the 10 classes, we have defined a binary classification sub-problem by considering one class against all the other classes.
We have compared GP-EP and SkewGP2, that is a SkewGP
with latent dimension $s=2$ (for the same reason 
outlined in the previous section). We have initialised $r_i$ by taking $2$ random samples from the training data.
We have also considered two different kernels: RBF and the Neural Network kernel \citep{williams1998computation}.
Table \ref{tab:1} reports the  accuracy for each of the 10 binary classification sub-problems. For the RBF kernel, it can be noticed   that SkewGP2 outperforms GP-EP in all sub-problems.  For the NN kernel, the differences between the two models are less substantial (due to the higher performance of the NN kernel on this dataset) but in any case in favor of SkewGP2.
We have also reported, for both the models, the  computational time\footnote{More precisely, the table reports the average computational time for the RBF and NN kernel case.} (in minutes) needed to optimize the hyperparameters, to compute the posterior and to compute the predictions for all instances in the test set. This shows that SkewGP2 is also faster than GP-EP.\footnote{This is due to both the efficiency of lin-ess and
the batch approximation of the marginal likelihood.}
The last row reports the accuracy on the original multi-class
classification problem obtained by using the one-vs-rest heuristic, with the only goal of showing that the more accurate  estimate of the probability by SkewGP leads also to an increase in accuracy for one-vs-rest. A multi-class Laplace's approximation for GP classification  was developed by \cite{williams1998bayesian} and other implementations are for instance discussed by \cite{hernandez2011robust} and \cite{chai2012variational}, we plan to address multi-class classification in future work.


\begin{table}
\centering
 \rowcolors{2}{blue!6}{white}  
\begin{tabular}{|c|cc|cc|cc|}
    \rowcolor{blue!20} 
    & \multicolumn{2}{c|}{RBF kernel} & \multicolumn{2}{c|}{NN kernel} & \multicolumn{2}{c|}{}\\
& \multicolumn{2}{c|}{Accuracy}  & \multicolumn{2}{c|}{Accuracy} & \multicolumn{2}{c|}{Time}\\
    Class & GP-EP & SkewGP2 & GP-EP & SkewGP2 & GP-EP & SkewGP2  \\
  \hline
  0 &  0.945 &  \bf{0.955} & 0.956 &  \bf{0.958}   &   80.7 &  67.8 \\
  1 &  0.988&  \bf{0.990} & 0.988 &  \bf{0.991} &  125.7  &  68.0 \\
  2 &  0.923  &  \bf{0.934} &0.943 &  0.943 &  130.9 &  70.7 \\
  3 &  0.951 &  \bf{0.961}  & 0.953 &  \bf{0.961} &   79.2  &  56.9 \\
  4 &  0.932  &  \bf{0.946} & \bf{0.946} &  0.944 &  121.1 &  49.3 \\
  5 &  0.967  &  \bf{0.980} & 0.975 &  \bf{0.983} &  110.5 &  53.0 \\
  6 &  0.917  &  \bf{0.922} & 0.924 &  \bf{0.933} & 115.7 &  65.1 \\
  7 &  0.970  &  \bf{0.974} & 0.981 &  \bf{0.983}&   113.6 &  66.3 \\
  8 &  0.969 &  \bf{0.979} &  0.980 &  \bf{0.981} &121.5  &  51.4 \\
  9 &  0.980  &  \bf{0.982} &   0.983 &  \bf{0.986}  & 137.0 &  55.0 \\
  \hline
  \text{One-vs-rest} & 0.785 & \bf{0.821} & 0.823  &  \bf{0.844} & 
&  \\
  \hline
\end{tabular}
 \caption{Image classification}
\label{tab:1}
\end{table}


Our  goal is assessing  the accuracy but also the quality of the probabilistic predictions.
Figure \ref{fig:6}, plot (a1), shows, for the RBF kernel case and for each instance
in the test set of the binary sub-problem relative to class 8 vs.\ rest,  the value of  of the mean predicted probability of class \textit{rest} for  SkewGP2 (x-axis) and GP-EP (y-axis).
Each instance is represented as a blue point.
The red points highlight the instances that were misclassified
by GP-EP. Figure (a2) shows the same plot, but the red points
are now the instances that were misclassified
by SkewGP2.
By comparing (a1) and (a2), it is evident that SkewGP2 provides a  higher quality of the probabilistic predictions.

 SkewGP2 also returns a better estimate of its own uncertainty.
This is showed in plots (b1) vs.\ (b2).
For each instance in the test set and for each sample
from the posterior, we have computed the predicted class
(the class that has probability greater than $0.5$). For each test instance, we have then computed the standard deviation of all these predictions and used it to color the scatter plot
of the mean predicted probability.
In this way, we have a visual representation
of  first order (mean predicted probability)
and second order (standard deviation of the predictions)
uncertainty.
Plot \ref{fig:6}(b1) is relative to GP-EP and shows
that GP-EP confidence is low only for the instances
whose mean predicted probability is close to $0.5$.
This is not reflected in the value 
of the mean predicted probability for the misclassified
instances (compare plot (a1) and (b1)).
We have also computed the histogram of the standard deviation of the predictions for those instances that were misclassified
by GP-EP in Figure (c1). Note that, the peak of the histogram
corresponds to very low standard deviation, that means
GP-EP has misclassified instances that have low second order uncertainty. This implies that the model is overestimating its confidence.
Conversely, the second order uncertainty of  
 SkewGP2 is clearly consistent, see plot (a2) and (b2), and in particular the histogram in (c2) -- the peak is in correspondence of high values of the standard deviation of the predictions.
 In other words, SkewGP2 has mainly misclassified  instances with high second order uncertainty, that is what we expect from a calibrated probabilistic model.
We have reported additional examples of the better calibration of SkewGP2  for the MNIST and German-road sign dataset in appendix.

\begin{figure}[htp]
\centering
\begin{tabular}{c @{\qquad} c }
    \includegraphics[width=.48\linewidth]{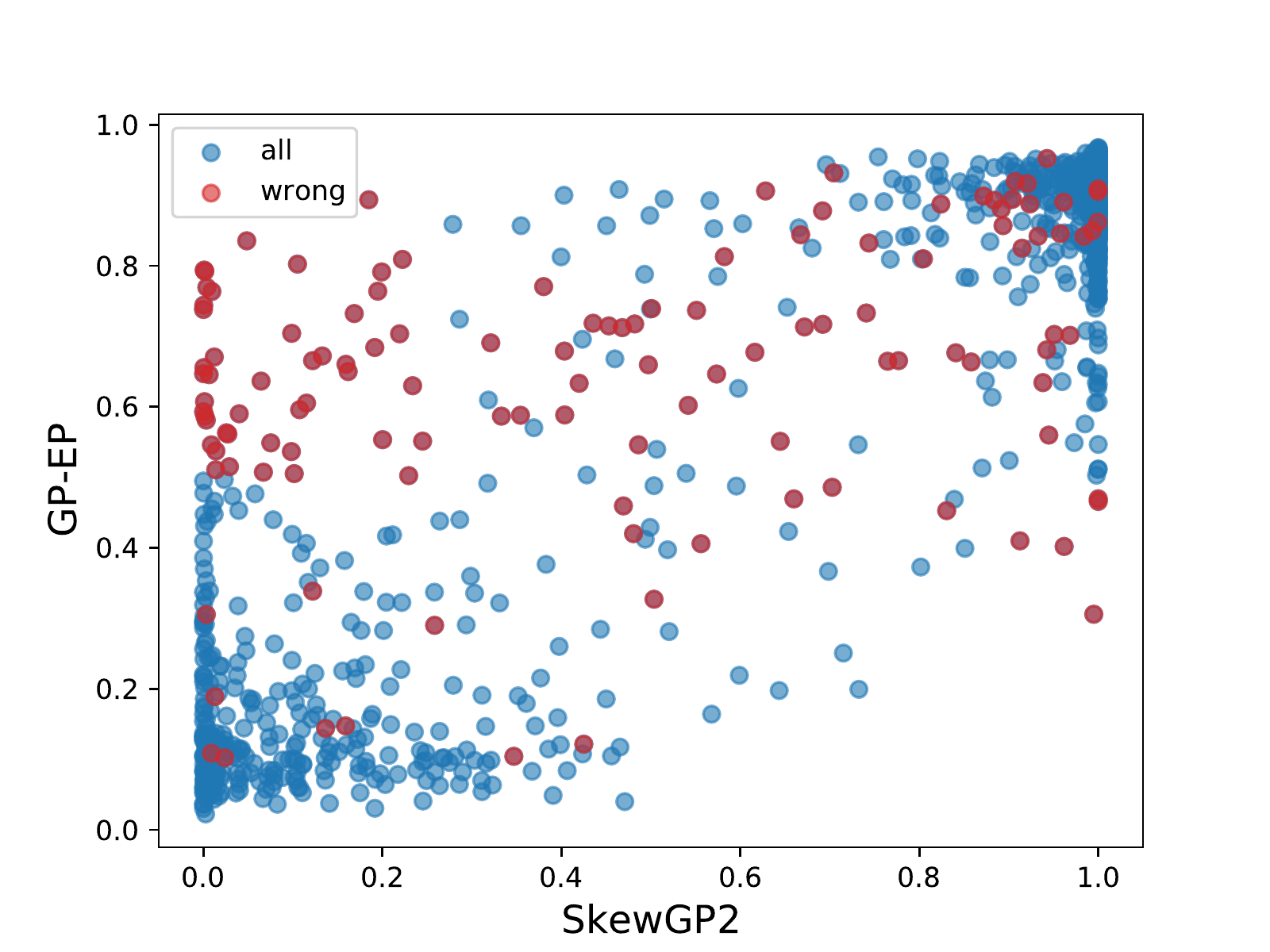} &
    \includegraphics[width=.48\linewidth]{{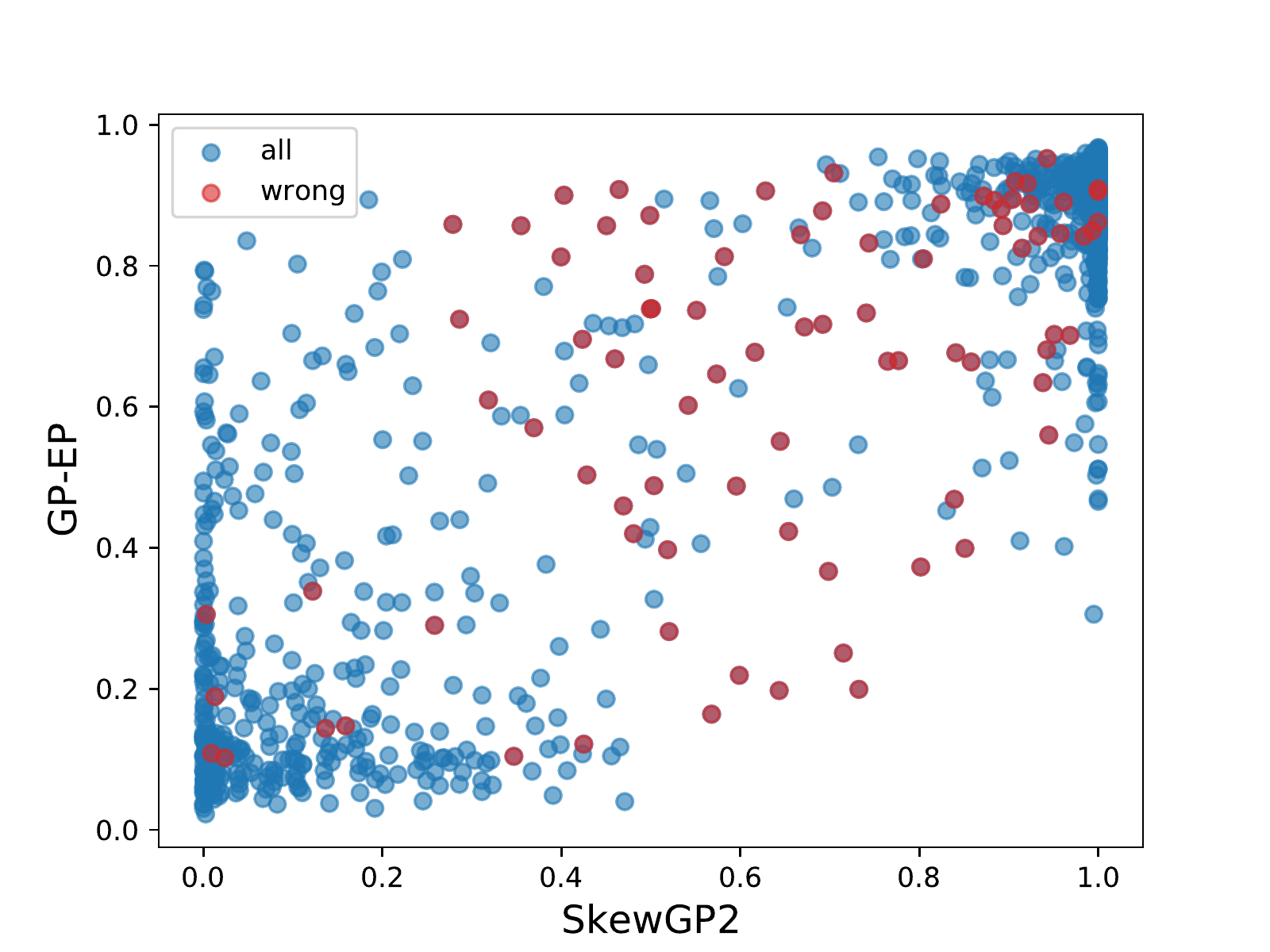}} \\
    \small (a1) & \small (a2)\\
    \includegraphics[width=.49\linewidth]{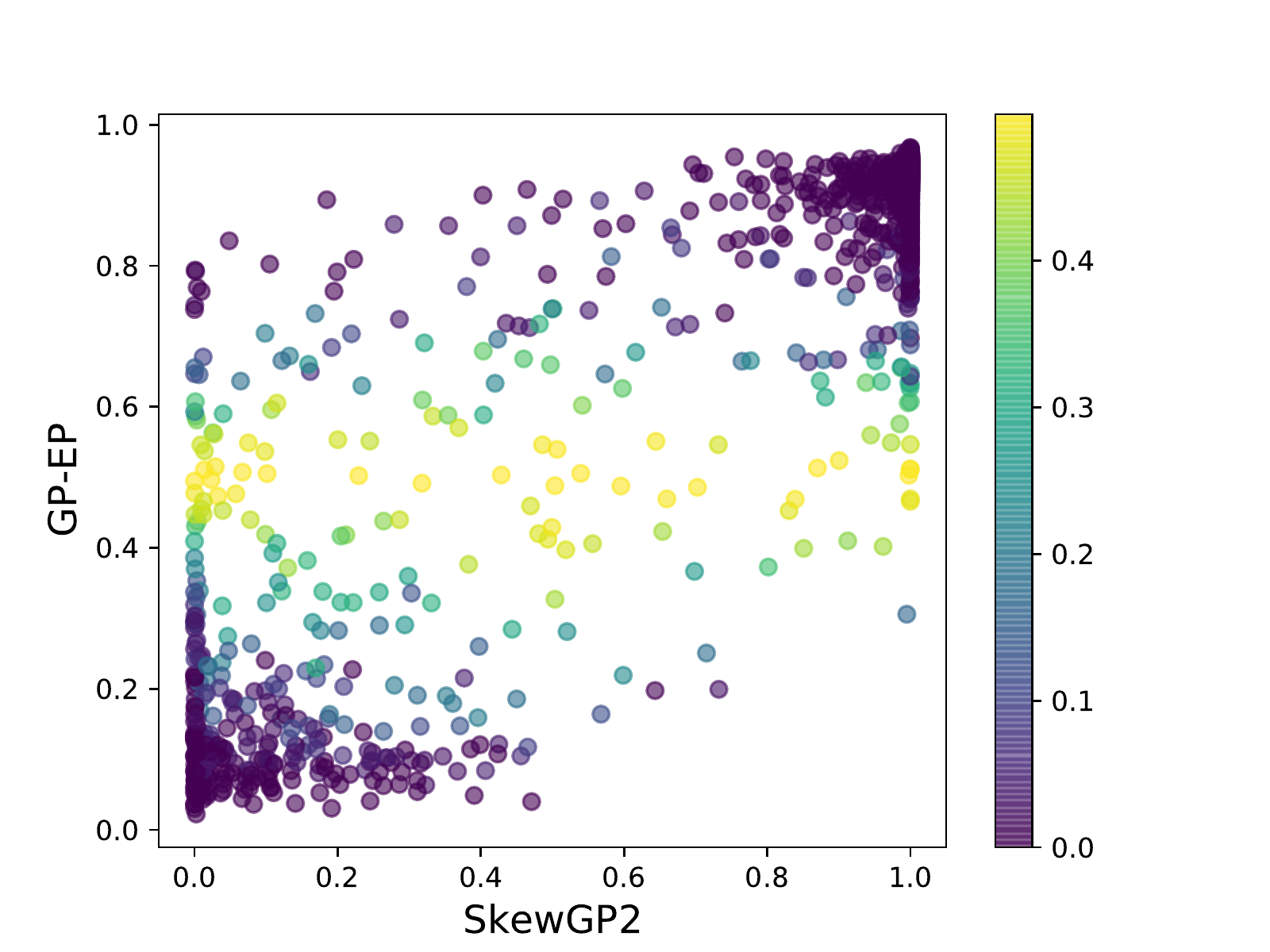} &
    \includegraphics[width=.49\linewidth]{{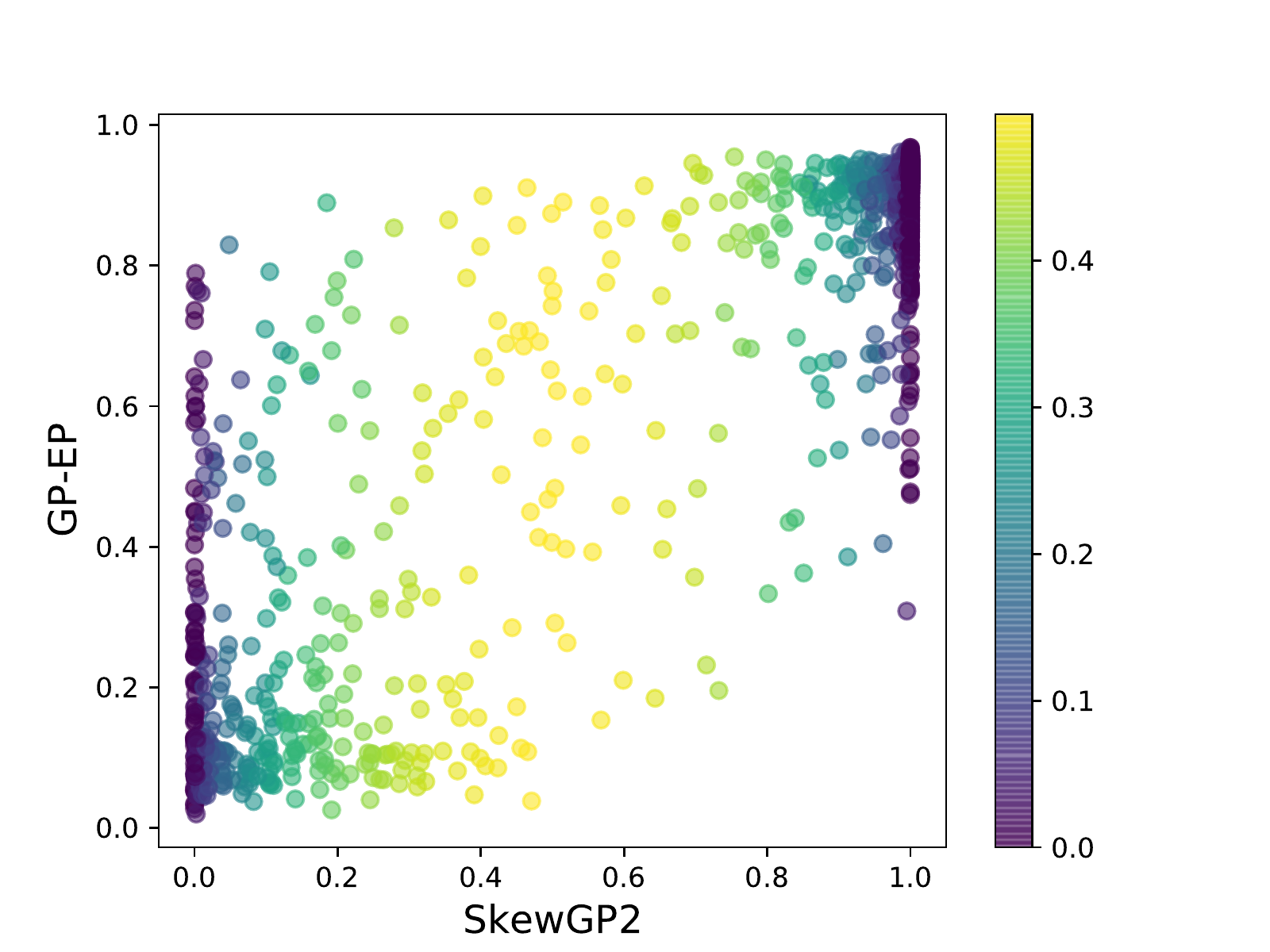}} \\
    \small (b1) & \small (b2)\\
    \includegraphics[width=.42\linewidth]{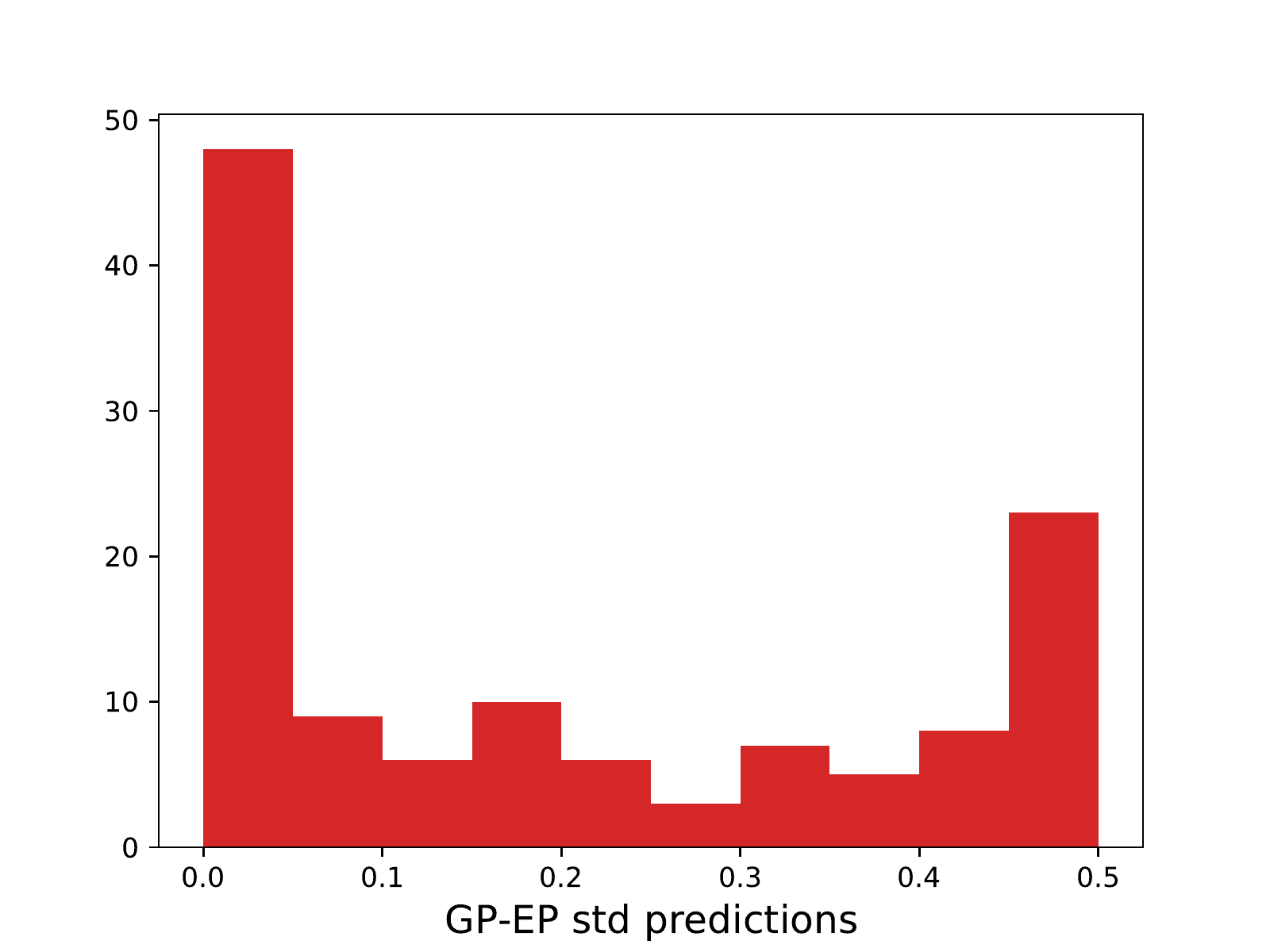} &
    \includegraphics[width=.42\linewidth]{{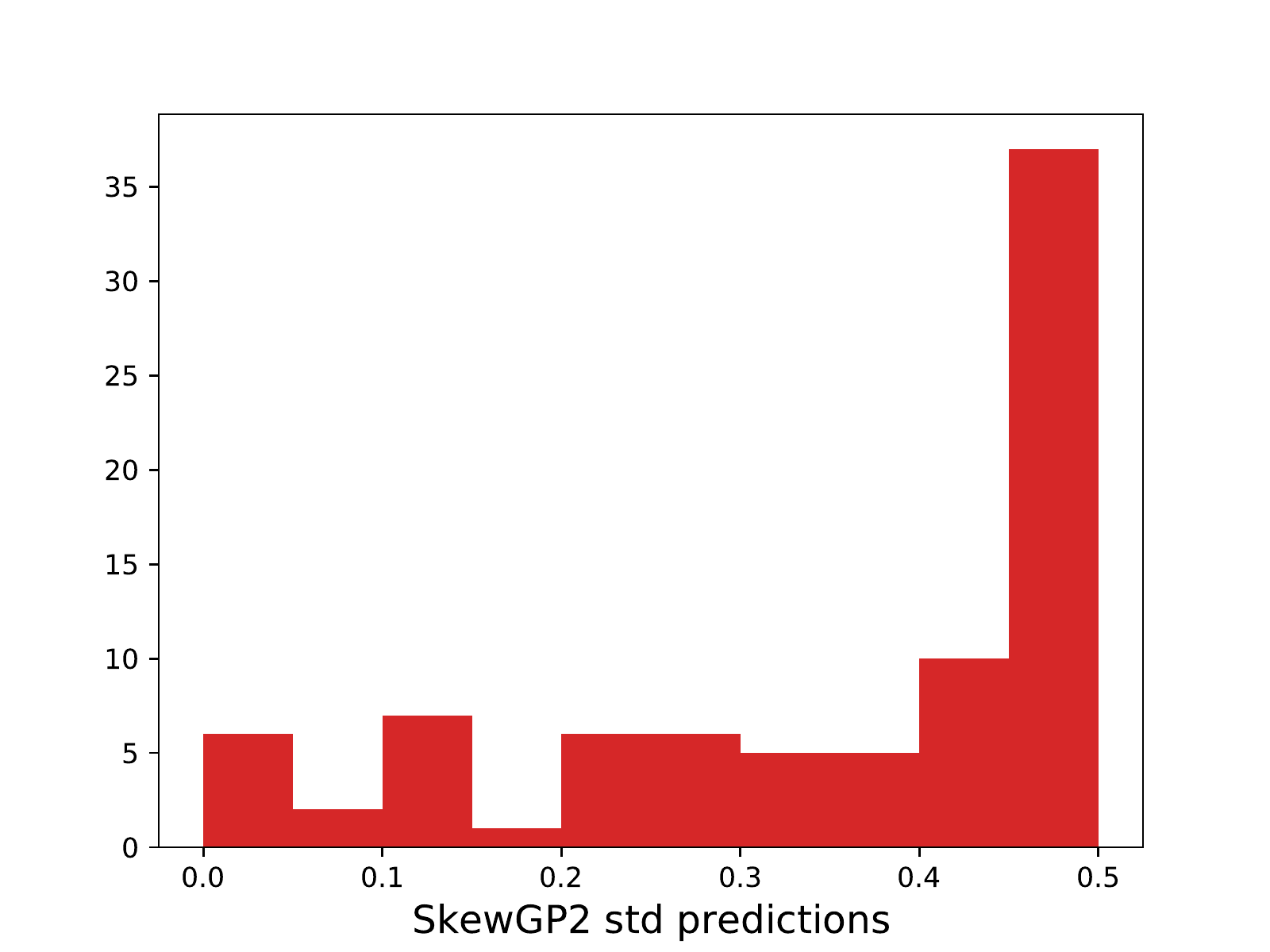}} \\
    \small (c1) & \small (c2)\\
   \end{tabular}
\caption{Fashion MNIST dataset}
\label{fig:6}
\end{figure}

}

%
%

\section{Conclusions}
We  have introduced the Skew Gaussian process (SkewGP) as an alternative to Gaussian processes for classification.
We have shown  that SkewGP and the probit likelihood are conjugate and provided  marginals and closed form conditionals.  We have also shown that SkewGP contains the GP as a special case and, therefore, SkewGPs inherit all good properties of GPs and increase their flexibility.    The SkewGP prior was applied in classification  showing  improved  performance over GPs (Laplace's method and Expectation Propagation approximations).

As future work, we plan to study other more native ways to parametrize the skewness matrix $\Delta$ that do not rely on an underlying kernel. Moreover, we plan to investigate the possibility of using inducing points, as for sparse GPs, to reduce the computational load for matrix operations (complexity $O(n^3)$ with storage demands of $O(n^2)$) {as well as deriving the
posterior for the multi-class classification problem.}


\bibliographystyle{spbasic}
	\bibliography{biblio}

%

\appendix
\section{Proofs}

\paragraph{Proposition \ref{prop:1}}
To prove that, we exploit Kolmogorov's extension theorem
\citep{orbanz2009construction}.
Suppose that a family $\mathcal{F}^I$ of probability measures are the I-finite-dimensional marginals of an infinite-dimensional measure $\mathcal{F}$ (a ``stochastic process''). Each measure
$\mathcal{F}^I$ belongs to the finite-dimensional
subspace of dimension $I$. Given two marginals $\mathcal{F}^I,\mathcal{F}^J$, as marginals of the same measure $\mathcal{F}$, they must be marginals of each other, that is  
\begin{equation}
\label{eq:projective}
\mathcal{F}^I=\mathcal{F}^{J\downarrow I} \text{ whether }I \subset J,
\end{equation}
where $\cdot \downarrow I$ denotes the projection onto the subspace of dimension $I$.
A family of probability measures that satisfies \eqref{eq:projective} is called a projective family. The  Kolmogorov's extension theorem states that  any projective family on the finite-dimensional subspaces of an infinite-dimensional product space uniquely defines a stochastic process on the space.
This means that we can define a nonparametric Bayesian model from a finite-dimensional distribution by simply verifying that \eqref{eq:projective} holds. From \eqref{eq:marginalFinDim}, it then immediately
follows that the definition \eqref{eq:SGprocess} uniquely defines a stochastic process and, therefore, SkewGP is a
well-defined stochastic process.

\paragraph{Theorem \ref{th:1}}
We aim to derive the posterior of $f(X)$.
The joint distribution of $f(X),\mathcal{D}$ is 
\begin{align}
\nonumber
 &p(\mathcal{D}|f(X))p(f(X))\\
 \nonumber
 &=\Phi_n(\Dmat\bl[f];I_n)\;\text{SUN}_{n,s}(\bxi,\Omega,\Delta,\bgamma,\Gamma)\\
 \nonumber
 &=\Phi_n(\Dmat\bl[f];I_n)\;\phi_n(\bl[f]-\bxi;\Omega)\frac{\Phi_s\left(\bgamma+\Delta^T\bar{\Omega}^{-1}\Domega^{-1}(\bl[f]-\bxi);\Gamma-\Delta^T\bar{\Omega}^{-1}\Delta\right)}{\Phi_s\left(\bgamma;\Gamma\right)} \\
 \label{eq:numjoint}
 &\propto \phi_n(\bl[f]-\bxi;\Omega)\Phi_n(\Dmat \bl[f];I_n)\Phi_s\left(\bgamma+\Delta^T\bar{\Omega}^{-1}\Domega^{-1}(\bl[f]-\bxi);\Gamma-\Delta^T\bar{\Omega}^{-1}\Delta\right)
\end{align}
where we denoted $\bl[f]= f(X) \in \mathbb{R}^n$ and omitted the dependence on $X$. First, note that 
$$
\begin{aligned}
&\Phi_n(\Dmat\bl[f];I_n) \\
&=\Phi_n\left(\Dmat\bxi+(\bar{\Omega}\Domega \Dmat^T)^{T}\bar{\Omega}^{-1}\Domega^{-1}(\bl[f]-\bxi);(\Dmat\Omega \Dmat^T+I_n)-\Dmat \Domega \bar{\Omega}\Domega \Dmat^T \right)
\end{aligned}
$$ 
Therefore, we can write
\begin{align}
\nonumber
 &\Phi_n(\Dmat\bl[f];I_n)\Phi_s\left(\bgamma+\Delta^T\bar{\Omega}^{-1}\Domega^{-1}(\bl[f]-\bxi);\Gamma-\Delta^T\bar{\Omega}^{-1}\Delta\right) \\
 \nonumber
 &=\Phi_n\left(\Dmat\bxi+(\bar{\Omega}\Domega \Dmat^T)^{T}\bar{\Omega}^{-1}\Domega^{-1}(\bl[f]-\bxi);(\Dmat\Omega \Dmat^T+I_n)-\Dmat \Domega \bar{\Omega}\Domega \Dmat^T \right)\\
 \nonumber
 &\cdot\Phi_s\left(\bgamma+\Delta^T\bar{\Omega}^{-1}\Domega^{-1}(\bl[f]-\bxi);\Gamma-\Delta^T\bar{\Omega}^{-1}\Delta\right) \\
 \label{eq:prodcdf}
 &=\Phi_{s+n}(m;M)
\end{align}
with
$$
m=\begin{bmatrix}
   \bgamma+\Delta^T\bar{\Omega}^{-1}\Domega^{-1}(\bl[f]-\bxi)\\
   \Dmat\bxi+(\bar{\Omega}\Domega \Dmat^T)^{T}\bar{\Omega}^{-1}\Domega^{-1}(\bl[f]-\bxi)
  \end{bmatrix},
$$
and
$$
M=\begin{bmatrix}
  \Gamma-\Delta^T\bar{\Omega}^{-1}\Delta & 0\\
  0 & (\Dmat\Omega \Dmat^T+I_n)-\Dmat \Domega \bar{\Omega}\Domega \Dmat^T 
  \end{bmatrix}.
$$
From \eqref{eq:numjoint}--\eqref{eq:prodcdf} and the definition of the PDF of the SUN distribution \eqref{eq:sun}, 
we can easily show  that we can rewrite \eqref{eq:numjoint} as a SUN distribution with updated parameters:
$$
\begin{aligned}
\tilde{\xi} & =\xi,\\
\tilde{\Omega} &= \Omega, \\
\tilde{\Delta} &=[\Delta,~~\bar{\Omega}\Domega \Dmat^T],\\
\tilde{\bgamma}& =[\bgamma,~~\Dmat\xi]^T, \\
\tilde{\Gamma}&=\begin{bmatrix}
         \Gamma & \quad\Delta^T \Domega \Dmat^T\\
        \Dmat \Domega \Delta & \quad (\Dmat \Omega \Dmat^T + I_n) \end{bmatrix}.
\end{aligned}
$$

\paragraph{Corollary \ref{co:ml}} This follows directly
from the above proof by observing that 
$\Phi_{s+n}(\tilde{\bgamma},\tilde{\Gamma})$ is the normalization constant of the posterior and, therefore, the marginal likelihood is
$$
\begin{aligned}
\frac{\Phi_{s+n}(\tilde{\bgamma},\tilde{\Gamma})}
{\Phi_{s}(\bgamma,\Gamma)}.
\end{aligned}
$$

\paragraph{Corollary \ref{co:predictive}} 
 Denote with $\hat{f}(\hat{X})=[f(X)^T,f({\bx^*})^T]^T$
and  observe  that the predictive distribution is 
by definition
$$
\int\int \Phi(f^*;1)p(f^*|\bl[f])p(\bl[f]|\mathcal{D}) d\bl[f] df^*,
$$
with $f^*:=f({\bx^*})$ and $\bl[f] = f(X)$. Note that we have omitted the dependence
on $\bx^*,X$ for easier notation ($p(f^*|\bl[f])$ corresponds to $p(f^*|\bx^*,X,\bl[f])$). We can write the posterior as
$$
p(\bl[f]|\mathcal{D})=\dfrac{p(\mathcal{D}|\bl[f])p(\bl[f])}{p(\mathcal{D})},
$$
and so
$$
\begin{aligned}
&\int\int \Phi(f^*;1)p(f^*|f)p(\bl[f]|\mathcal{D}) d\bl[f] df^*\\
&=\dfrac{1}{p(\mathcal{D})}\int\int \Phi(f^*;1)p(f^*|\bl[f])p(\mathcal{D}|\bl[f])p(\bl[f]) d\bl[f] df^*\\
&=\dfrac{1}{p(\mathcal{D})}\int\int \Phi(f^*;1)p(\mathcal{D}|\bl[f])p(\bl[f],f^*) d\bl[f] df^*.\\
\end{aligned}
$$
Observe that 
$$
\Phi(f^*;1)p(\mathcal{D}|\bl[f])=\Phi(f^*;1)\Phi_n(\Dmat\bl[f];I_n)=\Phi_{n+1}(\hat{\Dmat}\hat{f};I_{n+1})
$$
with $\hat{\Dmat}=\text{diag}(2y_1-1,\dots,2y_{n+1}-1)$
and $y_{n+1}=0.5$. Note that $2y_{n+1}-1=2 \cdot 0.5-1=0$ and this is the reason why we have introduced the dummy class value $1/2$.

 Observe  that
$$
\int\int \Phi_{n+1}(\hat{\Dmat}\hat{f};I_{n+1})p(\bl[f],f^*) d\bl[f] df^*
$$
is the marginal likelihood of a SkewGP posterior
corresponding to the augmented dataset
 $\hat{X}=[X^T,{\bx^*}^T]^T$, $\hat{y}=[y^T,1/2]^T$.
 Therefore, we have that
 $$
 \frac{1}{p(\mathcal{D})}\int\int \Phi_{n+1}(\hat{\Dmat}\hat{f};I_{n+1})p(\bl[f],f^*) d\bl[f] df^*=\dfrac{\frac{\Phi_{s+n+1}(\tilde{\bgamma}^*,\tilde{\Gamma}^*)}
{\Phi_{s}(\bgamma,\Gamma)}}{\frac{\Phi_{s+n}(\tilde{\bgamma},\tilde{\Gamma})}
{\Phi_{s}(\bgamma,\Gamma)}},
 $$
 where $\tilde{\bgamma}^*,\tilde{\Gamma}^*$ are the corresponding matrices  of the posterior computed for the augmented dataset. 

 \paragraph{Proposition \ref{prop:frechet}} 
 This follows by the Fr{\'e}chet inequality:
 $$
 Pr(A_1,A_2,\dots,A_b)\geq \max\left(0, \sum_{i=1}^b Pr(A_i)-(b-1)\right),
 $$
 where $A_i$ are  events. In fact, note that 
 $$
 \Phi_{s+n}(\tilde{\bgamma};\tilde{\Gamma})=Pr(u_{1}\geq\tilde{\bgamma}_1,\dots,u_{s+n}\geq \tilde{\bgamma}_{s+n})
 $$
 where $Pr$ is computed w.r.t.\ the PDF of a multivariate distribution with zero mean and covariance $\tilde{\Gamma}$.
 
 \paragraph{Theorem \ref{th:predictive}}
 The proof follows straightforwardly from that of Corollary
 \ref{co:predictive}.
\subsection{Additional image classification examples}
We have defined a binary sub-problem from the German Traffic Sign data by considering  Speed-Limit 30 vs.\ 50 and
from MNIST digit dataset by considering 3 vs.\ 5.
We have compared GP-L vs.\ SkewGP$_2$ (both with RBF kernel).
Table \ref{tab:1} reports the average accuracy
and shows again SkewGP$_2$ outperforms GP-L
(GP-EP achieves lower accuracy than GP-L in both cases).
\begin{table}[htp!]
\centering
 \rowcolors{2}{blue!6}{white}   
  \begin{tabular}{|c | c|c|c|}
    \rowcolor{blue!20}              
       \hline
    \textbf{Dataset}     & \textbf{classes}  & \textbf{GP-L}  & \textbf{SkewGP$_2$}        \\ \hline
   German Traffic Sign & Speed-Limit 30 vs.\ 50  & 0.96  & 0.98 \\
    MNIST & 3 vs.\ 5 & 0.80  & 0.90\\
    \hline
  \end{tabular}
  \caption{Image classification}
  \label{tab:1}
\end{table}

Our  goal is assessing  the accuracy but also the quality of the probabilistic predictions.
Figure \ref{fig:6}, plot (a1), shows, for each instance
in the test set (one of the fold) of the MNIST datatset , the value of  of the mean predicted probability of class $5$
for  SkewGP$_2$ (x-axis) and GP-L (y-axis).
Each instance is represented as a blue point.
The mean predicted probability ranges in 
$[0.41,0.53]$ for GP-L and in $[0.25,0.8]$ for SkewGP$_2$.
The red points highlight the instances that were misclassified
by GP-L (plot (a2) reports the images of some of the misclassified instances included in the rectangle).
Figure (b1) shows the same plot, but the red points
are now the instances that were misclassified
by SkewGP$_2$.
By comparing (a) and (b), it is evident that SkewGP$_2$ provides a  higher quality of the probabilistic predictions.

 SkewGP$_2$ also returns a better estimate of its own uncertainty.
This is showed in plots (c1) and (c2).
For each instance in the test set and for each sample
from the posterior, we have computed the predicted class
(the class that has probability greater than $0.5$). For each test instance, we have then computed the standard deviation of all these predictions and used it to color the scatter plot
of the mean predicted probability.
In this way, we have a visual representation
of  first order (mean predicted probability)
and second order (standard deviation of the predictions)
uncertainty.
Plot \ref{fig:6}(c1) is relative to GP-L and shows
that GP-L confidence is low only for the instances
whose mean predicted probability is close to $0.5$.
This is not reflected in the value 
of the mean predicted probability for the misclassified
instances (compare plot (a1) and (c1) and note that
the red spot in (a1) is outside the yellow area in (c1)).
Conversely, the second order uncertainty of  
 SkewGP$_2$ is clearly consistent with plot (b1).


\begin{figure}
\centering
\begin{tabular}{c @{\qquad} c }
    \includegraphics[width=.48\linewidth]{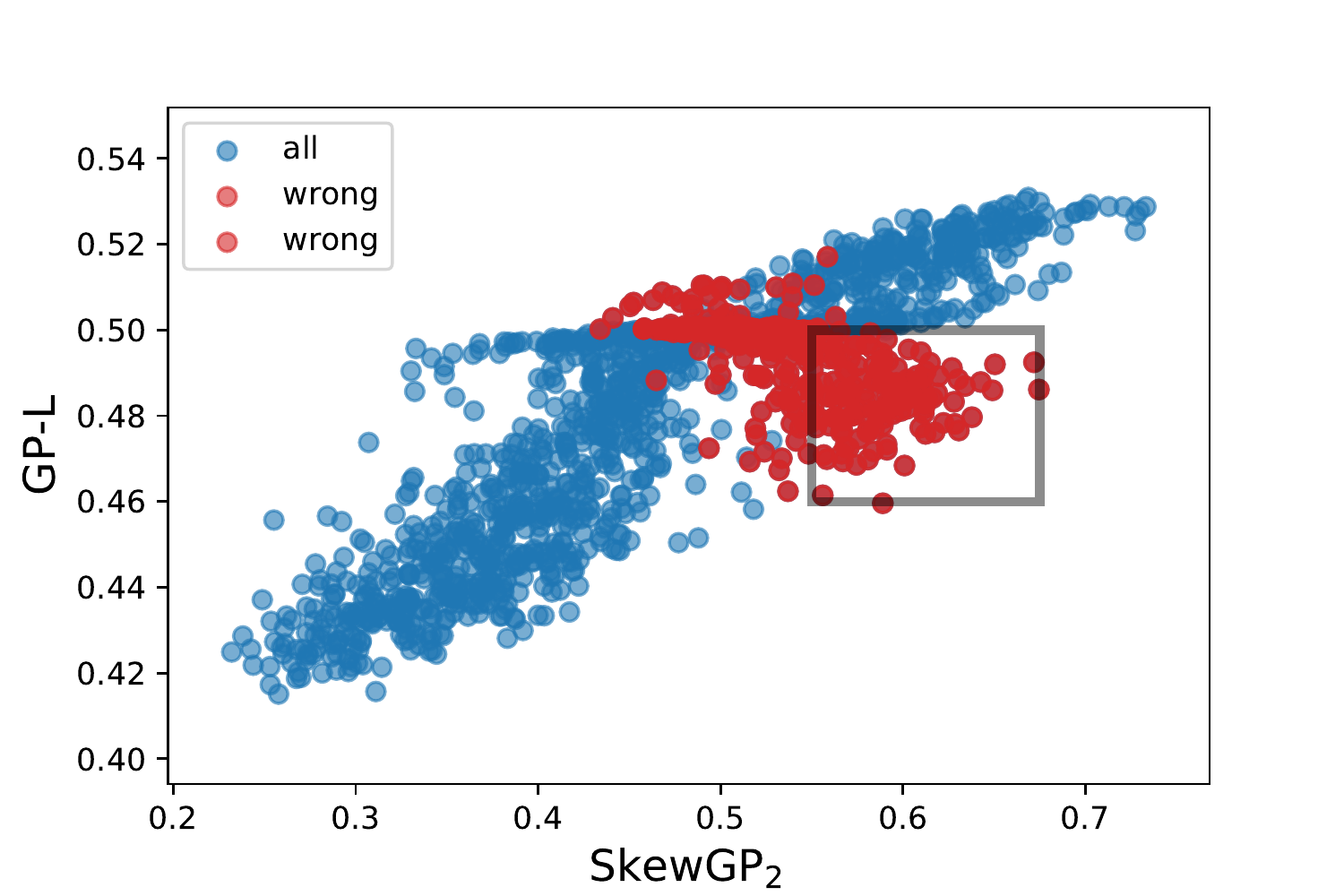} &
    \includegraphics[width=.28\linewidth,trim={3.5cm 1cm 3.5cm 0 }, clip]{{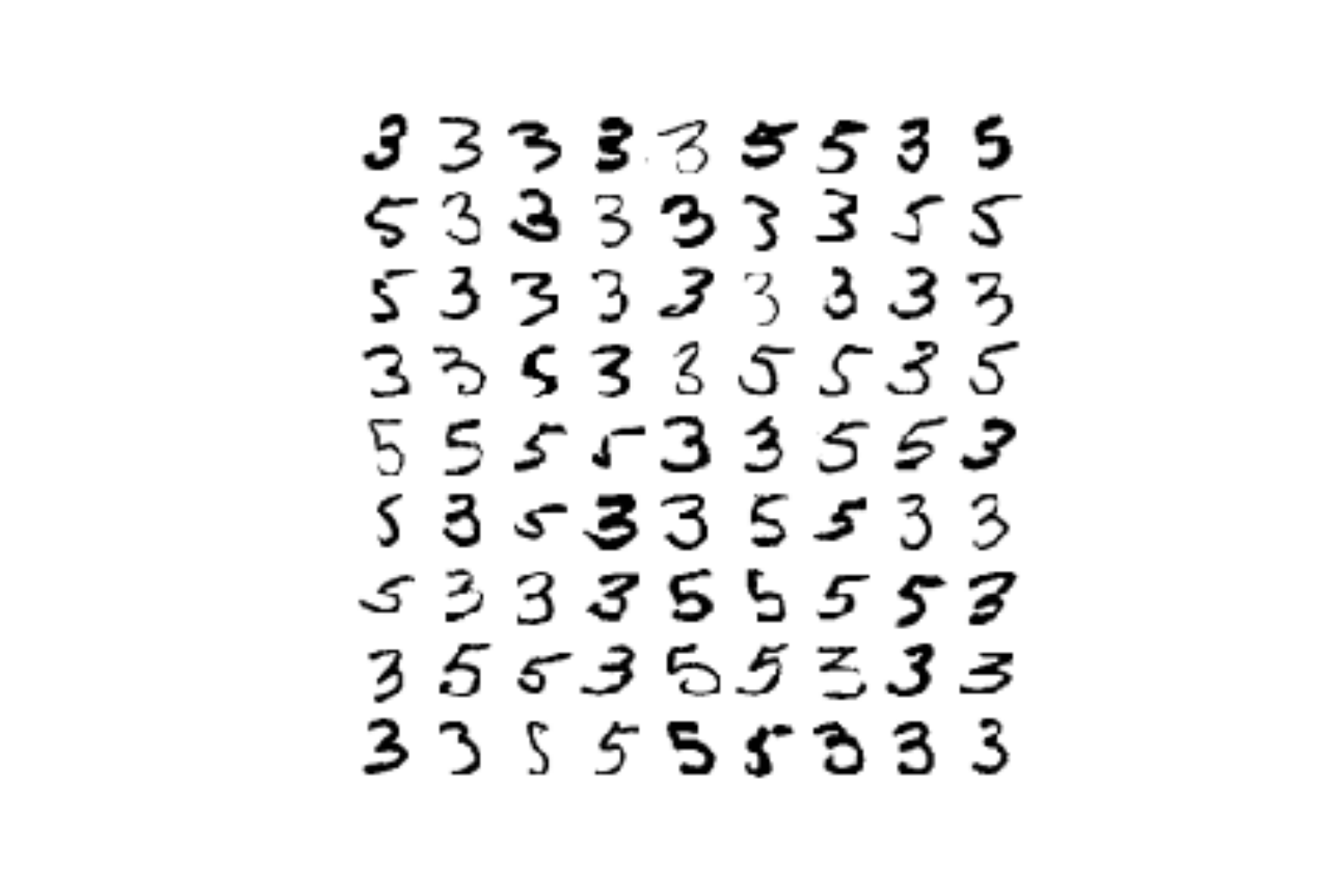}} \\
    \small (a1) & \small (a2)\\
    \includegraphics[width=.48\linewidth]{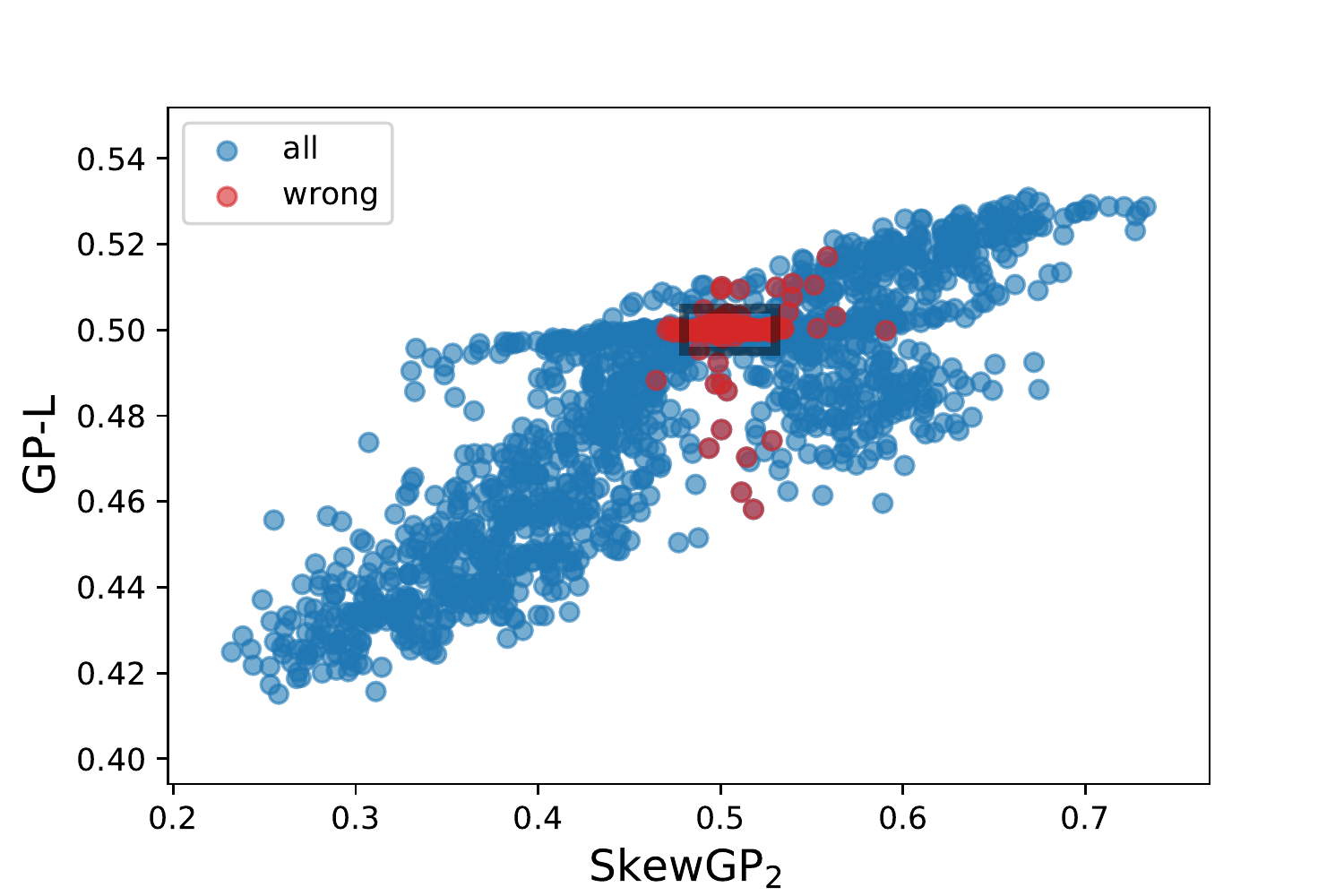} &
    \includegraphics[width=.28\linewidth,trim={3.5cm 1cm 3.5cm 0 }, clip]{{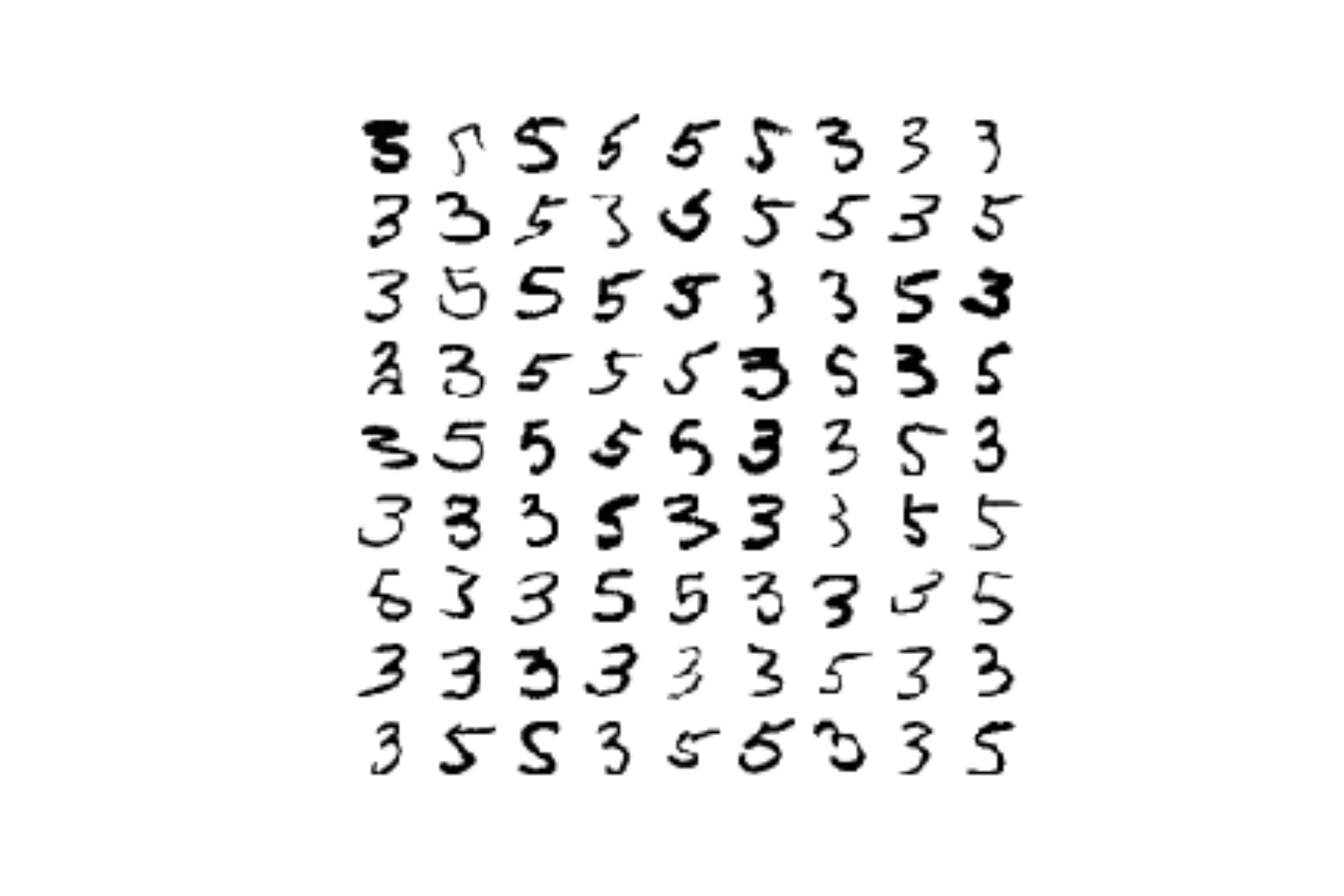}} \\
    \small (b1) & \small (b2)\\
    \includegraphics[width=.48\linewidth]{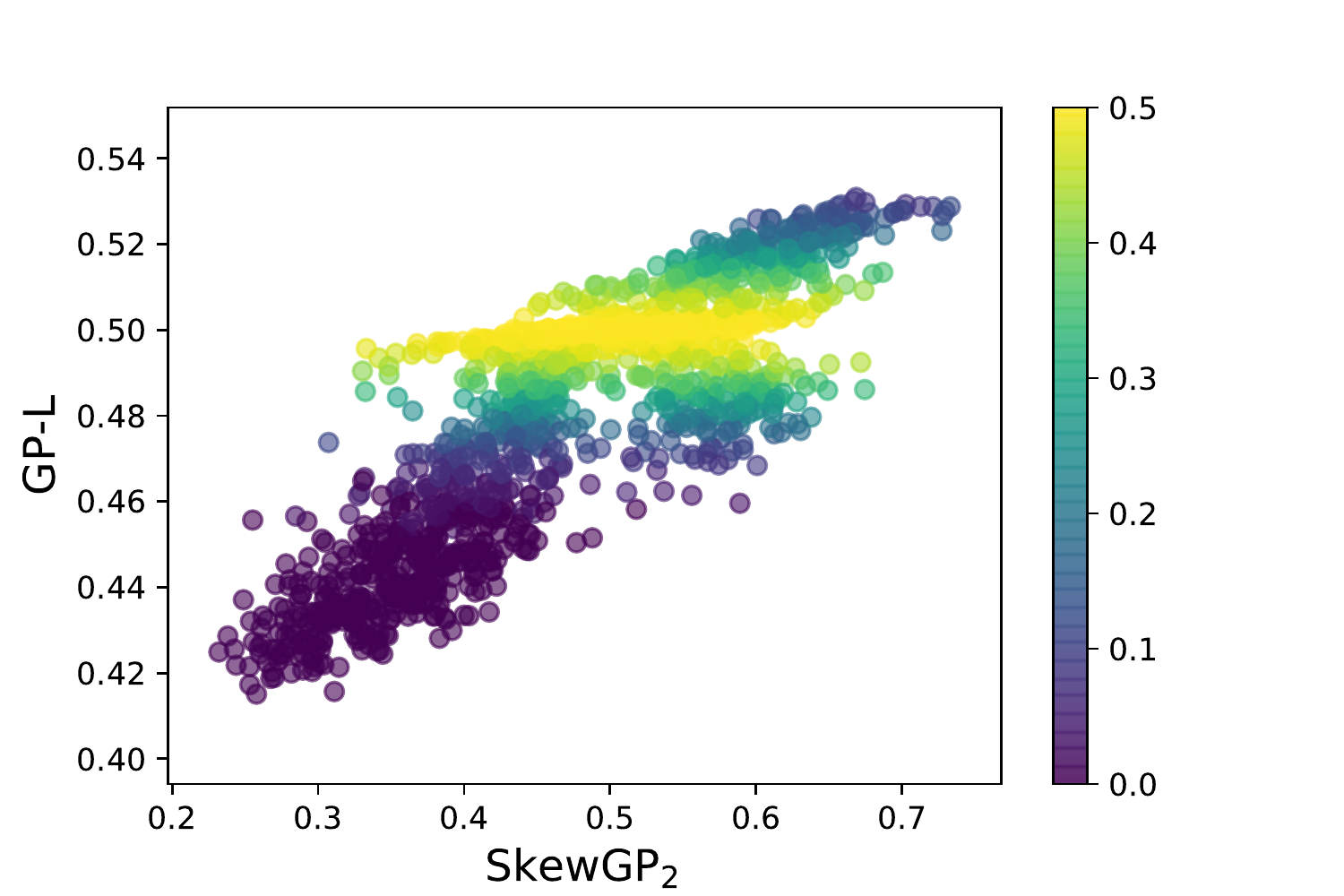} &
    \includegraphics[width=.48\linewidth]{{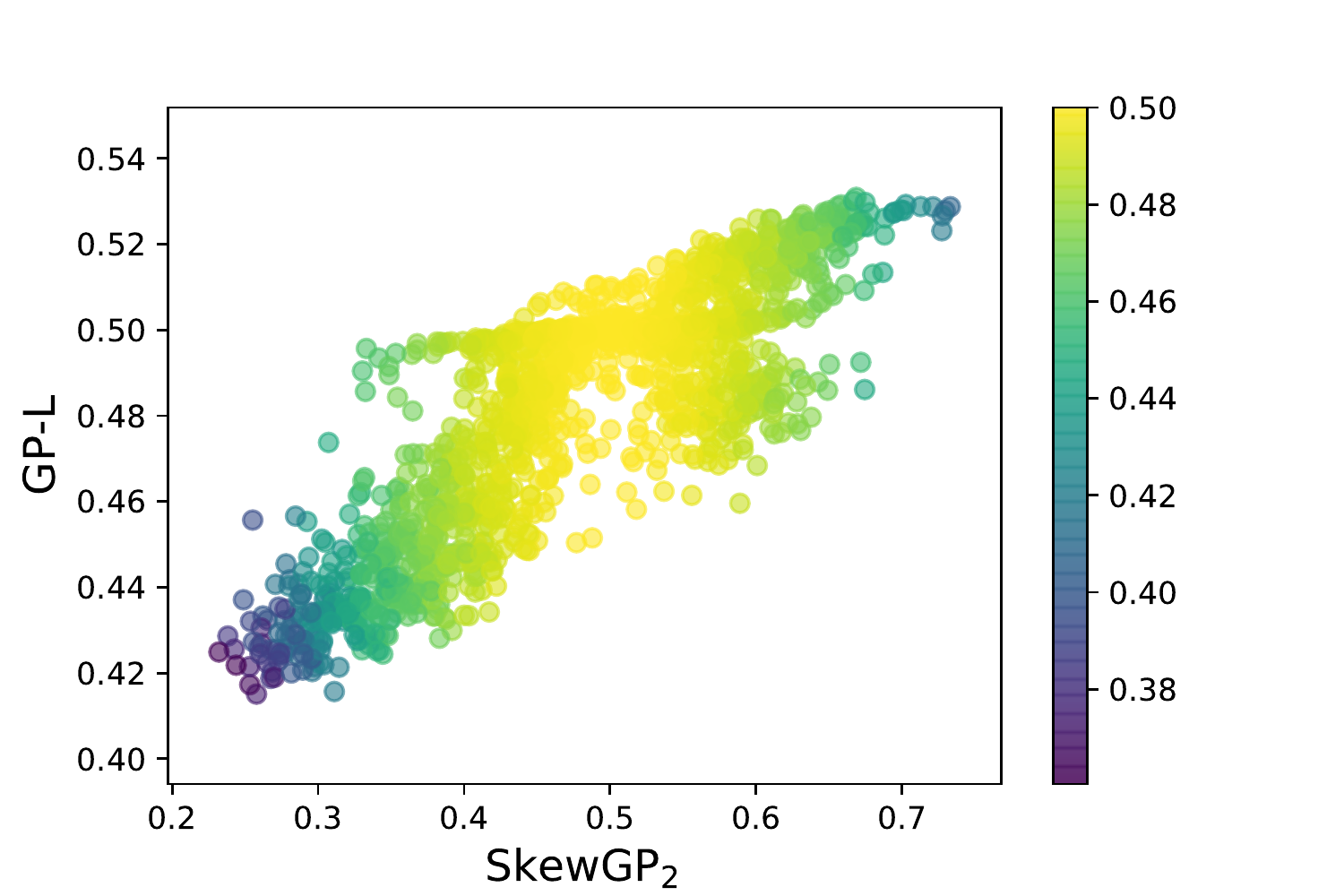}} \\
    \small (c1) & \small (c2)\\
   \end{tabular}
\caption{MNIST dataset}
\label{fig:6}
\end{figure}

\begin{figure}
\centering
\begin{tabular}{c @{\qquad} c }
    \includegraphics[width=.48\linewidth]{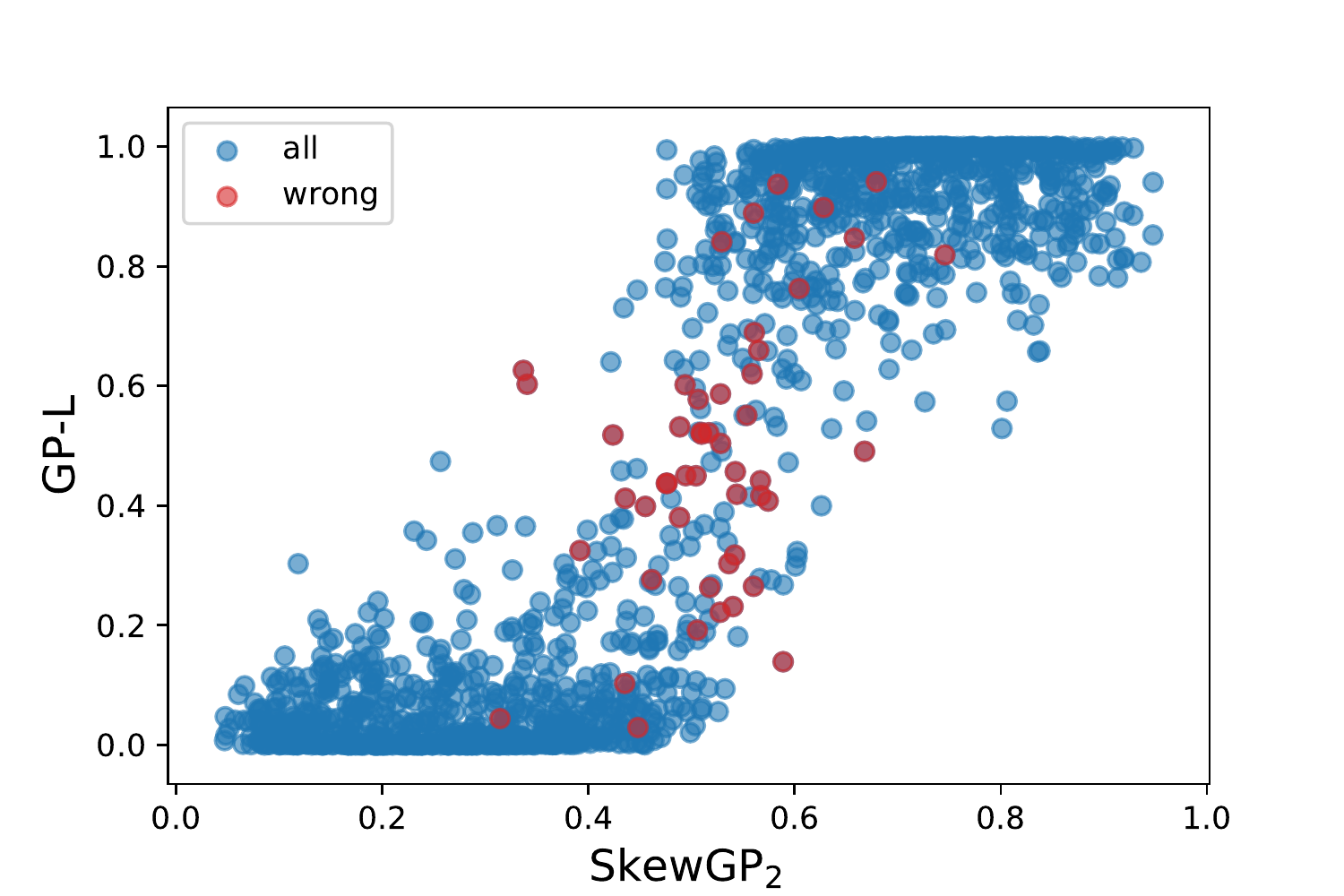} &
    \includegraphics[width=.42\linewidth,trim={1.0cm 0.2cm 1.0cm 0 }, clip]{{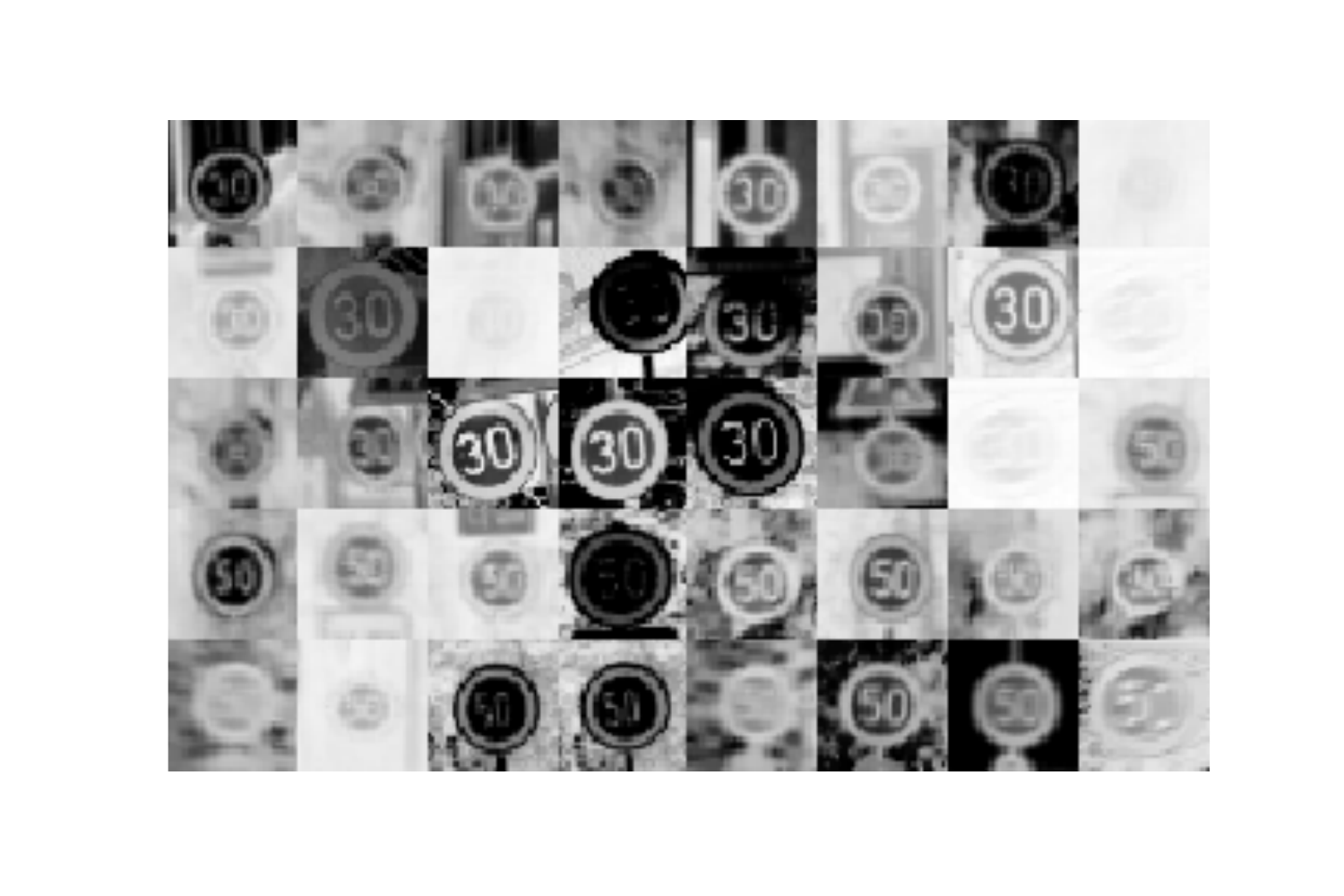}} \\
    \small (a1) & \small (a2)\\
    \includegraphics[width=.48\linewidth]{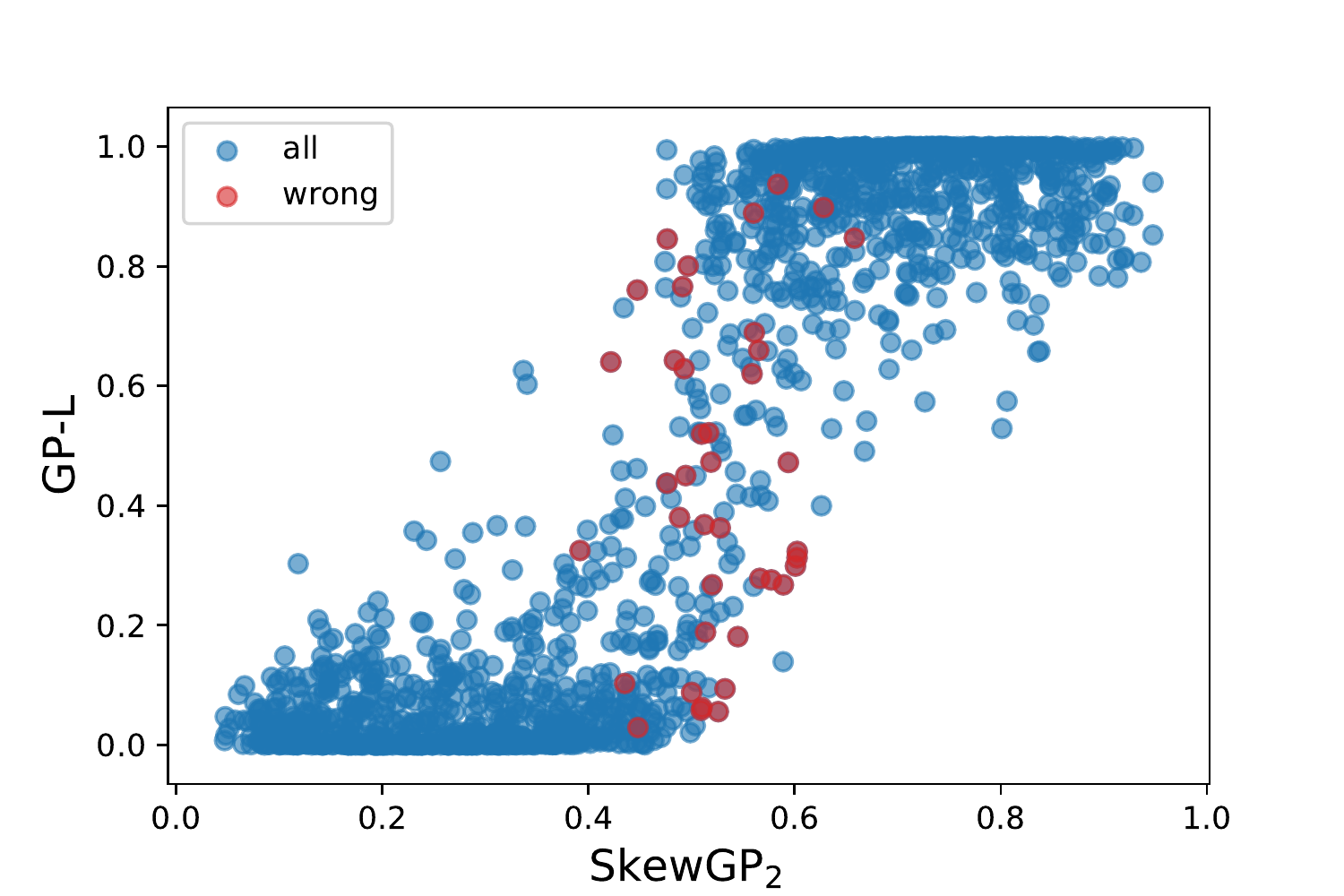} &
    \includegraphics[width=.42\linewidth,trim={1.0cm 0.2cm 1.0cm 0 }, clip]{{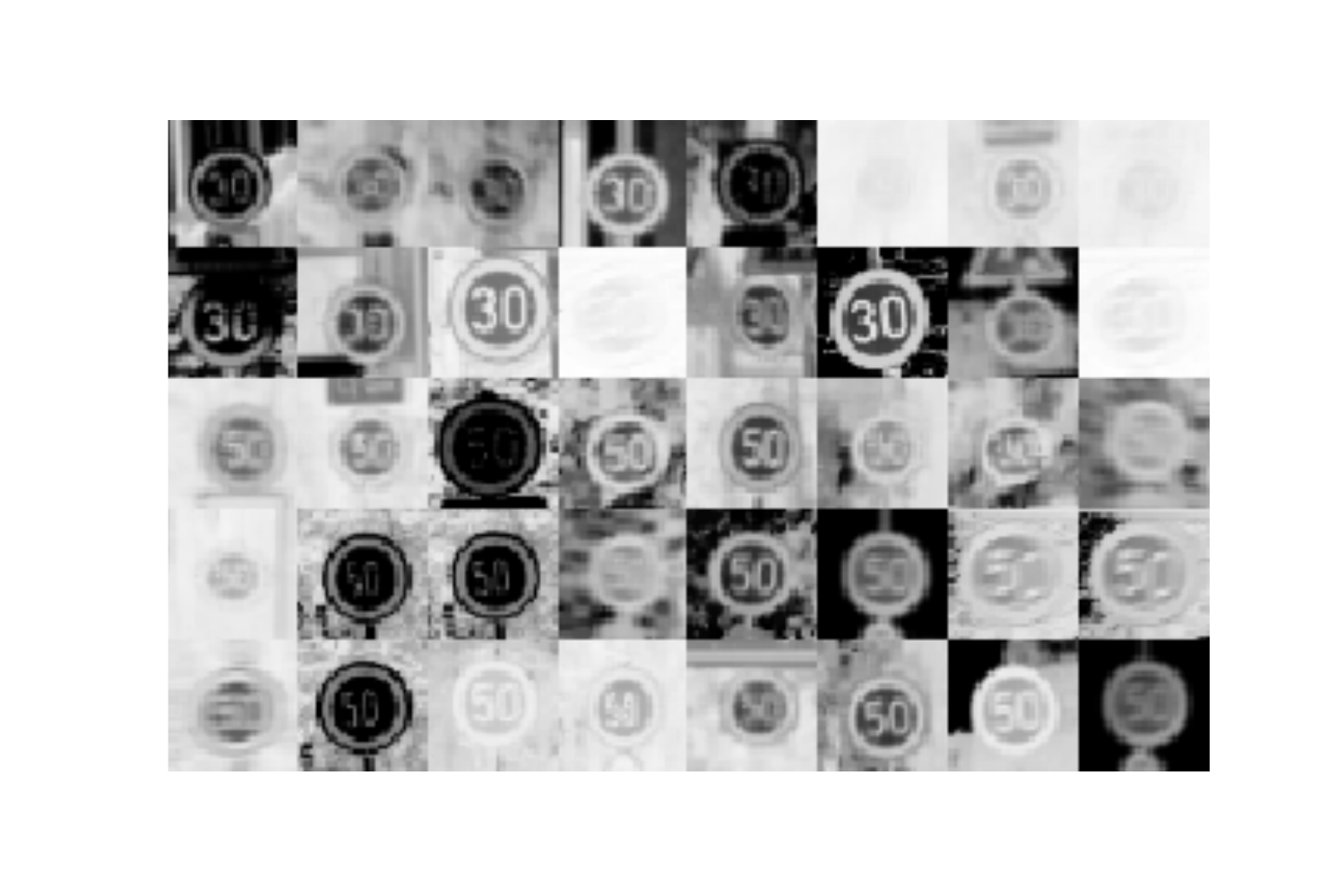}} \\
    \small (b1) & \small (b2)\\
    \includegraphics[width=.48\linewidth]{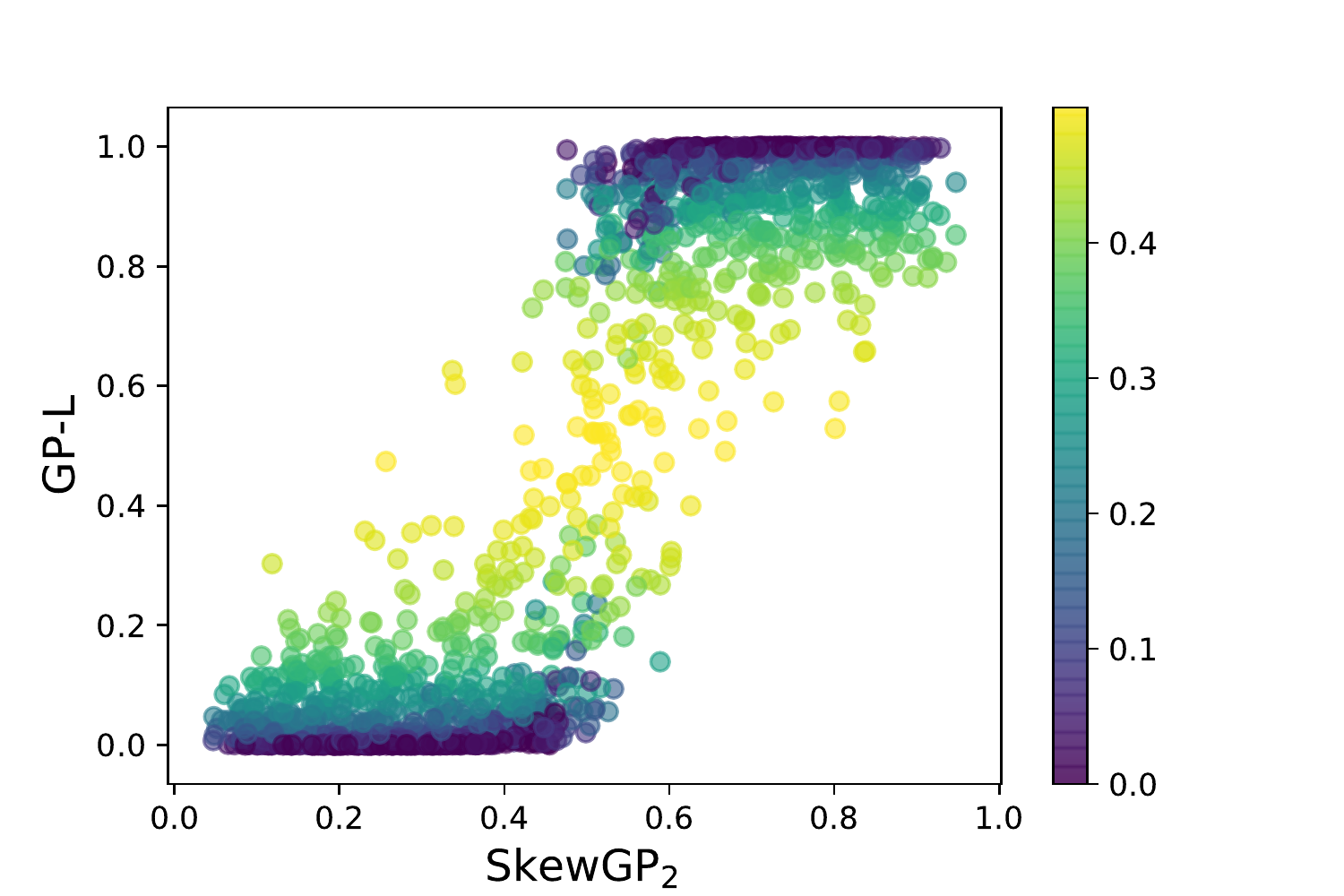} &
    \includegraphics[width=.48\linewidth]{{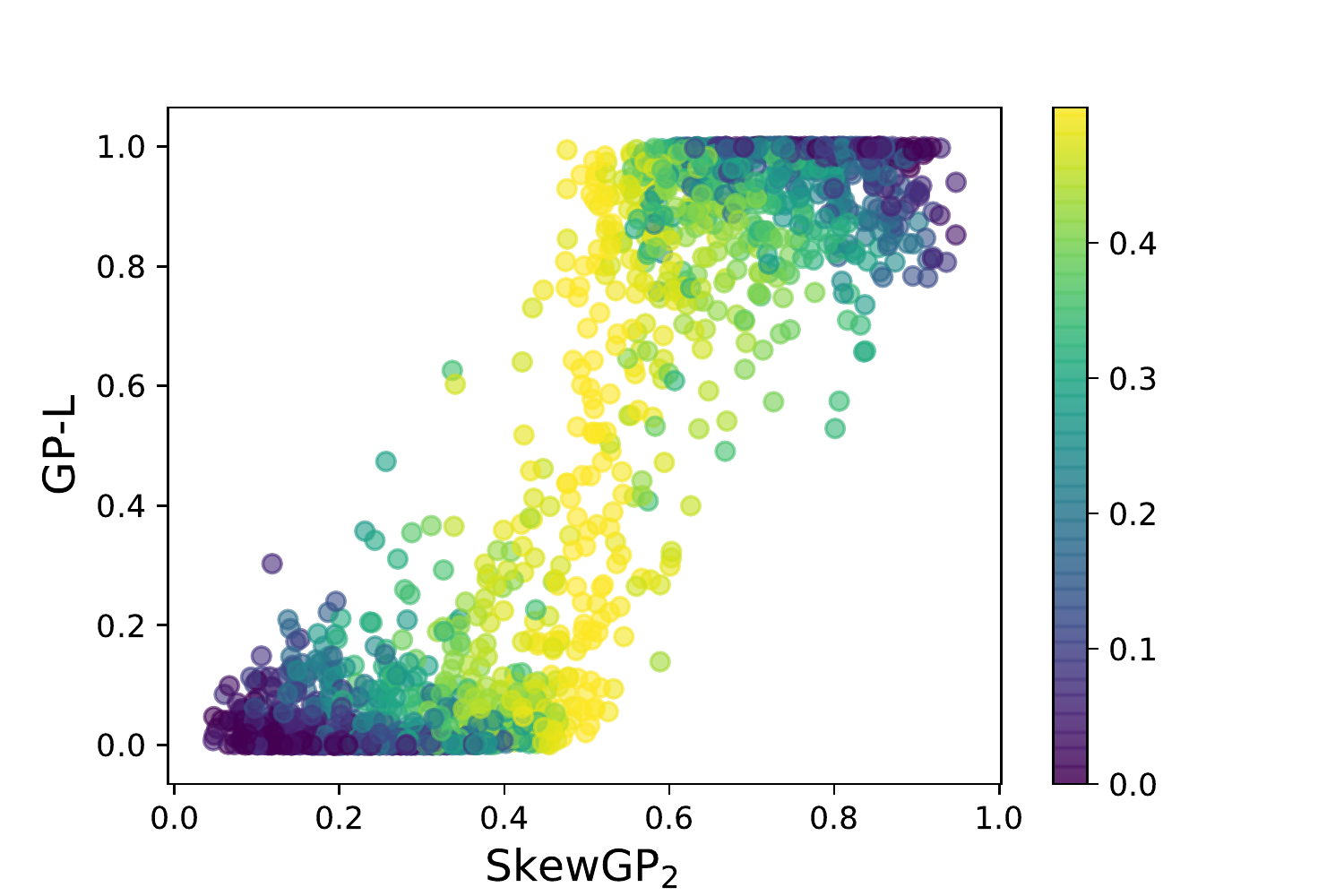}} \\
    \small (c1) & \small (c2)\\
   \end{tabular}
\caption{German-road sign dataset}
\label{fig:7}
\end{figure}

Figure \ref{fig:7} shows a similar  plot for the ``German road-sign'' dataset. Plot \ref{fig:7}(c1) is relative to GP-L and shows that GP-L confidence is low only for the instances
whose mean predicted probability is in $[0.3,0.7]$.
This is not reflected in the value 
of the mean predicted probability for the misclassified
instances (compare plot (a1) and (c1) and note that
some of the red points in (a1) are outside the yellow area in (c1)).
Conversely, \ref{fig:7}(c2) shows that the second order uncertainty of    SkewGP$_2$ is clearly consistent with plot (b1).

\end{document}